\ifdefined\pdfoutput\pdfoutput=1\fi % arXiv builds with pdfLaTeX when it sees this line
\documentclass{banglab}
\usepackage{multirow}
\usepackage{pgfplots}
\pgfplotsset{compat=1.18}

\theoremstyle{plain}

\theoremstyle{definition}

\theoremstyle{remark}

\title{Input-Conditioned Search for Efficient Test-Time Scaling of Recursive Reasoners}
\logos{udem,mila}

\author[1,2]{Haochen Shi}
\author[3]{Tianwen Fu}
\author[1,2]{Qianggang Ding}
\author[1,2]{Yinhan Lu}
\author[1,2]{Yizhan Li}
\author[1,2]{Huan Zhang}
\author[3]{Yajie Zhao}
\author[4]{Xingdi Yuan}
\author[5,\dagger]{Hengyang Lu}
\author[1,2,\dagger]{Bang Liu}

\affiliation[1]{Mila -- Quebec AI Institute}
\affiliation[2]{Universit\'e de Montr\'eal}
\affiliation[3]{University of Southern California}
\affiliation[4]{Microsoft Research}
\affiliation[5]{Ambarella Inc.}
\contribution[\dagger]{Corresponding author}

\abstract{%
People often tackle difficult problems by thinking more deeply or exploring more possibilities.
Similarly, recursive reasoners, which reason by repeatedly applying a learned update rule to latent states, can also scale inference through more updates (depth) or additional trajectories generated by perturbing latent states (width). 
When people hit a dead end in their thinking, they may step back and approach the problem from a fresh angle.
However, neither depth scaling nor perturbation-based width scaling of recursive reasoners can reliably achieve this: both may converge to incorrect attractors.
To address this, we propose to explore diverse reasoning traces by applying different update maps to the same latent state, without any additional training. We realize this idea through \emph{Input-Conditioned Search (ICS)}, which searches over input conditions to induce alternative update maps from the pretrained reasoner.
Across six reasoning tasks, ICS achieves the highest mean accuracy, while requiring up to two orders of magnitude fewer model FLOPs than the evaluated scaling baselines. 
Controlled experiments confirm that ICS can redirect trajectories from the same state. Our theoretical analysis and rigorous numerical
verification also confirm that ICS can locally eliminate incorrect attracting fixed points.
Together, these findings establish ICS as an effective axis for efficient test-time scaling and motivate further research on shaping recursive reasoning during training and at test time.
}

\date{\today}
\correspondence{\email{hylu@ambarella.com}, \email{bang.liu@umontreal.ca}}
\github{\href{https://github.com/icsearch/ICS}{github.com/icsearch/ICS}}
\projectpage{\href{https://icsearch.github.io/}{icsearch.github.io}}
\huggingface{\href{https://huggingface.co/hcshi/ICS}{huggingface.co/hcshi/ICS}}

\begin{document}

\maketitle

\begin{figure}[t]
\centering
\includegraphics[width=0.86\linewidth]{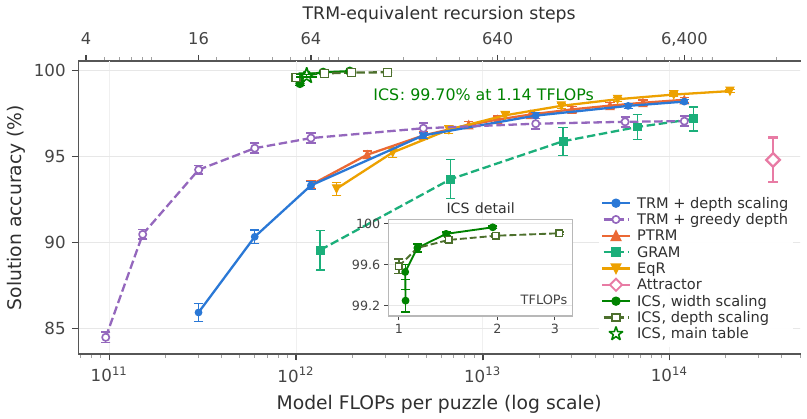}
\caption{Sudoku test-time scaling without a task certificate at selection. All curves are the average of three trained seeds. The Attractor point uses the $3{,}072$-puzzle subset and is marked separately.}
\label{fig:scaling}
% \vspace{-1.0em}
\end{figure}

\section{Introduction}
\label{sec:introduction}
People often respond to difficult problems by thinking longer or exploring more possibilities, yet their reasoning can still reach a dead end.
Recursive reasoners face a similar limitation. They produce answers by repeatedly applying a learned update rule to a latent state, and can scale test-time computation with fixed weights through longer trajectories (depth) or multiple perturbed trajectories (width)~\citep{trm2025,ptrm2026,gram2026}. 
However, both strategies retain the reasoner's original input-conditioned update map: depth iterates the same map for longer, while perturbation-based width explores additional trajectories under the same map. 
% Our diagnostics identify incorrect attracting states from which continued iteration and the tested perturbations fail to recover a valid solution. 
Our diagnostics identify failures that persist under both continued iteration and random perturbations, as trajectories converge to incorrect attractors. This raises a central question: 
\emph{how can a frozen recursive reasoner use additional test-time computation to overcome such failures and discover correct solutions efficiently?}
% \emph{if more computation under the same update map cannot escape such failures, how should a frozen recursive reasoner use it to reason correct solutions efficiently?}

When their reasoning stalls, people may step back and approach the same problem from a fresh angle.
We ask whether a recursive reasoner can do the same without retraining. To achieve this goal, we investigate shaping recursive dynamics as an additional dimension of test-time scaling: changing the update map that governs reasoning without retraining the model.
Because each recursive update depends on both the learned
parameters and the input, varying the input condition
provides a way to induce alternative update maps while
keeping the parameters fixed. 
We introduce \emph{Input-Conditioned Search (ICS)}, which searches over input conditions to explore these alternative maps and restarts reasoning from the same initial latent state under each condition.
% We ask whether a recursive reasoner can do the same by changing the update map itself at inference time, without retraining.
% Because each recursive update depends on both the learned parameters and the input, varying the input condition induces alternative update maps while keeping the parameters fixed.
% We introduce \emph{Input-Conditioned Search (ICS)}, which uses test-time computation to search over input conditions and restarts reasoning under each one.

We instantiate ICS through two operations. 
\emph{Hypothesis injection} incorporates partial model predictions into the input as tentative assumptions.
\emph{Validity-preserving restatements} express the same task in equivalent representations and map predictions back to the original representation.
Model predictions guide both which hypotheses to expand and which answer to select. Throughout the search, the original task remains the
criterion for answer correctness.

Our analysis establishes that input conditioning can locally eliminate incorrect attracting fixed points. We verify this mechanism with computer-assisted proofs in separately trained $4\times4$ Sudoku reasoners under continuous variation of the input embeddings. Controlled interventions on hypotheses and restatements also show that changing the input condition can redirect trajectories from the same initialization. Recovery further depends on the starting state and the duration of conditioning, motivating input-conditioned search combined with fresh initialization.

% Controlled interventions on hypotheses and restatements show that changing the input condition can redirect trajectories from the same initialization.
% They further show that recovery depends on the starting state and the duration of conditioning, motivating search over input conditions combined with fresh initialization.

Across six reasoning tasks, ICS achieves the highest mean accuracy both with and without a task correctness checker at answer selection, tying at $100\%$ on Light-Up. Without a checker, ICS achieves higher mean accuracy than the evaluated scaling baselines on the remaining five tasks with approximately $10$--$195\times$ fewer model FLOPs; on Sudoku-Extreme, it reduces errors by approximately $92\%$ relative to the strongest baseline at comparable compute.
Together, these results demonstrate efficient test-time scaling through input-conditioned search without additional training, and motivate broader methods for shaping recursive dynamics to improve access to correct solutions.

\section{Preliminaries}
\label{sec:preliminaries}
\label{sec:setting}

\begin{figure}[t]
  \centering
  \includegraphics[width=\linewidth]{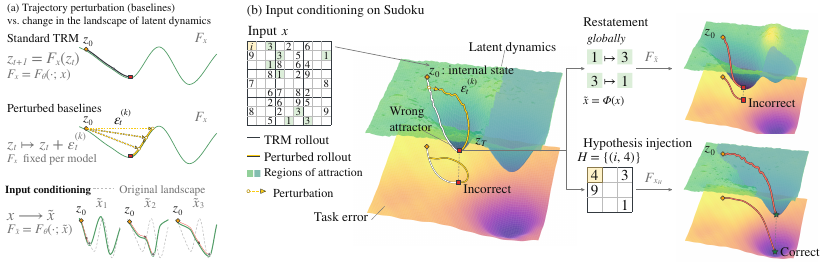}
  \caption{ICS lifts efficient test-time scaling of recursive reasoners by shaping the update rules.}
  \label{fig:his-mechanism}
  % \vspace{-1.0em}
\end{figure}

\textbf{Frozen recursive reasoners.}
A recursive reasoner repeatedly updates an internal latent
state to refine its prediction.
Formally, a model $\mathcal{M}_\theta$  receives an input $x\in\mathcal{V}^{L_{x}}$, a sequence of $L_x$ tokens over the vocabulary $\mathcal{V}$ with task-specific given positions $G(x) \subset \{1, \dots, L_x\}$. $\mathcal{M}_\theta$ \emph{reasons} towards the answer $\hat{y} \in \mathcal{V}^{L_{y}}$ by iterating a learned update $F_\theta(z_{t-1};x)$ of its latent state $z \in \mathcal{Z} = \mathbb{R}^d$ from an initial state $z_0$, and reads the answer and a halting signal from the state:

\begin{equation}
z_{t} = F_\theta(z_{t-1};\, x), \qquad \hat{y}_t = g_\theta(z_t;\, x), \qquad q_t = Q_\theta(z_t),
\qquad t = 1, 2, \dots
\label{eq:map}
\end{equation}

where the decoder $g_\theta$ decodes the answer $\hat{y}$ from the latent $z_t$ and $Q_\theta$ produces a halting score that can be used for stopping or candidate selection. The learned parameters $\theta$ are frozen at test time; thus, for a given input $x$, they induce a fixed latent update map $F_x := F_\theta(\cdot;,x)$. One application of $F_x$ is a \emph{recursion step}, and the resulting $\tau = (z_0, z_1, \dots)$ is a \emph{reasoning trajectory} from the initialization $z_0$. 
The Tiny Recursive Model (TRM)~\citep{trm2025}, our primary
backbone, uses a small shared network to recursively update
an internal reasoning state and an answer representation.
In our notation, the complete state is $z=(z_H,z_L)$:
$z_H$ represents the current answer, $z_L$ supports its
iterative refinement, and the decoder reads $z_H$.
A recursion step can contain multiple applications of the
shared network. In this work, we study reasoning puzzles, where the input token length is equal to the output length $L:=L_y = L_x$.

\textbf{Test-time scaling of recursive reasoners.}
Each latent state $z_t$ can be decoded by $g_\theta$ into a candidate answer $\hat y_t$. Additional recursion steps therefore provide further opportunities to discover a correct answer, but need not improve coverage of correct solutions (Appendix~\ref{app:sigma}).
Two common scaling axes are \emph{depth}, which increases the number $D$ of updates per trajectory, and \emph{width}, which increases the number $K$ of trajectories.
Width scaling can be implemented by perturbing a latent state $z_t$ with a perturbation $\varepsilon^{(k)}_t$ and continuing recursion from the resulting states. 
Previous work uses perturbations sampled from fixed noise distributions or learned distributions~\citep{ptrm2026,gram2026}.
For a fixed input $x$, consider $K$ trajectories with $D$ updates each. Their latent states form the collection $\mathcal{P}(x)
=
\bigl(z_t^{(k)}\bigr)_{
1\leq k\leq K, 1\leq t\leq D
}$.
% \begin{equation}
% \mathcal{P}(x)
% =
% \bigl(z_t^{(k)}\bigr)_{
% 1\leq k\leq K, 1\leq t\leq D
% }.
% \label{eq:scaling}
% \end{equation}

\textbf{Answer selection and evaluation.}
Discovering a correct candidate and selecting the answer to return are distinct requirements of test-time scaling. 
Let $V(x,y)\in\{0,1\}$ indicate whether $y$ satisfies the original task defined by $x$.
Let $\mathcal{C}(x)$ denote the candidate answers made available by the inference procedure, expressed in the original task representation.
Their solution coverage is $\mathrm{Coverage}(x)
    =
    \mathbf{1}\!\left[
    \exists\,y\in\mathcal{C}(x):V(x,y)=1
    \right].$
% \begin{equation}
%     \mathrm{Coverage}(x)
%     =
%     \mathbf{1}\!\left[
%     \exists\,y\in\mathcal{C}(x):V(x,y)=1
%     \right].
%     \label{eq:generation-selection}
% \end{equation}
% Let $V(x,y)\in\{0,1\}$ be a task-specific correctness checker that indicates whether $y$ satisfies the original task defined by $x$. Certificate-based selection has access to this checker during inference.
% % Let $\mathcal{C}(x)=(g_\theta(z_t^{(k)};x))_{k,t}$ denote the decoded candidates. Their solution coverage is 
% \begin{equation}
%     \mathrm{Coverage}(x)
%     =
%     \mathbf{1}\!\left[
%     \exists\,y\in\mathcal{C}(x):V(x,y)=1
%     \right].
%     \label{eq:generation-selection}
% \end{equation}
We distinguish two answer-selection regimes: $ \hat{y}_{\mathrm{cert}}=
    c_{\mathrm{cert}}\bigl(\mathcal{C}(x);x, V\bigr),
    \ 
    \hat{y}_{\mathrm{model}}   =c_{\mathrm{model}}\bigl(\mathcal{P}(x);x\bigr).$
% \begin{equation}
%     \hat{y}_{\mathrm{cert}}=
%     c_{\mathrm{cert}}\bigl(\mathcal{C}(x);x, V\bigr),
%     \qquad
%     \hat{y}_{\mathrm{model}}   =c_{\mathrm{model}}\bigl(\mathcal{P}(x);x\bigr).
% \end{equation}
For a fixed candidate pool, certificate-based selection returns a valid answer whenever one is present. Model-based selection instead uses model-derived information, such as halting scores, learned scores, or prediction agreement, without access to $V$ during selection.
With a task certificate $V$ available, we evaluate \emph{reasoning efficiency}: the ability to discover a verified solution as a function of inference cost. Without the certificate at selection, we evaluate the complete procedure by its final-answer accuracy and inference cost, using $V$ only to assess correctness. Each regime is evaluated with its corresponding search and stopping rules.

\section{Input-Conditioned Search}
\label{sec:method}

Input-Conditioned Search (ICS) allocates test-time computation to exploring alternative input conditions for the original task $x$ (Figure~\ref{fig:his-mechanism}). Each conditioned input $\tilde{x}$ induces an update map $F_{\tilde{x}}:=F_\theta(\cdot;\tilde{x})$ under the same
frozen model. ICS restarts reasoning from $z_0$ under these conditions to generate candidate answers.
We instantiate this search through two operations: 
\emph{hypothesis injection}, which incorporates partial model predictions into the input as tentative assignments, and 
\emph{validity-preserving restatements}, which express the same task in equivalent representations. Predictions are mapped back to the original task representation when necessary, and answer correctness is always defined by the original input $x$.

% HIS adds input conditioning as a dimension of test-time search (Fig.~\ref{fig:his-mechanism}). It constructs tentative hypotheses $H$ from the recursive reasoner $\mathcal{M}_\theta$'s predictions, incorporates each hypothesis into an input $x_H$, and runs the frozen reasoner under each resulting condition. Each hypothesis defines an input-conditioned update map $F_{x_H}:=F_\theta(\cdot;x_H)$. HIS additionally explores validity-preserving restatements $x'=\Phi(x)$ under $F_{x'}$ to further improve the input-conditioning search, mapping their prediction back through the corresponding inverse answer transformation $\Phi^{-1}$. 
% Throughout this search, the model parameters remain frozen, and returned answers are evaluated against the original task $x$. 

\subsection{Input-conditioned Reasoning}
\label{sec:hypotheses}

For a conditioned input $\tilde{x}$, we run the frozen reasoner from a fresh initialization:
\begin{equation}
    z_0^{(\tilde{x})} = z_0,
    \qquad 
    z_t^{{\tilde{x}}} = 
    F_{\tilde{x}}\!\left(z_{t-1}^{(\tilde{x})}\right),
\quad t=1,\ldots,T_{\tilde{x}},
\label{eq:conditioned-rollout}
\end{equation}

where $T_{\tilde{x}}$ is the allocated recursion depth. The resulting state provides a decoded candidate and a halting score. 
The final state provides a raw decoded prediction
$\hat{y}_{\tilde{x}} = g_\theta(z_{T_{\tilde{x}}}^{(\tilde{x})};\tilde{x})$
and a halting score
$q_{\tilde{x}} = Q_\theta(z_{T_{\tilde{x}}}^{(\tilde{x})})$.
We construct the conditioned inputs $\tilde{x}$ through two operations.

\textbf{Hypothesis injection.}
A hypothesis specifies tentative assignments $H=\{(i,v_i): i \in I_H\}$, where $I_H$ is a set of task-specific mutable positions disjoint from $G(x)$ and each $v_i$ is an admissible state token at position $i$.
We construct the conditioned input by
% $    (x_H)_i = 
%     \begin{cases}
%         v_j, & i \in I_H, \\
%         x_i, & i \notin I_H.
%     \end{cases}$.
\begin{equation}
    (x_H)_i = 
    \begin{cases}
        v_j, & i \in I_H, \\
        x_i, & i \notin I_H.
    \end{cases}
\label{eq:hypothesis-input}
\end{equation}
This preserves the original givens while expressing the tentative assignments as additional input information. Specifically, the assignments specify digits in Sudoku-Extreme (Sodoku), path membership in Maze-Hard, light placement in Light-Up, and cell colors in Nurikabe, Tapa, and Heyawake. We then apply Equation~\ref{eq:conditioned-rollout} with $\tilde{x}=x_H$. The resulting predictions and scores guide hypothesis expansion and answer selection.

% For each conditioned input, we initialize a fresh reasoning trajectory and evaluate it for $T_h$ recursion steps:
% \begin{equation}
%     z_0^{(H)} = z_0,
%     \qquad
%     z_t^{(H)} = F_{x_H}\!\left(z_{t-1}^{(H)}\right),
%     \quad t=1, \ldots, T_h.
%     \label{eq:hypothesis-rollout}
% \end{equation}
% After $T_h$ recursion steps, we obtain a candidate answer $\hat{y}_H = g_\theta(z_{T_h}^{(H)};x_H)$ and its halting score $q_{H} = Q_\theta(z_{T_h}^{(H)})$. Hypothesis injection thus turns model predictions into input conditions for subsequent reasoning.

\textbf{Validity-preserving restatements.}
A restatement expresses the same task in an equivalent representation through an invertible, task-specific transformation $\Phi$.
% We also construct alternative representations of the same task using an invertible, task-specific transformation $\Phi$. 
The transformation acts correspondingly on inputs and answers and preserves task validity: 
% $V\!\left(\Phi(x), \Phi(y)\right) = V(x,y)$.
\begin{equation}
    V\!\left(\Phi(x), \Phi(y)\right) = V(x,y)
\label{eq:validity-preserving}
\end{equation}
We apply Equation~\ref{eq:conditioned-rollout}
with $\tilde{x}=\Phi(x)$ and map the decoded prediction
back to the original representation with $\hat{y} = \Phi^{-1}\!\left(\hat{y}_{\Phi(x)}\right)$. The frozen reasoner need not be equivariant to $\Phi$, so equivalent task representations can produce different candidate answers after inverse mapping.

% We run the frozen reasoner on $x'=\Phi(x)$ and map its decoded prediction $\hat{y}'$ back to the original representation as $\hat{y}=\Phi^{-1}(\hat{y}')$. The recursive reasoner is not required to be equivariant to $\Phi$, so these equivalent task representations can produce distinct candidates after inverse mapping $\Phi^{-1}$.

\textbf{Search and answer selection}
\label{sec:search}

ICS starts with running the frozen reasoner $\mathcal{M}_\theta$ on the original input $x$. If none of the resulting candidates meets the configured acceptance rule $c_{cert}$ or $c_{model}$, it proceeds to hypothesis-conditioned search. Each search node corresponds to a hypothesis $H$ and retains the model predictions obtained from $x_H$. The root corresponds to $H=\varnothing$. Expanding a node introduces an additional assignment: 
% $ H' = H \cup \{(i, v)\},  i \notin G(x)\cup I_H$.
\begin{equation}
    H' = H \cup \{(i, v)\},
    \qquad i \notin G(x)\cup I_H,
    \label{eq:hypothesis-extension}
\end{equation}
If hypothesis search ends without an accepted answer,
ICS explores validity-preserving restatements of the
original input $x$. This schedule allocates computation to additional input
conditions when earlier stages have not accepted an answer.

% HIS starts with running the frozen reasoner $\mathcal{M}_\theta$ on the original input $x$. If none of the resulting candidates meets the configured acceptance rule $c_{cert}$ or $c_{model}$, it proceeds to hypothesis-conditioned search. Each search node corresponds to a hypothesis $H$ and retains the model predictions obtained from $x_H$. The root corresponds to $H=\varnothing$. Expanding a node introduces an additional assignment:
% \begin{equation}
%     H' = H \cup \{(i, v)\},
%     \qquad i \notin G(x)\cup I_H,
%     \label{eq:hypothesis-extension}
% \end{equation}
% where $i$ is a task-specific mutable position and $v$ is a proposed state token.
% Each child is evaluated using the conditional rollout in Equation~\ref{eq:hypothesis-rollout}.

\textbf{Proposal and expansion.}
At each parent node $H$, we use the decoder's token score to propose additional assignments. Specifically, we rank eligible positions by increasing prediction margin: 
% $m_i^{(H)} = \ell_{i,(1)}^{(H)}-\ell_{i,(2)}^{(H)}$.
\begin{equation}
    m_i^{(H)}
    =
    \ell_{i,(1)}^{(H)}-\ell_{i,(2)}^{(H)},
    \label{eq:proposal-margin}
\end{equation}
where $\ell_{i,(1)}^{(H)}$ and $\ell_{i,(2)}^{(H)}$ are the largest and second-largest decoder logits at position $i$ from the parent rollout.
At each position, we propose alternative state tokens permitted by the task's vocabulary, excluding the current predicted state. When multiple alternatives are available, we order them by the decoder's token logits ranking. Each proposed assignment $(i,v)$ create a child $H' = H\cup \{(i,v)\}$. We cap the number of proposals per parent and truncate the pooled proposals to the remaining total child budget. For each remaining child $H'$, we run the frozen reasoner on $x_{H'}$ for $T_h$ recursion steps from $z_0$, as in Equation~\ref{eq:conditioned-rollout}.

\textbf{Candidate ranking.}
For each child $H'$, let $\hat y_{H'}$ denote its raw decoded prediction and $q_{H'}$ its halting score.  Let $\Pi_x(\hat{y})$ restore the original givens in a prediction $\hat{y}$.
% Given a raw prediction $\hat{y}$, we restore the original givens from $x$ and denote the resulting prediction by $\Pi_x(\hat{y})$.
% We rank children search nodes in ascending lexicographic order of $\left( \sum_{(i,v)\in H'} \mathbf{1}[\hat y_{H', i}\ne v], 
% \quad
% \mathbf{1}[\hat y_{H'}\ne \Pi_x(\hat y_{H'})],
% \quad
% -q_{H'}
% \right)$
\begin{equation}
\left(
\sum_{(i,v)\in H'}
\mathbf{1}[\hat y_{H', i}\ne v],
\quad
\mathbf{1}[\hat y_{H'}\ne \Pi_x(\hat y_{H'})],
\quad
-q_{H'}
\right)
\label{eq:child-ranking}
\end{equation}
So we first prioritize child nodes with fewer assignments that contradict their parent's hypothesis $H$ and the givens. Among ties, we prefer predictions with a higher halting score. These consistency checks use the original input $x$, the injected assignments in relevant hypotheses $H$, and the task's output format, without requiring a task certificate. If no candidate meets the acceptance rule, we retain up to $W$ highest-ranked children for expansion at the next search level, where $W$ is the beam width.

\textbf{Termination and answer selection.}
We apply the acceptance rule to candidates expressed in the original task representation and corrected by $\Pi_x$. When a task certificate $V$ is available, we accept a candidate when 
$V(x, \Pi_x(\hat{y}))=1$. When the task certificate is unavailable, we use the model-derived scores or consistency between predictions to stop recursion and select an answer.

\section{Performance and Inference Efficiency}
\label{sec:performance}

\begin{table}[t]
\centering
\small
\setlength{\tabcolsep}{3pt}
\resizebox{\linewidth}{!}{%
\begin{tabular}{lcccccc}
\toprule
Protocol & Sudoku & Maze & Light-Up & Nurikabe & Tapa & Heyawake\\
\midrule
\multicolumn{7}{l}{\emph{Task certificate available at selection}}\\
Standard TRM & $86.0\pm0.6$ & $91.6\pm0.3$ & $99.7\pm0.5$ & $83.8\pm2.8$ & $72.8\pm5.4$ & $86.4\pm3.3$\\
 \quad + depth scaling  & $98.2\pm0.1$ & $91.6\pm0.3$ & $\mathbf{100.0\pm0.0}$ & $84.9\pm2.7$ & $75.4\pm5.5$& $87.8\pm1.5$\\
PTRM  & $98.3\pm0.1$ & $92.4\pm0.3$ & $\mathbf{100.0\pm0.0}$ & $86.1\pm2.3$ & $76.4\pm6.9$ & $88.7\pm1.5$\\
GRAM & $97.5\pm0.4$ & $90.8\pm0.6$ & $99.7\pm0.5$ & $91.9\pm1.8$ & $52.3\pm5.5$ & $97.1\pm1.3$\\
EqR  & $98.8\pm0.1$ & $95.0\pm1.9$ & $91.4\pm0.5$ & $77.1\pm0.5$ & $26.7\pm0.9$ & $31.6\pm2.7$\\
Attractor  & $94.8\pm1.3^{\dagger}$ & $80.7\pm1.4$ & $\mathbf{100.0\pm0.0}$ & $89.0\pm2.2$ & $82.2\pm4.6$ & $94.2\pm2.2$\\
\textbf{ICS} & $\mathbf{99.94\pm0.02}$ & $\mathbf{100.0\pm0.1}$ & $\mathbf{100.0\pm0.0}$ & $\mathbf{95.9\pm1.3}$ & $\mathbf{95.9\pm0.9}$ & $\mathbf{99.1\pm0.0}$\\
\midrule
\multicolumn{7}{l}{\emph{No task certificate at selection}}\\
Standard TRM & $86.0\pm0.6$ & $91.6\pm0.3$ & $99.7\pm0.5$ & $83.8\pm2.8$ & $72.8\pm5.4$ & $86.4\pm3.3$\\
 \quad + depth scaling& $98.1\pm0.1$ & $85.1\pm0.9$ & $99.7\pm0.5$ & $80.9\pm1.7$ & $67.7\pm4.1$ & $84.3\pm3.0$\\
PTRM& $98.3\pm0.1$ & $91.8\pm0.3$ & $\mathbf{100.0\pm0.0}$ & $84.9\pm3.1$ & $74.9\pm6.2$ & $88.7\pm1.5$\\
GRAM, LPRM & $97.2\pm0.7$ & $84.7\pm1.6$ & $99.1\pm0.9$ & $86.1\pm3.8$ & $43.1\pm5.3$ & $93.6\pm0.5$\\
GRAM, majority & $96.6\pm0.6$ & $85.2\pm1.4$ & $99.4\pm1.0$ & $86.1\pm4.0$ & $43.1\pm1.5$ & $93.6\pm0.5$\\
EqR & $98.8\pm0.1$ & $87.4\pm1.8$ & $80.2\pm0.5$ & $57.7\pm1.0$ & $17.4\pm0.9$ & $18.8\pm1.3$\\
Attractor & $94.8\pm1.3^{\dagger}$ & $53.6\pm1.6$ & $\mathbf{100.0\pm0.0}$ & $87.6\pm1.8$ & $78.1\pm5.7$ & $92.2\pm2.3$\\
\textbf{ICS} & $\mathbf{99.70\pm0.08}$ & $\mathbf{99.1\pm0.2}$ & $\mathbf{100.0\pm0.0}$ & $\mathbf{88.4\pm5.9}$ & $\mathbf{86.7\pm0.9}$ & $\mathbf{94.5\pm1.0}$\\
\bottomrule
\end{tabular}}
\caption{Answer accuracy (\%) on six tasks. $\dagger$: Attractor Models is evaluated on a fixed $3{,}072$-instance Sudoku-Extreme subset. Appendix~\ref{app:sudoku-subset} compares all methods on the same subset.
}
\label{tab:performance}
% \vspace{-1.2em}
\end{table}

\begin{figure}[t]
\centering
\setlength{\abovecaptionskip}{4pt}
\includegraphics[width=\linewidth]{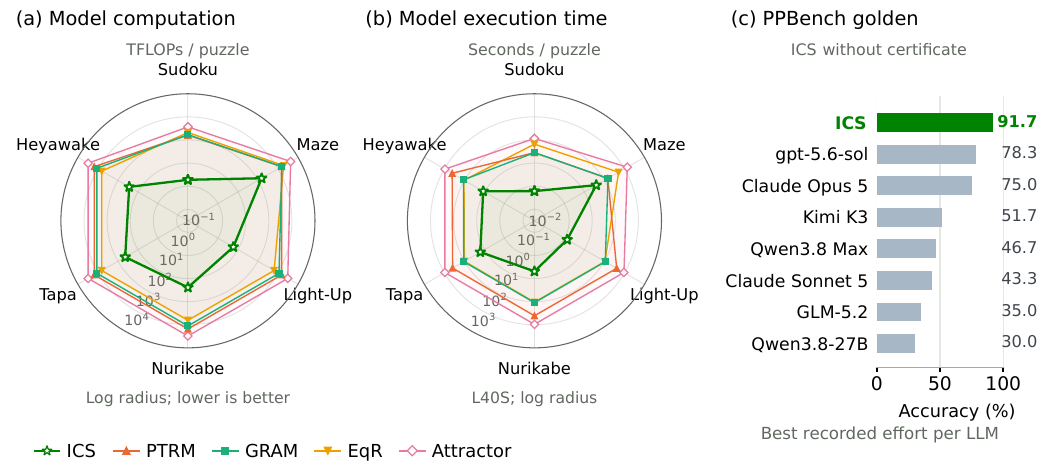}
\caption{\textbf{Six-task efficiency and PPBench golden-set accuracy.}
Both radar panels use logarithmic radii, farther inward indicates lower cost.Appendix~\ref{app:llm-evaluation} gives
complete LLMs results.}
\label{fig:golden-coverage}
\label{fig:efficiency-golden}
% \vspace{-1.4em}
\end{figure}

\begin{table}[t]
\centering
\small
\setlength{\tabcolsep}{6pt}
\begin{tabular}{lrrrr}
\toprule
& \multicolumn{2}{c}{Certificate selection} & \multicolumn{2}{c}{Model selection}\\
\cmidrule(lr){2-3}\cmidrule(lr){4-5}
Components & Accuracy (\%) & TFLOPs & Accuracy (\%) & TFLOPs\\
\midrule
A: TRM + depth scaling & 87.85 & 0.38 & 87.85 & 0.38\\
AB: + hypotheses & 99.92 & 1.39 & 99.75 & 1.10\\
AC: + restatements & 98.83 & 1.19 & 98.82 & 1.19\\
ABC: full ICS & 99.94 & 1.39 & 99.81 & 1.12\\
\bottomrule
\end{tabular}
\caption{Sudoku component ablation on the same eight test windows and trained model.  Appendix~\ref{app:ablation} provides more details for the complete ablation study.}
\label{tab:ablation-main}
% \vspace{-1.0em}
\end{table}

\begin{figure}[t]
 \centering
 \includegraphics[width=\linewidth]{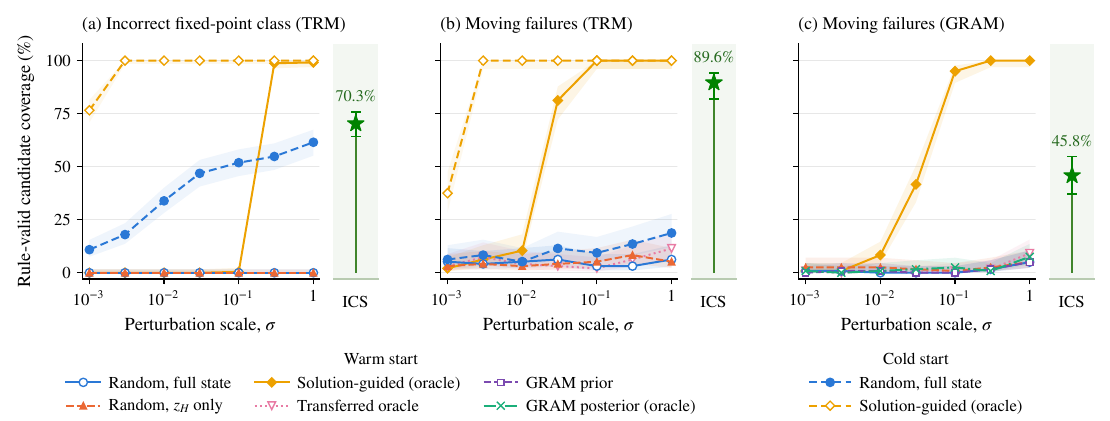}
 \caption{\textbf{State interventions and ICS recover different failures.} Oracle direction requires label access and more computation(Appendix~\ref{app:sigma}).}
 \label{fig:sigma}
 % \vspace{-1.5em}
\end{figure}

\textbf{Experimental setup.}
\label{sec:experimental-setup}
We evaluate ICS on six reasoning tasks: Sudoku-Extreme and Maze-Hard~\citep{hrm2025}, and four tasks from PPBench~\cite{ppbench2026}, including Light-UP, Nurikabe, Tapa, and Heyawake. We use Maze-Hard for our main maze experiments because standard TRM already achieves near-perfect accuracy on Maze-Unique~\citep{eqr2026}, leaving little headroom for comparing accuracy gains from test-time scaling (Appendix~\ref{app:maze-unique}). We train and evaluate all recursive models with the same training and test splits across all six tasks. Dataset construction, preprocessing, and training configurations are detailed in Appendix~\ref{app:experimental-details}.

\textbf{Baselines.}
We compare ICS with standard TRM~\citep{trm2025}, two TRM depth-scaling baselines, PTRM~\citep{ptrm2026}, GRAM~\citep{gram2026}, EqR~\citep{eqr2026}, and Attractor Models~\citep{attractor2026}. ICS operates on frozen TRM backbones. \emph{TRM + depth scaling} runs each instance for the full allocated depth budget. It returns the candidate with the highest halting score $q_t$ along the entire trajectory under the model-based answer selection. \emph{TRM + greedy depth} instead stops each instance at the first task-valid candidate under certificate-based selection, or at the first candidate with $q_t > 0$ under model-based selection.
For an evaluation set of $N$ instances, greedy depth shares a total budget of $6400N$ recursion steps: computation continues on the remaining active instances until this budget is exhausted or all instances have stopped. This matches the average compute budget of 6400-step depth scaling while allowing different instances to receive different amounts of computation. For GRAM, we report its performance with both LPRM and majority-vote selection. Results for each recursive-reasoning method are reported as the mean and standard deviation across three training seeds. For each task, ICS, Standard TRM, TRM with two depth-scaling, and PTRM use the same three trained TRM checkpoints for a fair comparison.

\textbf{Evaluation and inference cost.}
We measure solution accuracy by whether the returned answer $\hat{y}_t$ satisfies the original task rules, using the same correctness criterion for all methods. We evaluate all methods under two answer selection protocols. Under \emph{certificate-based selection}, a task certificate is available during inference, allowing us to measure recursive reasoning efficiency with the accuracy and computational cost of verified solution discovery.  Under \emph{model-based selection}, each method uses its own selection rule without the task certificate, evaluating the complete procedure's ability to return a correct answer. The task certificate is still used for final scoring under both protocols. We report inference cost as model FLOPs and execution time, averaged over all evaluated instances, including those that stop early or return incorrect answers. Detailed inference budgets, compute accounting, and profiling procedures are provided in Appendix~\ref{app:experimental-details}.

\textbf{LLM comparison.}
We additionally compare ICS without certificate-based selection against LLMs on the official PPBench golden set. All LLMs use a shared prompt template and a shared output limit of 128k tokens, including reasoning tokens and excluding input tokens. Attempts that fail to return a valid answer within this limit are counted as incorrect. The corresponding configuration details can be found in Appendix~\ref{app:llm-evaluation}.

\textbf{ICS achieves the best mean performance across all six tasks.}
As shown in Table~\ref{tab:performance}, under both certificate-based and model-based answer selection, ICS achieves the highest mean accuracy on all tasks and ties on Light-Up. With a task certificate available, ICS substantially improves verified solution discovery on all tasks. On Tapa, it reaches $95.9\%$, exceeding the second best method by $13.7$ percentage points. 
The gains persist under model-based selection. On Maze-Hard, ICS achieves $99.1\%$, compared with $91.8\%$ for the second best method. On Sudoku-Extreme, ICS achieves $99.7\%$ without certificate-based selection, exceeding EqR's $98.8\%$. Relative to PTRM's $98.3\%$ on the same trained TRM checkpoints, ICS reduces the remaining errors by approximately $82\%$.

\textbf{Test-time scaling.}
Figure~\ref{fig:scaling} compares test-time scaling on Sudoku-Extreme under model-based answer selection. At comparable compute, ICS substantially reduces the remaining errors: it achieves $99.70\%$ accuracy at $1.14$ TFLOPs per puzzle, reducing errors by approximately $92\%$ relative to the second best method (TRM + greedy depth), evaluated at approximately $1.20$ TFLOPs per puzzle.
Increasing the baselines' budgets does not close this gap.
None of the evaluated scaling baselines matches ICS accuracy
at its reported high-budget setting, despite using over
$100\times$ the test-time computation. Even providing these baselines with a task certificate at selection leaves their accuracy below model-selected ICS (Table~\ref{tab:performance}). 
Furthermore, we scale ICS along both the width (more hypotheses) and depth (more recursion steps) axes, achieving accuracies of $99.96\%$ and $99.91\%$ with 1.94 and 3.10 TFLOPs of test-time compute, respectively.
The two TRM depth-scaling curves also reveal a budget-dependent trade-off between computation allocation and answer selection. Greedy depth performs better at small budgets, consistent with early stopping concentrating computation on unresolved instances. At larger budgets, full-depth best-$q$ selection overtakes greedy stopping: continued evaluation allows the model to reconsider its answer after an early positive halting score, whereas greedy stopping commits to that first acceptance.

\textbf{The efficiency gains extend across all six tasks.}
Figure~\ref{fig:golden-coverage}(a,b) compares model computation and execution time across tasks under model-based answer selection. Across the six tasks, ICS achieves the results in Table~\ref{tab:performance} with approximately $10$--$525\times$ less test-time computation than the evaluated scaling baselines. ICS also achieves the lowest measured execution time per puzzle on all six tasks.

\textbf{ICS outperforms the evaluated LLMs on the PPBench golden set.}
Figure~\ref{fig:golden-coverage}c compares model-selected
ICS with each LLM's strongest evaluated configuration.
Without a task certificate at selection, ICS achieves
$91.7\%$ mean accuracy across three training seeds,
exceeding the strongest evaluated LLM result of
$78.3\%$ by $13.3$ percentage points. Per-task results and evaluation configurations are provided in Appendix~\ref{app:llm-evaluation}.

\textbf{Contributions of the input-conditioning operations.}
We ablate the two input-conditioning components on a random sampled subset of the Sudoku-Extreme test set (49,152 instances) using one trained TRM backbone (Table~\ref{tab:ablation-main}). Under model-based selection, removing hypothesis search reduces accuracy from $99.81\%$ to $98.82\%$, whereas removing validity-preserving restatements yields $99.75\%$. Certificate-based selection shows the same pattern: the corresponding accuracies are $99.94\%$, $98.83\%$, and $99.92\%$, respectively. Thus, at the evaluated settings, hypothesis search provides the larger contribution, while restatements offer an additional improvement. Full configurations and proposal ablations are reported in Appendix~\ref{app:ablation}.

\section{Discussions}
\label{sec:mechanisms}

\begin{figure}[t]
  \centering
  \setlength{\abovecaptionskip}{4pt}
  \includegraphics[width=\linewidth]{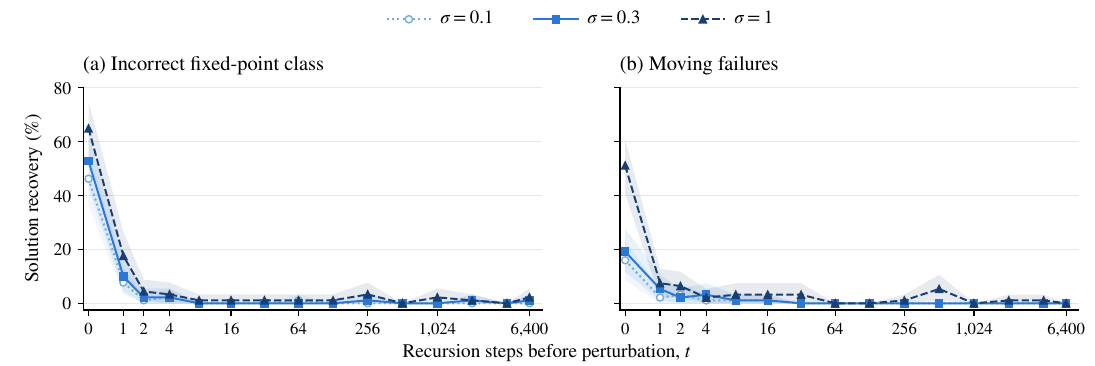}
  \caption{\textbf{Recovery depends on when the state is perturbed.} Shading shows pointwise 95\% instance-bootstrap intervals. The time axis uses $\log(1+t)$ spacing. $t=0$ denotes fresh initialization.}
  \label{fig:perturbation-time}
  % \vspace{-1.4em}
\end{figure}

The preceding experiments establish ICS's accuracy and computational advantages. We now examine two aspects of its design: the state from which reasoning begins and the input-conditioned
update map under which it proceeds. We first study latent-state interventions under the fixed original-input map $F_x$. We then examine hypothesis injection and validity-preserving restatements as two ways to change the input condition while keeping model parameters fixed. Together, these experiments investigate how input conditioning and fresh initialization affect access to correct solutions.

\subsection{Recovery depends on where exploration begins.}
\label{sec:recovery-initialization}
We first examine exploration under the original-input update map $F_x$. Repeated updates can leave the latent state almost unchanged even when the decoded answer is incorrect. Conversely, a correct answer can be produced while the latent state continues to change (Appendix~\ref{app:latent-trajectories}). These observations distinguish answer correctness from latent convergence.
We therefore examine whether interventions at different states along a reasoning trajectory can recover valid solutions while keeping the update map fixed.

\textbf{Perturbation direction and initialization both matter.}
We compare latent-state perturbations with hypothesis-conditioned rollouts, corresponding to the hypothesis-injection operation of ICS, on both TRM and GRAM backbones. For the same failed instances, warm-start perturbations modify a saved failed state and continue recursion under $F_x$. On TRM, we additionally evaluate cold-start perturbations applied to the fresh initialization $z_0$, followed by recursion under the same map.
Figure~\ref{fig:sigma} shows that cold-start random
perturbations recover substantially more instances than
warm-start random perturbations, which recover none across
the evaluated scales. Oracle perturbations can also
recover failures that resist random perturbations, indicating that recovery depends on direction as well as initialization, while both learned prior and posterior perturbations of GRAM yield substaintially lower recover rate. Both Oracle directions and the GRAM posterior use the ground-truth label and serve as diagnostic controls. ICS instead changes the input condition and recomputes from $z_0$ and achieves significantly higher candidate coverage than the tested label-free perturbations on both backbones, including cold-start random perturbations.

\textbf{Recovery becomes difficult early in failed trajectories.}
We next examine how perturbation effectiveness changes along the original-input trajectory.
On the selected Sudoku-Extreme failure cohorts, we apply Gaussian perturbations to $z_H$ at different recursion steps, holding the relative perturbation scale fixed. 
We measure whether any of 16 trials produces a correct answer within 64 subsequent updates under $F_x$. Across all three perturbation scales, recovery drops sharply after only two recursion steps for both the incorrect fixed-point class and moving failures
(Figure~\ref{fig:perturbation-time}).
For these instances, the tested perturbations become less effective early in the trajectory. 
This motivates exploring from fresh initialization, rather than restricting exploration to states reached after failure.

\subsection{Input conditioning changes recursive dynamics.}
\label{sec:input-conditioned-dynamics}
The preceding interventions change the latent state while continuing recursion under $F_x$. We now examine the complementary choice explored by ICS: changing the input condition that determines the recursive update map.

\begin{figure}[t]
  \centering
  \setlength{\abovecaptionskip}{4pt}
  \includegraphics[width=\linewidth]{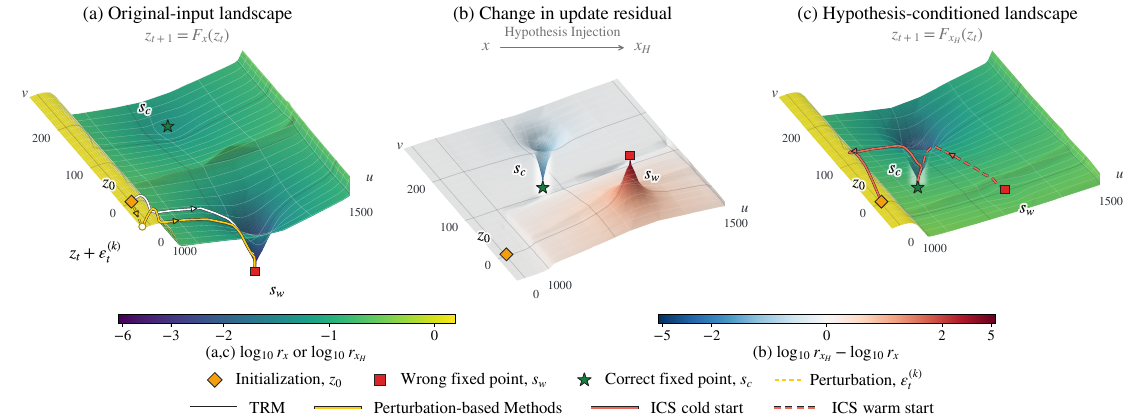}
  \caption{The landscapes show the relative one-step update residuals. Smaller residuals indicate less change in the latent state after one recursion step. 
}
\label{fig:input-conditioning}
% \vspace{-1.4em}
\end{figure}

\textbf{Hypothesis injection changes the update landscape.}
Figure~\ref{fig:input-conditioning} compares the relative update-residual landscapes induced by $x$ and $x_H$ on a shared two-dimensional section for an incorrect near-fixed point.
The surfaces interpolate model evaluations of $r_u(z)=\|F_u(z)-z\|_2/\|z\|_2$ for $u\in\{x,x_H\}$, and the overlaid paths are projections of recorded latent trajectories.
Under the landscape shaped by $F_x$, both the unperturbed TRM trajectory and the illustrated perturbed trajectory are captured by a wrong attractor and approach the same incorrect near-fixed point $s_w$.
At $s_w$, hypothesis injection increases the relative one-step update residual from approximately $2.4\times10^{-7}$ to $0.225$, so the original state is no longer nearly stationary under $F_{x_H}$.
Both the warm-start trajectory from $s_w$ and the cold-start trajectory from $z_0$ reach the same correct near-fixed point $s_c$, yielding a valid answer to the original task. 

\textbf{Input conditioning can eliminate incorrect attracting fixed points.}
Our theory and computer-assisted proofs establish that input conditioning can locally eliminate an incorrect attracting fixed point in a frozen recursive reasoner. 
In this mechanism, continuous variation of the input embedding causes an incorrect attracting fixed point to collide with an unstable fixed point, and the pair disappears through a local fold (saddle-node) bifurcation~\citep{kuznetsov2004bifurcation}.
We derive sufficient conditions for this mechanism and verify them with rigorous numerical certificates in separately trained $4 \times 4$ Sudoku reasoners (Appendix~\ref{app:fold-theory} and~\ref{app:small-model-certificate}).
Numerical continuation in the Maze-Hard and Sudoku-Extreme models provides complementary evidence of folds in incorrect fixed-point branches (Appendix~\ref{app:fixed-point-continuation}). 
These results establish that changing the input can remove an incorrect attracting state from the reasoner's dynamics, providing a mechanism for overcoming persistent failures without retraining.

\textbf{Fresh initialization can enable recovery.}
In the extended Sudoku study, restarting from $z_0$ recovers solutions under input conditions for which continuation from the saved failed state remains incorrect. The retained incorrect near-fixed states include cases with estimated local spectral radii below one. Recovery also depends on whether the hypothesis agrees with the reference solution and corrects an erroneous prediction (Appendix~\ref{app:hypothesis-interventions}). 
These findings motivate combining input conditioning with recomputation from a fresh latent state.

% \emph{Fixed-point continuation distinguishes displacement from disappearance.}
% However, the warm-start recovery is not universal. Extended Sudoku study shows that wrong near-fixed states can persist under hypothesis conditioning, including states with estimated local spectral radii below one, while restarting from $z_0$ enables recovery.
% Recovery depends on initialization and on whether the hypothesis agrees with the reference solution and corrects an erroneous prediction (Appendix~\ref{app:hypothesis-interventions}). 

\textbf{Validity-preserving restatements also change recursive dynamics.}
After aligning latent coordinates across equivalent task representations, we find that restatements alter recursive updates and can redirect trajectories from the same initialization (Appendix~\ref{app:restatement-dynamics})

Together, these findings show that input conditioning can redirect reasoning trajectories and locally eliminate incorrect attracting fixed points.
They motivate shaping recursive dynamics as an additional dimension of test-time scaling. ICS realizes this principle by searching over input conditions that induce alternative update maps, using hypothesis injection and validity-preserving restatements. Combined with fresh initialization, this search enables efficient test-time scaling without additional model training, as demonstrated by our accuracy and inference-cost results.

\section{Related Work}
\label{sec:related-work}

\paragraph{Recursive and looped reasoning.}
Reusing a learned transformation to refine a latent state has a long history, from Universal Transformers~\citep{dehghani2019universal} and adaptive computation time~\citep{graves2016adaptive,banino2021pondernet} to recurrent networks that extrapolate to harder instances by iterating longer~\citep{schwarzschild2021can,bansal2022endtoend} and looped or recurrent-depth transformers that trade parameters for depth~\citep{giannou2023looped,saunshi2025reasoning,geiping2025scaling}, now pretrained at scale with learned depth allocation~\citep{zhu2025ouro}.
Iterative neural solvers have also been developed for constraint problems such as Sudoku, including recurrent relational networks~\citep{palm2018recurrent}, differentiable satisfiability layers~\citep{wang2019satnet}, energy-based iterative reasoning~\citep{du2022irem,du2024ired} and the recursive reasoners
HRM~\citep{hrm2025} 
% couples high- and low-level recurrent modules operating at different timescales, 
and TRM~\citep{trm2025}
% simplifies this design with a single shared network that refines a latent reasoning state and an embedded answer
. Both models re-inject the problem input at every
recursion step, as does the recall architecture of \citet{bansal2022endtoend}, which keeps the
problem instance from being forgotten over many iterations.
% \citet{bansal2022endtoend} showed that re-injecting the problem input at every iteration prevents degeneration over long rollouts; ICS exploits the complementary observation that \emph{changing} this input changes the recursive dynamics themselves.
ICS instead changes this input and each input condition induces a different update map under the same
frozen model.
% We build on trained TRM reasoners and organize their test-time computation through input-conditioned search, keeping parameters frozen.

\paragraph{Test-time scaling and answer selection.}
For language models, %additional inference compute is commonly spent on sampling many candidates and selecting among them by majority vote~\citep{selfconsistency2023}, learned outcome or process verifiers~\citep{cobbe2021training,lightman2024lets}, or structured search over intermediate steps~\citep{yao2023tree}.%
test-time compute is typically spent on sampling many candidates and selecting among them ~\citep{selfconsistency2023,cobbe2021training,lightman2024lets,yao2023tree}.
Repeated sampling could increase coverage, 
% which translates into accuracy when an automatic verifier is available but is much harder to exploit without one~\citep{brown2024large}, and the best allocation of test-time compute depends on problem difficulty~\citep{snell2025scaling}.
but converting coverage into accuracy is much harder without an automatic verifier ~\citep{brown2024large}.
% Our certificate-based and model-based protocols mirror this distinction between discovering and selecting correct answers.
% For recursive reasoners, test-time scaling has so far been realized by deeper recursion and by sampling latent trajectories: PTRM~\citep{ptrm2026} injects Gaussian noise during recursion and selects with the halting head, and GRAM~\citep{gram2026} learns stochastic latent transitions via variational inference and selects by majority vote or a latent process reward model.
For recursive reasoners, PTRM~\citep{ptrm2026} diversifies candidates by injecting Gaussian noise into the latent state during recursion, and GRAM~\citep{gram2026} by sampling from learned stochastic latent transitions.
% ICS instead diversifies candidates by varying the input condition, which our experiments show to be substantially more compute-efficient than latent perturbation.
ICS instead changes the update map; this recovers more solutions than label-free latent perturbations on the same failed instances, and ICS uses less compute overall\ref{fig:golden-coverage}.

\paragraph{Equilibrium and attractor views of reasoning.}
% Deep equilibrium models define outputs as fixed points of a learned map~\citep{bai2019deep}, and associative-memory models retrieve patterns as attractors of an energy landscape~\citep{hopfield1982neural,ramsauer2021hopfield}.
% \citet{anil2022path} argue that equilibrium models exploit test-time computation best when they are \emph{path independent}, converging to the same attractor regardless of initialization.
Equilibrium models compute their outputs as fixed points of a learned map~\citep{bai2019deep}. \citet{anil2022path} call a recurrent computation path independent if it converges to the same limiting behavior from any initial state, and find that equilibrium models with this property better exploit test-time computation. Two recent recursive reasoners build these ideas into training: Attractor Models ~\citep{attractor2026} solve for the fixed point of TRM's recursive update instead of unrolling it; EqR~\citep{eqr2026} randomizes the initial state to encourage path independence and restarts from multiple random initial states under the same update map at test time. However, convergence alone does not ensure task correctness(Section~\ref{sec:mechanisms}).
ICS explores a complementary direction by searching over input-conditioned update maps with a frozen model, without requiring path independence. It restarts every candidate from $z_0$, allocating test-time computation to trajectories generated under alternative update maps.
% ICS instead uses a frozen standard TRM, which is trained from a single fixed initial state and need not be path independent. It therefore restarts every candidate from the same $z_0$ and varies the map rather than the initial state.

% Attractor Models~\citep{attractor2026} refine backbone predictions with a fixed-point solver trained via approximate implicit gradients, and EqR~\citep{eqr2026} trains with randomized initialization and path noise to encourage solution-aligned attractors, then scales inference through longer rollouts and stochastic restarts with convergence-based selection.
% These works highlight the importance of aligning latent convergence with task correctness.
% Our diagnostics show that trained recursive reasoners can instead converge to incorrect near-fixed points, produce correct answers on moving trajectories, and recover from failures only from some starting states.
% Rather than reshaping attractors during training, ICS changes the update map at test time by conditioning on hypotheses and restarting from a fresh state.

\paragraph{Search over partial assignments and equivalent views.}
Branching on tentative variable assignments is the backbone of classical constraint solving, from DPLL~\citep{davis1962machine} to conflict-driven clause learning~\citep{marquessilva1999grasp}, and learned models have been used to choose the branching variable
% guide such search
~\citep{selsam2019learning}.
Evaluating a model on symmetry-transformed inputs is another source of candidate diversity, used by TRM on ARC-AGI~\citep{trm2025}, aggregated across views as a product of experts~\citep{franzen2025product}, and combined with parameter updates in test-time training~\citep{akyurek2025surprising}.
ICS combines both ideas while keeping parameters frozen: the model's own decoder margins select where and how to branch, the reasoner itself, rather than a hand-designed propagator, completes the solution under each hypothesis without a task certificate, and validity-preserving restatements provide additional views, although our ablation identifies hypotheses as the larger source of improvement.

\section{Conclusion}
\label{sec:conlusion}

We introduced Input-Conditioned Search (ICS), which allocates test-time computation to exploring alternative update maps as an additional scaling axis without retraining the model. 
% Our instantiation combines hypothesis injection and validity-preserving restatements, recomputing candidates from a fresh latent state under each condition.
Across six reasoning tasks, ICS achieves the highest mean accuracy while using substantially less model computation and execution time than baselines.
Controlled experiments show that ICS can effectively redirect recursive updates and trajectories.
% latent convergence alone does not guarantee correctness, that both input-conditioning operations can redirect recursive updates and trajectories, and that recovery depends on the starting state. 
% These findings motivate searching over input conditions
% together with fresh initialization.
Our theoretical analysis and rigorous numerical verification establish that ICS can locally eliminate incorrect attracting fixed points. 
Together, these findings demonstrate efficient test-time scaling through ICS without additional training and motivate further research on shaping recursive reasoning dynamics during training and at test time.
% More broadly, ICS establishes input conditioning as an effective dimension of test-time search and opens opportunities to discover or learn better conditions and search strategies for shaping recursive reasoning dynamics.

% \section{Conclusion}
% \label{sec:conlusion}

% We introduced Hypothesis-Injection Search (HIS), which searches over input conditions and recomputes candidates with a frozen recursive reasoner.
% Across six reasoning tasks, HIS achieves the highest mean
% accuracy, including ties, under both certificate-based and
% model-based answer selection, while using substantially less
% model computation and execution time than the evaluated
% scaling baselines.
% Our mechanism experiments show that latent convergence alone
% does not guarantee correctness, and that recovery depends on
% both the update map and the starting state.
% HIS acts on these two factors through input conditioning and fresh initialization. More broadly, these findings open new opportunities for efficient test-time scaling by shaping the update rules of recursive reasoning, motivating future methods that learn to modify these rules for more effective solution discovery.
% \input{ICS_sections/statement}
\section*{Acknowledgments}
This work is supported by the Canada NSERC Discovery Grant (RGPIN-2021-03115) and the Amazon Research Award (ARA) program.

\clearpage
\bibliography{references}
\bibliographystyle{plainnat}
% \section{Appendix}
\clearpage
\appendix
\section{Experimental Details}
\label{app:experimental-details}

\subsection{Datasets and Splits}
\label{datasets}

We evaluate six reasoning tasks: Sudoku-Extreme and
Maze-Hard~\citep{hrm2025}, and Light-Up, Nurikabe, Tapa,
and Heyawake from PPBench~\citep{ppbench2026}.
All methods are trained and evaluated on the same training and test partitions of every task with 3 seeds.

\paragraph{Sudoku-Extreme and Maze-Hard.}
Each Sudoku-Extreme puzzle consists of $9\times9$ grids.
A valid answer must preserve the given clues and satisfy the row, column, and subgrid constraints.
We use the official split of the Sudoku-Extreme dataset, with $1{,}000$ training instances and a test pool of $422{,}786$ instances.
Each Maze-Hard puzzle contains $30\times30$ grids and requires finding a shortest path between the designated endpoints. We also use the official split of the dataset, with training and test sets each contain $1{,}000$ instances.

\paragraph{PPBench test-set construction.}
For each of the four PPBench tasks, we reserve the $15$ instances belonging to the official \texttt{golden\_300} selection. With only $15$ instances per task, a single prediction changes per-seed accuracy by $6.67$ percentage points, making comparisons sensitive to individual instances. To obtain a more informative comparison, we additionally sample $100$ instances per task from the non-golden pool using the fixed random seed \texttt{20260808}. For Tapa, we sample $50$ additional instances because fewer instances are available. The golden and additionally sampled instances together form the held-out test set for each task.

\paragraph{PPBench training data and input representation.}
After removing the held-out instances, we use all remaining instances with $\max(H,W)\leq24$ for training, where $H$ and $W$ denote the puzzle dimensions. We represent each puzzle on a $26\times26$ canvas. The main six-task comparison uses this representation consistently within each PPBench task. Two Light-Up golden instances do not fit this canvas and are excluded from that evaluation. The resulting test sets contain $113$ Light-Up, $115$ Nurikabe, $65$ Tapa, and $115$ Heyawake instances. To make an additional evaluation on the whole Light-Up golden set, we further trained 3 TRM seeds on the data that fits in a $40\times 40$ canvas. The results on the whole Light-Up golden set are consistent with the universal protocol (trained on $26 \times 26$, with all three TRM seeds achieving $100\%$ accuracy on the full Light-Up golden set. All scaling methods based on the same TRM seeds also achieve $100\%$ accuracy on the full Light-Up golden set.

\paragraph{Complete golden-set evaluation.}
For the comparison with LLMs, we evaluate all $15$ golden instances from each task, giving $60$ instances in total. To accommodate the two larger Light-Up instances, this comparison uses separately trained $40\times40$ Light-Up models with three training seeds. The other three tasks retain their $26\times26$ configurations. The main six-task table continues to report the $26\times26$ Light-Up results to maintain consistency with other PPbench tasks.

\subsection{Models and Training}
\label{app:models-training}

\paragraph{Training seeds and shared backbones.}
For each task, standard TRM, TRM + depth scaling, TRM + greedy depth, PTRM, and ICS use the same three trained TRM checkpoints. This allows us to compare their test-time procedures on a common trained reasoner. GRAM, EqR, and Attractor Models are trained separately using their respective architectures and training procedures, with the same task-specific data splits. We report solution accuracy as the mean and standard deviation across three training seeds. Model parameters remain frozen during inference. Following PTRM~\citep{ptrm2026}, we augment PPBench training inputs with partial puzzle states taken from trajectory prefixes. 

\subsection{Maze-Unique and the choice of maze benchmark}
\label{app:maze-unique}

We additionally evaluate Maze-Unique~\citep{eqr2026} to assess the remaining accuracy headroom for test-time scaling. We train three TRM models on its training split using the official TRM training configuration.

Before evaluating task validity, we restore the original givens in each prediction by copying the wall, start, and goal tokens from the input. This input-only projection removes errors in reproducing these given tokens without using a reference solution or a task certificate for answer selection.

The three standard TRM models achieve rule-valid accuracies of $100.0\%$, $100.0\%$, and $99.8\%$. ICS and PTRM each achieve $100.0\%$ on all three trained backbones. These results leave little headroom for distinguishing accuracy improvements from additional test-time computation in this setting.

We therefore use Maze-Hard for the main maze experiments, where the remaining errors permit a more informative comparison of test-time scaling methods.

\subsection{Inference Protocols}
\label{app:inference-protocols}

\paragraph{Standard TRM.}
Standard TRM stops at the first recursion step with $q_t>0$, or at its standard depth limit if the threshold is never reached, and returns the corresponding prediction $\hat{y}_t$. We evaluate this same output under both selection regimes, without additional trajectories or retries.

\paragraph{TRM + depth scaling.}
For each instance, we execute the full depth budget of $D$ recursion steps, regardless of intermediate halting scores, and consider the decoded candidates produced throughout the trajectory. Under model-based selection, we return the candidate with the highest halting score:
\begin{equation}
t^\star = \operatorname*{arg\,max}_{1\leq t\leq D} q_t,
\qquad
\hat y = g_\theta(z_{t^\star};x).
\end{equation}
Under certificate-based selection, we return a valid candidate whenever one occurs along the trajectory. Both regimes execute the full depth budget.
Our main comparison uses $D=6{,}400$ to match the budget of PTRM and GRAM.

\paragraph{TRM + greedy depth.}
Greedy depth stops each instance at the first task-valid candidate under certificate-based selection, or at the first candidate with $q_t>0$ under model-based selection. The corresponding prediction is returned immediately. For an evaluation set of $N$ instances, we allocate a shared budget of $6{,}400N$ recursion steps. Computation continues on instances that have not yet stopped until the shared budget is exhausted or all
instances have stopped. Thus, individual instances may receive more than $6{,}400$ steps, while the total budget matches that of fixed-depth scaling over the same evaluation set. Under model-based selection, instances that have not triggered $q_t>0$ when the shared budget is exhausted return the prediction decoded at their final recursion step. In our experiments, several instances remain unsolved with more than 2.9 million recursion steps.

\paragraph{PTRM.}
PTRM~\citep{ptrm2026} generates $K$ trajectories from fresh initial states using the frozen TRM backbone. Each trajectory runs for $D$ recursion steps with early halting disabled. Noisy trajectories receive Gaussian perturbations with standard deviation $\sigma$ to the low-level recurrent state $z_L$ during recursion. We follow the hyperparameter settings reported in their paper. Each trajectory contributes its final decoded answer and corresponding halting score. Model-based selection returns the terminal candidate with the highest halting score. Certificate-based selection returns a valid answer whenever one exists among these terminal candidates. Both regimes evaluate the full pool, requiring $KD=6{,}400$ recursion steps per instance.

\paragraph{GRAM.}
GRAM~\citep{gram2026} generates $K=100$ stochastic trajectories using its learned prior, each evaluated for $D=64$ recursion steps with early halting disabled. Each trajectory contributes its terminal prediction and learned LPRM score. We evaluate two model-based selection rules on this candidate pool. LPRM selection returns the terminal candidate with the highest LPRM score. Majority selection returns the most frequent complete decoded answer, with ties resolved by first occurrence in sampling order. Certificate-based selection returns a valid answer whenever one exists in the same terminal candidate pool. All selection rules use the full configured pool.

\paragraph{EqR and Attractor Models.}
We use the authors' officially released implementations of EqR~\citep{eqr2026} and Attractor Models~\citep{attractor2026}. For Attractor Models, we identified and corrected a history-window indexing error in the Anderson solver of the released implementation. This error caused non-finite training losses in our runs. All reported Attractor Models checkpoints were trained and evaluated using the corrected solver.
We will release the correction together with our ICS implementation and trained checkpoints to facilitate reproduction of our results by the community.

\subsection{Matched-subset comparison on Sudoku-Extreme}
\label{app:sudoku-subset}

Evaluating the high-budget search configuration of Attractor
Models on the full Sudoku-Extreme test set is computationally expensive. We therefore evaluate this configuration on a fixed subset comprising the first $3{,}072$ instances in test-loader order. To enable a direct comparison, we also report all other methods on exactly the same instances.

Table~\ref{tab:sudoku-subset} reports accuracy under both certificate-based and model-based answer selection, summarized as the mean and sample standard deviation across three training seeds. Each method retains its corresponding inference budget.

ICS achieves the highest mean accuracy under both protocols on this shared subset: $99.95\%$ with certificate-based selection and $99.65\%$ with model-based selection, compared with the strongest baseline means of $97.48\%$ and $97.49\%$, respectively. Thus, its accuracy advantage also holds when all methods are evaluated on the same test instances.

\begin{table}[t]
\centering
\small
\begin{tabular}{lcc}
\toprule
Method & Certificate selection & Model selection\\
\midrule
Standard TRM & $84.80\pm1.18$ & $84.80\pm1.18$\\
TRM + depth scaling$^{\dagger}$ & $97.48\pm0.63$ & $97.49\pm0.75$\\
PTRM & $97.47\pm0.96$ & $97.47\pm0.96$\\
GRAM, LPRM & $94.99\pm0.30$ & $94.69\pm0.52$\\
GRAM, majority & $94.99\pm0.30$ & $94.09\pm0.56$\\
EqR & $97.48\pm0.16$ & $97.46\pm0.17$\\
Attractor Models & $94.76\pm1.30$ & $94.76\pm1.30$\\
\textbf{ICS} & $\mathbf{99.95\pm0.07}$ & $\mathbf{99.65\pm0.27}$\\
\bottomrule
\end{tabular}
\caption{Solution accuracy (\%) on the same first $3{,}072$ Sudoku-Extreme test instances. Values are means and sample standard deviations across three training seeds. Each method retains its corresponding inference budget; this is not an equal-compute comparison. Standard TRM evaluates the same native 16-step output in both columns. The two GRAM rows share the same certificate coverage. $\dagger$: depth scaling uses a $6{,}400$-step budget. Its two columns come from separate archived numerical executions with different batch configurations, rather than two selectors applied to an identical trajectory pool.}
\label{tab:sudoku-subset}
\end{table}

\subsection{ICS Configuration}
\label{app:his-configuration}

\paragraph{Task-specific hypotheses.}
ICS constructs hypotheses over task-specific cell states while preserving the original clues and structural
constraints. Table~\ref{tab:his-hypotheses} summarizes the assignment tokens used for each task. Each extension assigns a cell not already assigned by the current hypothesis, using an admissible state token that differs from the current prediction.

\begin{table}[t]
\centering
\small
\begin{tabular}{ll}
\toprule
Task & Hypothesis assignment \\
\midrule
Sudoku-Extreme & A digit from $1$ to $9$ \\
Maze-Hard      & A traversable cell belongs to the path \\
Light-Up       & A lamp is placed in an empty cell \\
Nurikabe       & A cell is shaded or unshaded \\
Tapa           & A cell is shaded or unshaded \\
Heyawake       & A cell is shaded or unshaded \\
\bottomrule
\end{tabular}
\caption{Task-specific assignments used for ICS.}
\label{tab:his-hypotheses}
\end{table}

For Maze-Hard, hypotheses mark path membership while preserving walls, the start, and the goal. For Heyawake, cell states are encoded together with clues and room boundaries; hypothesis injection changes only the shading component of the token and preserves the remaining information.

\paragraph{Validity-preserving restatements.}
For Sudoku-Extreme, we compose digit relabeling, permutations of bands and stacks, row permutations within bands, column permutations within stacks,
and transposition. For Maze-Hard and the PPBench tasks, we use grid rotations and reflections. All task annotations are transformed consistently with the grid. For tasks represented on a padded canvas, we transform the puzzle region and embed the transformed puzzle back into the canvas. Decoded predictions are mapped back to the original coordinates, with inverse digit relabeling applied where necessary.

\paragraph{Search parameters.}
Table~\ref{tab:his-search-settings} lists the task-specific search settings. Every hypothesis-conditioned rollout starts from a fresh initialization $z_0$ and runs for 16 recursion steps. Each rollout on a validity-preserving restatement also uses 16 steps.

The proposal limit bounds the number of cell--value assignments proposed per parent, while the node budget bounds the total number of hypothesis children across all search levels. We use the margin-based proposer and consider up to two admissible alternative values per position, ordered by the decoder's token scores.

\begin{table}[t]
\centering
\small
\setlength{\tabcolsep}{4pt}
\resizebox{\linewidth}{!}{%
\begin{tabular}{lccccccc}
\toprule
Task & Original-input depths & Levels & Beam &
Proposals & Nodes & Views & Selection \\
\midrule
\multicolumn{8}{l}{\emph{Certificate-based selection}} \\
Sudoku-Extreme
& 20 & 6 & 4 & 48 & 96 & 32 & Certificate \\
Maze-Hard
& 20 & 6 & 4 & 48 & 96 & 8 & Certificate \\
All four PPBench tasks
& 20 & 6 & 4 & 48 & 96 & 8 & Certificate \\
\midrule
\multicolumn{8}{l}{\emph{Model-based selection}} \\
Sudoku-Extreme
& 20 & 8 & 3 & 6 & 24 & 32 & $q$ threshold \\
Maze-Hard
& 20 & 8 & 3 & 6 & 24 & 8 & $q$ threshold\\
Light-Up, Nurikabe
& 20 & 8 & 3 & 6 & 24 & 8 & $q$ threshold \\
Tapa, Heyawake
& 20 & 8 & 3 & 6 & 24 & 8 & Confirmation \\
\bottomrule
\end{tabular}}
\caption{ICS search configurations. Levels, beam width, proposals per parent, and evaluated hypothesis nodes specify upper limits. Views denotes the size of the final restatement set, including the identity representation. }
\label{tab:his-search-settings}
\end{table}

\paragraph{Certificate-based selection.}
A candidate is accepted when it satisfies the original task rules, after mapping it back to the original representation and restoring the original givens. Accepted instances leave the search.

\paragraph{$q$ threshold selection.}
For $q$ threshold selection, we accept a candidate when its halting score exceeds zero. This condition is checked at the end of each configured rollout.

\paragraph{Confirmation-based selection.}
For confirmation-based selection, we maintain the candidate with the highest halting score encountered so far. 

The selection method on each task is chosen by minimizing the error on the training set.

\subsection{Inference Cost Measurement}
\label{app:inference-cost}

\paragraph{Average inference cost.}
We report inference cost per evaluated instance. The denominator includes all evaluated instances, including those that stop early or return incorrect answers. Computation spent on unsuccessful hypotheses and rejected candidates contributes to the total cost. 

\paragraph{Model FLOPs.}
For each model and task, we record the number of executed forward calls at each batch shape.
We measure the FLOPs of these calls using PyTorch's \texttt{FlopCounterMode} on eager replays of inputs saved during execution, counting a multiply-add as two operations. For each shape, we average the measurements from up to two saved calls.

Let $n_s$ denote the number of executed calls with shape $s$, and let $f_s$ denote their profiled FLOPs per call. For $N$ evaluated instances, the reported computation is
\begin{equation}
    \overline{C}_{\mathrm{FLOPs}}
    =
    \frac{1}{N}\sum_s n_s f_s.
\end{equation}
The count covers the model operations supported by the profiler, primarily matrix multiplications.

\paragraph{Execution time.}
We measure elapsed execution time on NVIDIA L40S GPUs. All methods are measured end-to-end on the same data. Per-instance execution time is calculated by dividing the total execution time by the number of instances.

\paragraph{Used datasets and configurations.}
For the computational comparison, we evaluate a fixed subset of 3,072 Sudoku-Extreme instances and the complete test pools of the other tasks: 1,000 Maze-Hard, 113 Light-Up, 115 Nurikabe, 65 Tapa, and 115 Heyawake instances.

\subsection{LLM Evaluation Protocol}
\label{app:llm-evaluation}

\paragraph{Evaluation set.}
We evaluate LLMs on the official PPBench golden set, using the 15 instances selected for each of Light-Up, Nurikabe, Tapa, and Heyawake, for a total of 60 instances.

For this comparison, ICS uses the Light-Up models trained on a $40\times40$ canvas, which accommodates all 15 Light-Up golden instances. The other three tasks retain their $26\times26$ configurations. We report ICS accuracy averaged across three training seeds.

\paragraph{Prompt and inference budget.}
All LLMs receive the same prompt template, instantiated with the task rules, puzzle dimensions, and puzzle input. Each model makes a single attempt per instance under each evaluated configuration, with no retries. The output-token limit is $128$k tokens, including reasoning tokens and excluding input tokens. Responses that do not produce a valid answer within this limit are counted as incorrect.

\begin{quote}
\small
\begin{verbatim}
You are solving a {puzzle_name} puzzle.

Rules:
{puzzle_rules}

The puzzle has {H} rows and {W} columns.
Rows are numbered 1-{H} top to bottom,
columns 1-{W} left to right.

Puzzle:
{puzzle_to_solve}

Think carefully, then output your final answer as
exactly {H} lines of {W} characters each (no spaces)
inside a ```-fenced code block. Clue cells are part
of the grid and are written as . in the answer.
The last fenced block in your reply will be graded.
\end{verbatim}
\end{quote}

\paragraph{Answer extraction and scoring.}
We extract the answer from the last fenced code block and evaluate it against the original puzzle rules. A response is correct only if the extracted answer has the required format and satisfies the task constraints. Valid alternative solutions are accepted. Missing, malformed, and incorrect answers are all counted as incorrect.

Accuracy is the percentage of the 60 instances solved correctly. ICS uses model-based answer selection without access to the task certificate during selection; the certificate is used to score the returned answer.

\paragraph{Reported model configurations.}
For an LLM evaluated at multiple reasoning-effort settings, the main comparison reports the setting with the highest
aggregate accuracy on the golden set. Each setting is evaluated with one attempt per instance. Answers from different settings are not pooled to construct a per-instance best-of-settings result.

\paragraph{Per-task results.}
Table~\ref{tab:llm-golden-results} reports the number of correctly solved instances for each task. The main figure retains the strongest evaluated effort setting for each LLM. For ICS with model-based selection, the three training seeds solve 57, 54, and 54 instances, respectively, giving an overall accuracy of $91.7\pm2.9\%$.

\begin{table}[t]
\centering
\small
% \setlength{\tabcolsep}{5pt}
% \resizebox{\linewidth}{!}{%
\begin{tabular}{lrrrrr}
\toprule
Model / configuration
& Light-Up & Nurikabe & Tapa & Heyawake & Overall (\%) \\
\midrule
gpt-5.6-sol, xhigh
& 13 & 9 & 13 & 12 & 78.3 \\
gpt-5.6-sol, high
& 12 & 7 & 11 & 11 & 68.3 \\
gpt-5.6-sol, medium
& 9 & 2 & 6 & 3 & 33.3 \\
Claude Opus 5, high
& 11 & 11 & 14 & 9 & 75.0 \\
Kimi K3, max
& 13 & 4 & 10 & 4 & 51.7 \\
Qwen3.8 Max, max
& 13 & 5 & 8 & 2 & 46.7 \\
Claude Sonnet 5, max
& 12 & 2 & 9 & 3 & 43.3 \\
GLM-5.2, max
& 9 & 2 & 8 & 2 & 35.0 \\
Qwen3.8-27B (FP8)
& 10 & 2 & 3 & 3 & 30.0 \\
\midrule
ICS, training seed 0
& 15 & 15 & 13 & 14 & 95.0 \\
ICS, training seed 1
& 15 & 12 & 14 & 13 & 90.0 \\
ICS, training seed 2
& 15 & 13 & 13 & 13 & 90.0 \\
\bottomrule
\end{tabular}
% }
\caption{Results on the PPBench golden set. Task columns give correctly solved counts out of 15 instances; overall accuracy uses all 60 instances. All rows use answer selection without the task certificate. ICS rows correspond to separate training seeds.}
\label{tab:llm-golden-results}
\end{table}

\section{Latent Trajectory Diagnostics}
\label{app:latent-trajectories}

We examine latent trajectories to distinguish convergence of the recursive state from correctness of the decoded answer. The representative examples include both correct and incorrect near-stationary states, as well as correct answers produced on trajectories that continue moving.

\paragraph{Trajectory recording.}
We keep the model parameters and original input fixed, disable halting, and repeatedly apply $F_x$ from the
model's standard initialization $z_0$. For TRM, the recorded state includes both recurrent components, $z=(z_H,z_L)$. These diagnostic rollouts use FP32 forward computation and recurrent states.

\paragraph{State-update residual.}
We measure the relative change between consecutive latent states as
\begin{equation}
    r_t
    =
    \frac{\|z_{t+1}-z_t\|_2}{\|z_t\|_2}.
\label{eq:trajectory-residual}
\end{equation}
The norms are computed over the complete latent state. We also record the absolute update norm, state norm, decoded prediction, and halting score at each step. Small residuals indicate approximate stationarity; answer correctness is evaluated separately.

\paragraph{Two-dimensional visualization.}
We save a fixed 256-dimensional random projection of each latent trajectory. For visualization, we fit a two-dimensional principal component analysis (PCA) basis to the final 256 saved projections and apply that basis to the trajectory. Colored paths show this final window, and gray paths show the earlier approach where displayed. Black markers indicate the final recorded state.

Each example uses its own PCA basis and axis scales. The accompanying diagnostics show the step residuals, their distribution over the final 256 updates, and a return map of consecutive values of a fixed linear observable of the latent state.

\paragraph{Endpoint correctness.}
We evaluate each saved final prediction against the original task rules after restoring the original givens, and separately check whether its raw decode respects those givens. The correct/incorrect labels in the trajectory figures refer to these saved endpoints. For a moving trajectory, a correct endpoint does not imply that every preceding prediction was correct.

The examples are selected to illustrate distinct combinations of latent behavior and answer correctness. They show that an incorrect answer can persist near a stationary state, while a correct answer can occur during continued latent motion. Their two-dimensional geometry provides a visualization of the recorded trajectories; numerical classification uses the diagnostics described below.

\paragraph{Near-fixed points and periodic trajectories.}
We identify approximate stationarity and periodic returns using numerical residuals. For the trajectory examples shown here, we use a relative tolerance
$\tau_{\mathrm{fp}}=2.5\times10^{-6}$, ten times the FP32 residual floor used in these diagnostics.

Let $T$ denote the observation horizon. Starting from the final state $z_T$, we search for the smallest return period $p\in\{1,\ldots,64\}$ satisfying
\begin{equation}
    \frac{\|F_x^p(z_T)-z_T\|_2}{\|z_T\|_2}
    \leq \tau_{\mathrm{fp}},
\label{eq:period-return}
\end{equation}
where $F_x^p$ denotes $p$ successive applications of $F_x$. This endpoint return test uses the full latent state.

We additionally require a window of 16 consecutive $p$-step relative differences below the same tolerance within the recorded trajectory. For $p=1$, these are the full-state update residuals $r_t$. For $p>1$, we compute the persistence test from the saved random projections $u_t=Rz_t$, using
$\|u_{t+p}-u_t\|_2/\|u_t\|_2$.

A trajectory satisfying both tests is assigned to the near-fixed-point class when $p=1$, or the approximately periodic class when $p>1$. These labels describe numerical behavior over the observed horizon.

\begin{figure}[t]
\centering
\includegraphics[width=\linewidth]
{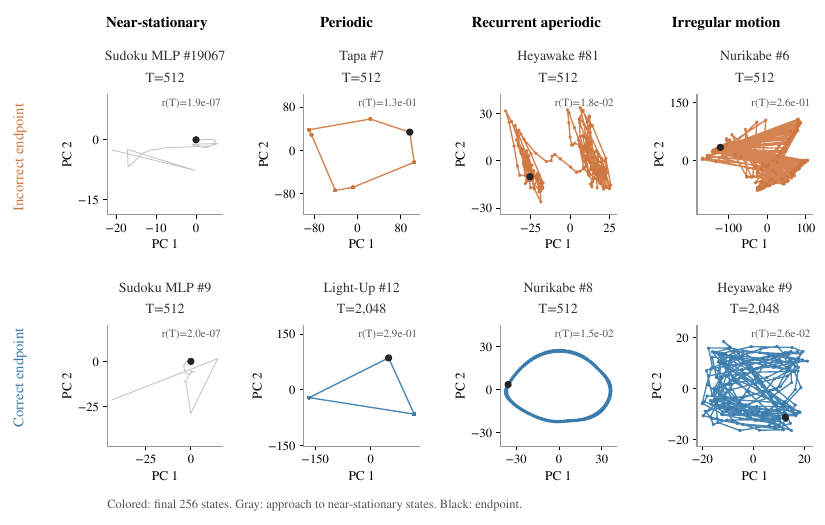}
\caption{
\textbf{Different latent behaviors can accompany either
correct or incorrect answers.}
Top: incorrect endpoints. Bottom: correct endpoints.
Columns show near-stationary, periodic, regular moving,
and irregular moving examples.
Each panel uses an independent PCA basis fitted to the
final 256 saved random projections.
Colored paths show this window; gray paths show the
earlier approach for the near-stationary examples.
Black markers indicate endpoints, and $r(T)$ reports
the final recorded relative update residual.
Axis scales differ across panels.
}
\label{fig:latent-overview}
\end{figure}

\begin{figure}[p]
\centering
\includegraphics[width=\linewidth]
{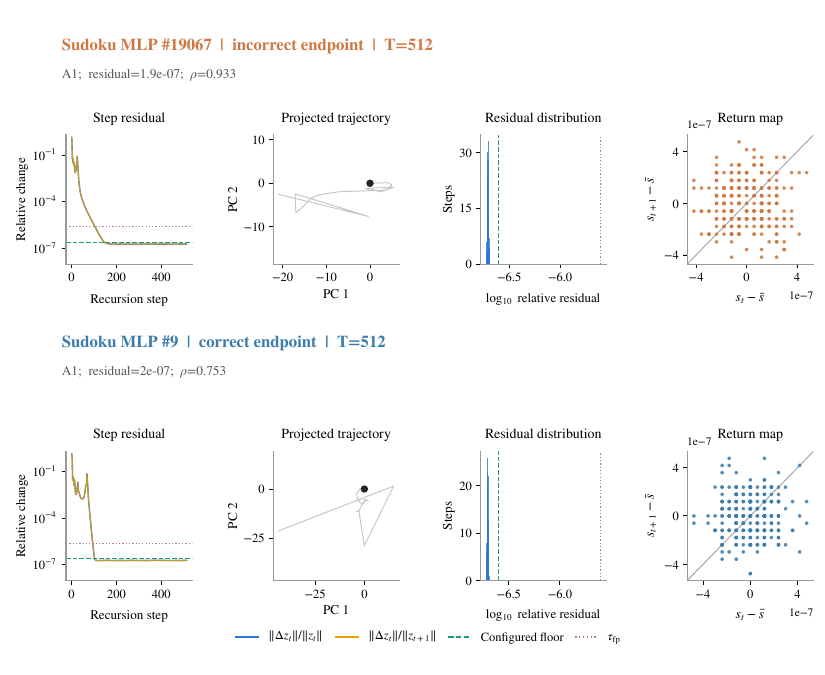}
\caption{
\textbf{Near-fixed points with different answer outcomes.}
Both Sudoku MLP trajectories approach numerical
stationarity within 512 updates.
The incorrect raw prediction satisfies the row, column,
and box constraints but violates an original given;
restoring that given produces an answer that fails the
task rules. The other endpoint is correct.
Columns show update residuals, the projected trajectory,
the final-window residual distribution, and a centered
scalar return map.
The two residual curves normalize the update norm by
$\|z_t\|_2$ and $\|z_{t+1}\|_2$, respectively.
}
\label{fig:latent-fixed}
\end{figure}

\begin{figure}[p]
\centering
\includegraphics[width=\linewidth]
{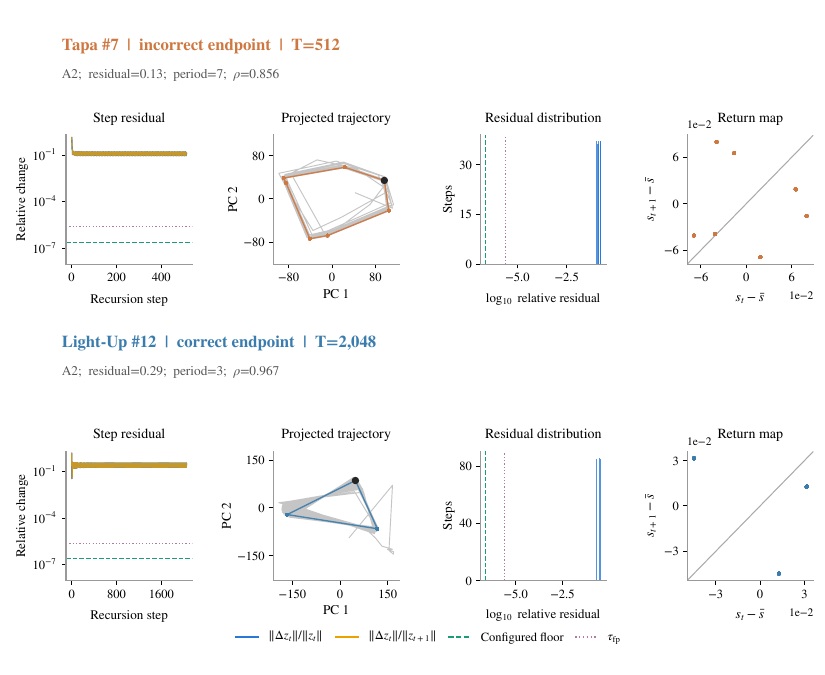}
\caption{
\textbf{Periodic returns with different answer outcomes.}
The Tapa trajectory has an incorrect endpoint and a
detected period of seven after 512 updates.
The Light-Up trajectory has a correct endpoint and a
detected period of three after 2,048 updates.
Both continue moving between successive states.
Period detection and local stability are evaluated
separately: the Light-Up example does not pass the
conservative spectral-radius criterion.
Correctness is checked at the saved endpoint.
Readouts follow Figure~\ref{fig:latent-fixed}.
}
\label{fig:latent-periodic}
\end{figure}

\begin{figure}[p]
\centering
\includegraphics[width=\linewidth]
{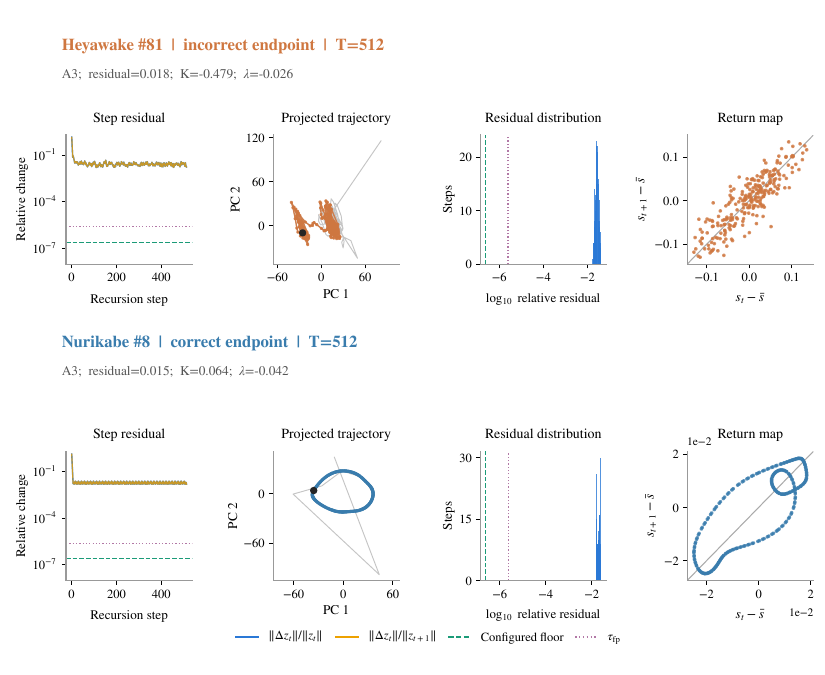}
\caption{
\textbf{Regular-motion examples with different answer outcomes.}
The Heyawake endpoint is incorrect and the Nurikabe
endpoint is correct, both after 512 updates.
Both satisfy the regular-motion diagnostic criteria,
despite their different projected appearances.
The Nurikabe trajectory exhibits a looping projection
while maintaining nonzero update residuals.
Its correct endpoint therefore does not require latent
stationarity.
Readouts follow Figure~\ref{fig:latent-fixed}.
}
\label{fig:latent-regular}
\end{figure}

\begin{figure}[p]
\centering
\includegraphics[width=\linewidth]
{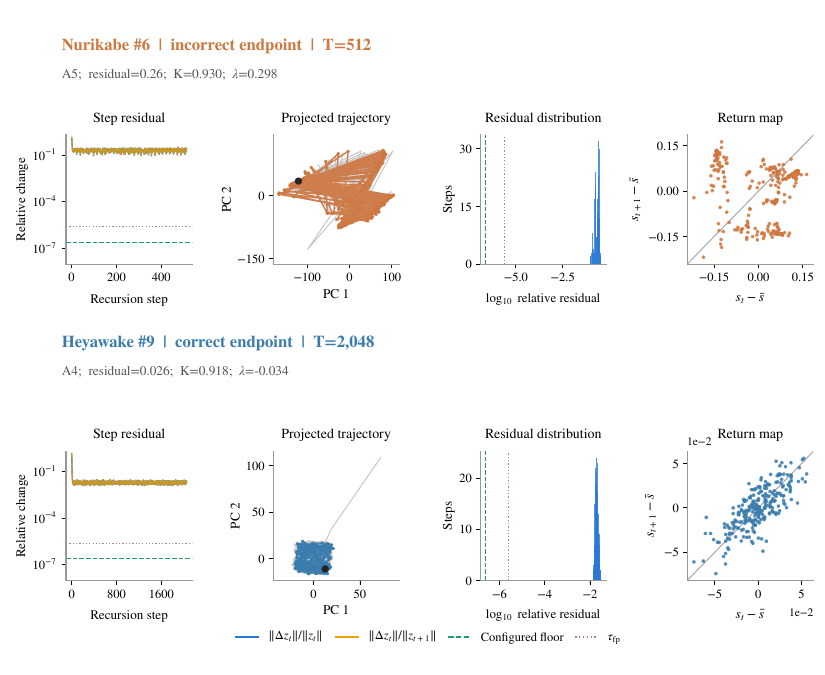}
\caption{
\textbf{Irregular-motion examples with different answer outcomes.}
The Nurikabe endpoint is incorrect after 512 updates;
the Heyawake endpoint is correct after 2,048 updates.
Their small/large-motion distinction uses the relative
projected diameter in
Equation~\ref{eq:trajectory-diameter}.
The Heyawake example satisfies the irregularity
criterion through its 0--1 statistic despite a negative
finite-time perturbation-growth estimate.
Readouts follow Figure~\ref{fig:latent-fixed}.
}
\label{fig:latent-irregular}
\end{figure}

\paragraph{Local stability diagnostics.}
For a detected return period $p$, we examine the linearized update $J_p = D(F_x^p)(z_T)$ using automatic-differentiation Jacobian--vector products. We estimate its spectral radius using 40 power iterations from each of two random initial vectors. For each initialization, we take the geometric mean of the final 30 growth ratios, then retain the larger estimate.

We use $\hat{\rho}<0.95$ as a conservative numerical criterion for local stability and record whether the iteration has converged. A trajectory that passes the return tests but does not meet this stability criterion remains in the corresponding near-fixed-point or periodic class without confirmed stability.
For example, the Light-Up trajectory in Figure~\ref{fig:latent-periodic} has a detected period of three, but its spectral-radius estimate does not pass this stability criterion.

\paragraph{Diagnostics of continued motion.}
For trajectories without a confirmed fixed-point or short-period return, we examine temporal regularity and sensitivity to nearby initial states.

We compute the correlation version of the 0--1 test for chaos~\citep{gottwald2009implementation} from a fixed scalar observable $s_t=a^\top z_t$, where $a$ is a random unit vector. The reported statistic $K$ is the median over 100 test frequencies sampled uniformly from $(\pi/5,4\pi/5)$.
We use up to the final 2,000 observations. Since this implementation uses correlation coefficients, finite-sample values of $K$ can be negative.

We also compute a finite-time perturbation-growth estimate $\lambda$ using nearby trajectories with stepwise renormalization~\citep{benettin1980lyapunov}. Starting from $z_T$, we use four random perturbation directions with initial norm $10^{-3}\|z_T\|_2$ and continue for up to 64 updates. We average the usable logarithmic growth rates across steps and directions, excluding readings from separation collapse or from the first crossing of the saturation threshold $0.1\|z_T\|_2$ onward. If no usable readings remain, the estimate is treated as unavailable.

We assign the regular-motion class when $K\leq0.2$ and either $|\lambda|<0.05$ or $\lambda$ is unavailable. We assign the irregular-motion class when $K\geq0.8$ or an available $\lambda$ exceeds $0.05$. Other cases remain unresolved. These operational classes summarize finite observations; the two diagnostics need not agree on every instance.

For irregular trajectories, we additionally distinguish small and large motion using the relative diameter
\begin{equation}
    d_{\mathrm{tail}}
    =
    \frac{
        \max_{i,j\in\mathcal{W}}\|u_i-u_j\|_2
    }{\|z_T\|_2},
\label{eq:trajectory-diameter}
\end{equation}
where $u_t=Rz_t$ is the saved random projection and $\mathcal{W}$ contains the final 256 recorded states. The threshold is $d_{\mathrm{tail}}=0.1$. The figure annotations A1--A5 denote, respectively, near-fixed points, periodic returns, regular motion, small irregular motion, and large irregular motion.

\paragraph{Latent behavior and answer correctness.}
Figure~\ref{fig:latent-overview} pairs correct and incorrect endpoints across these numerical classes. The two Sudoku MLP examples both approach near-fixed points, although only one returns a correct answer (Figure~\ref{fig:latent-fixed}). Periodic trajectories likewise occur with different endpoint outcomes (Figure~\ref{fig:latent-periodic}).

Correct answers also occur during continued latent motion. The Nurikabe example in Figure~\ref{fig:latent-regular} returns a valid answer while its update residual remains well above the near-fixed-point tolerance. Figure~\ref{fig:latent-irregular} shows a correct Heyawake endpoint with irregular-motion diagnostics. For this example, $K$ exceeds the irregularity threshold while the finite-time perturbation-growth estimate is negative; its classification is therefore supported by the former diagnostic.

Together, these examples show why latent stationarity and answer correctness must be evaluated separately. They illustrate possible combinations of the two, rather than their prevalence across tasks.

\subsection{Stability and robustness of incorrect states}
\label{app:stability-robustness}

Incorrect predictions can persist both under continued recursion and after substantial state perturbations. To characterize this persistence, we distinguish three complementary diagnostics: approximate stationarity, estimated local linear stability, and robustness to finite perturbations. These diagnostics complement the recovery comparisons in Figure~\ref{fig:sigma}.

\paragraph{Approximate stationarity and local stability.}
Among the 4,320 final failures of greedy depth scaling for the seed-0 Sudoku-Extreme TRM checkpoint, 239 instances belong to the incorrect near-fixed-point class. Their classification uses the residual and persistence tests described above. To examine local stability, we estimate the spectral radius of the full-state Jacobian,
\[
    \widehat{\rho}_x(w)
    \approx \rho\!\left(DF_x(w)\right),
\]
at each recorded near-fixed point $w$.

We use Jacobian--vector products in FP32, with the MATH attention backend where applicable. For each of two random initial vectors, we perform 40 power iterations and compute the geometric mean of the final 30 growth ratios.
We report the larger of the two estimates. An estimate passes our convergence check when the relative range of the final 10 growth ratios is below $1\%$ for both initial vectors.

Across all 239 instances, the lower quartile, median, and upper quartile of the estimated spectral radius are $0.893$, $0.919$, and $0.943$, respectively; the maximum is $0.9945$. The convergence check passes for 128 instances, whose estimates range from $0.7384$ to $0.9877$. Together with their small update residuals, these measurements provide numerical evidence of local attraction for a subset of incorrect states. The remaining 111 estimates are reported as unconverged, rather than interpreted as verified stability. Throughout, we retain the label \emph{near-fixed point}: an estimated spectral radius below one is distinct from a certified bound at an exact fixed point.

\paragraph{Robustness to finite perturbations.}
For each of the 239 instances, we draw 16 independent full-state Gaussian perturbations at each scale
\[
    \sigma \in
    \{0.001, 0.003, 0.01, 0.03, 0.1, 0.3, 1\}.
\]
The perturbation is scaled relative to the recorded state:
\[
    \widetilde z_0=w+\varepsilon,
    \qquad
    \varepsilon \sim
    \mathcal{N}\!\left(
        0,\frac{\sigma^2\|w\|_2^2}{d}I_d
    \right).
\]
We then apply 64 recursion steps under the unchanged map $F_x$. At every tested scale, all resulting endpoints retain the original incorrect decoded answer. Moreover, no rule-valid prediction is observed during any of these 64-step continuations. Thus, the failure persists across the tested random directions even when the perturbation magnitude is comparable to the state norm.

\paragraph{Decoded-answer retention and latent return.}
Retaining the same answer does not require returning to the same latent state. We therefore measure decoded-answer retention separately from latent return. Let $S_w$ contain the recorded state $w$ and the final 16 states of its reference trajectory. An endpoint $\widetilde z_{64}$ passes the latent-return test when
\[
    \min_{s\in S_w}\|\widetilde z_{64}-s\|_2
    \leq
    \max\!\left\{
        3f_{\mathrm{num}}\|w\|_2,\,
        \operatorname{diam}(S_w)
    \right\},
\]
where $f_{\mathrm{num}}$ is the calibrated relative numerical floor. Distances in this test are computed in the full latent space.

Table~\ref{tab:wrong-state-robustness} shows that endpoint answer retention remains at $100\%$, whereas latent return within 64 updates decreases as the perturbation scale grows. This separates robust persistence of an incorrect answer from return to its recorded latent neighborhood. Failure to satisfy the finite-horizon return test alone does not establish escape: a trajectory may converge slowly, approach another state with the same decode, or remain in motion.

\begin{table}[t]
\centering
\small
\setlength{\tabcolsep}{7pt}
\begin{tabular}{rrrr}
\toprule
$\sigma$
& Answer retention (\%)
& Solution discovery (\%)
& Latent return (\%) \\
\midrule
0.001 & 100.0 & 0.0 & 75.0 \\
0.003 & 100.0 & 0.0 & 64.3 \\
0.010 & 100.0 & 0.0 & 43.8 \\
0.030 & 100.0 & 0.0 & 25.5 \\
0.100 & 100.0 & 0.0 & 16.8 \\
0.300 & 100.0 & 0.0 & 11.8 \\
1.000 & 100.0 & 0.0 & 6.8 \\
\bottomrule
\end{tabular}
\caption{
Perturbation robustness of 239 incorrect near-fixed points from one Sudoku-Extreme TRM checkpoint. Each scale includes 16 perturbations per instance
(3,824 trials). Answer retention compares the raw endpoint decode with the original decode after 64 updates. Solution discovery records whether any rule-valid prediction appears during those updates, with the original givens restored. Latent return uses the full-state neighborhood criterion defined in the text. All entries are descriptive percentages over perturbation trials.
}
\label{tab:wrong-state-robustness}
\end{table}

\paragraph{Moving failures.}
The same finite-perturbation protocol also reveals persistent failure among the 4,073 irregular moving trajectories. Across the tested scales, approximately $99.8$--$100\%$ of endpoints satisfy the reference-set return criterion. At $\sigma=1$, only $0.13\%$ of endpoints yield a valid answer. For these trajectories, the return tolerance is determined by the diameter of the recorded reference tail. These measurements describe finite-horizon return to a sampled neighborhood; they do not identify a unique invariant attracting set.

Together, these diagnostics show that incorrect answers can persist in approximately stationary states with supporting local-stability evidence, as well as along moving trajectories. They motivate evaluating interventions by their ability to recover valid answers, rather than treating stationarity or perturbation robustness as a proxy for correctness.

\section{State Perturbations and Recovery}
\label{app:sigma}

\subsection{Perturbation Direction and Initialization}

\paragraph{Diagnostic cohorts.}
Figure~\ref{fig:sigma} compares state perturbations and hypothesis injection on three Sudoku-Extreme cohorts, using one trained checkpoint per backbone.
% 239 TRM instances in the incorrect fixed-point class,
% 96 TRM instances with moving failures, and 120 GRAM
% instances with moving failures.
Within each panel, all displayed methods use the same instances. Cold-start comparisons are available for the two TRM cohorts.

% The incorrect fixed-point cohort contains all 239 qualifying instances among 4,320 final failures of the extended greedy-depth run.
% Their saved states are obtained by casting the run's
% terminal states from BF16 to FP32 and continuing for
% 1,024 updates before classification.
% This cohort includes 39 instances that do not pass
% the numerical stability criterion; membership in the
% fixed-point class does not require confirmed attraction.

% The moving-failure states come from 512-step FP32
% classification trajectories starting at $z_0$.
% For GRAM, we use the first 120 qualifying instances
% in the recorded order.
% These selected failure cohorts support diagnostic
% comparisons rather than estimates of full-test accuracy.

\paragraph{Warm and cold initialization.}
Let $w$ denote the saved failed state of an instance. Warm-start interventions perturb $w$, whereas cold-start interventions perturb the model's standard initialization $z_0$. 
After applying an edit, we run 64 updates under the unchanged original input:
\begin{equation}
    z_{\mathrm{end}}^{(k)}
    =
    F_x^{64}\!\left(
        z_{\mathrm{start}}+\varepsilon^{(k)}
    \right),
    \qquad
    z_{\mathrm{start}}\in\{w,z_0\}.
\label{eq:perturbation-rollout}
\end{equation}
All diagnostic rollouts use FP32 computation with halting disabled. For GRAM, subsequent updates follow its deterministic prior-mean path.

\paragraph{Random perturbations.}
We evaluate $\sigma\in\{0.001,0.003,0.01,0.03,0.1,0.3,1\}$. For a full-state Gaussian perturbation,
\begin{equation}
    \varepsilon^{(k)}
    =
    \sigma
    \frac{\|z_{\mathrm{start}}\|_2}{\sqrt{d}}
    \xi^{(k)},
    \qquad
    \xi^{(k)}\sim\mathcal{N}(0,I_d).
\label{eq:random-state-perturbation}
\end{equation}
Its expected squared norm is $\sigma^2\|z_{\mathrm{start}}\|_2^2$. The high-level-only control applies the corresponding construction to $z_H$ and leaves $z_L$ unchanged.

Each perturbation is scaled relative to its own starting state or component. 
% Thus, equal $\sigma$ specifies equal relative scale, not necessarily equal absolute displacement between warm and cold starts or between the two random controls.

\paragraph{Solution-guided perturbations.}
The oracle control uses the reference answer to optimize an edit of the full latent state. We minimize next-update cross-entropy over nonpadding output positions using 30 Adam updates with learning rate 2. After each optimization step, the edit is projected onto the ball
$\|\varepsilon\|_2\leq
\sigma\|z_{\mathrm{start}}\|_2$.
Each scale uses eight random initializations, with initial edit norm $\min\{10,\sigma\|z_{\mathrm{start}}\|_2\}$.
The resulting edits are evaluated using Equation~\ref{eq:perturbation-rollout}.

The transferred-oracle control takes another instance's optimized edit at the same scale, normalizes its direction, and rescales it to the receiving state's norm budget. Donors are assigned cyclically within each recorded evaluation group. These transferred directions retain reference-answer information from their donors.

\paragraph{Learned perturbations on GRAM.}
We additionally evaluate directions drawn from GRAM's learned prior and reference-conditioned posterior. The posterior direction includes its mean displacement relative to the prior mean. Each direction is normalized and rescaled to $\sigma\|z_H\|_2$, applied to $z_H$, and followed by 64 deterministic prior-mean updates.  The posterior and solution-guided controls use the reference answer.

\paragraph{ICScomparison.}
For each instance, the ICS comparator evaluates up to eight model-proposed single-cell hypotheses. Each hypothesis changes the input and starts a fresh
64-step rollout from $z_0$ No reference answers are used to generate these hypotheses.

This comparison measures candidate coverage from one round of hypothesis proposals. It does not execute the complete multilevel ICS search or apply its final model-based answer selection.

\paragraph{Endpoint coverage and uncertainty.}
We decode each rollout at its 64-step endpoint, restore the original givens, and check validity against the original task. An instance is covered if at least one of the first eight attempts returns a rule-valid answer. The comparison therefore uses at most eight 64-step candidate rollouts per instance and condition. The oracle controls additionally incur the optimization cost described above. Coverage is measured at rollout endpoints, rather than by checking whether any intermediate state was correct. Shaded regions and error bars in Figure~\ref{fig:sigma} show pointwise nominal 95\% Wilson intervals, with instances as the statistical unit.

% Copy the single-page PDFs to the manuscript's figures/sigma_atlas/ directory.
% Each PDF is 5.5 in wide; four-row pages are 7.58 in (7.70 in for GRAM).
% 0.84\linewidth keeps that 5.5 in on banglab's 6.5 in line; at full width the
% tallest pages and their captions run off the page.
% Keep captions brief; put common definitions in a preceding paragraph.

\begin{figure}[p]
\centering
\includegraphics[width=0.84\linewidth]{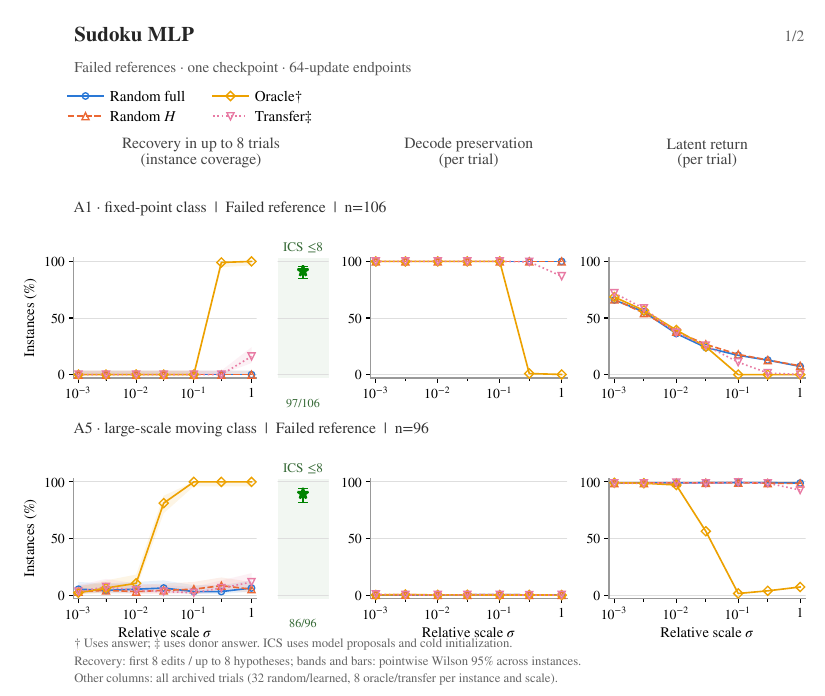}
\caption{Sudoku MLP: failed-reference diagnostics. Left: first-eight strict recovery coverage; ICS uses up to eight cold model hypotheses. Shading and bars are pointwise Wilson 95\% intervals across instances. Middle/right: per-trial decode preservation and latent return.}
\label{fig:sigma-sudoku-trm-seed0-p1}
\end{figure}
\clearpage

\begin{figure}[p]
\centering
\includegraphics[width=0.84\linewidth]{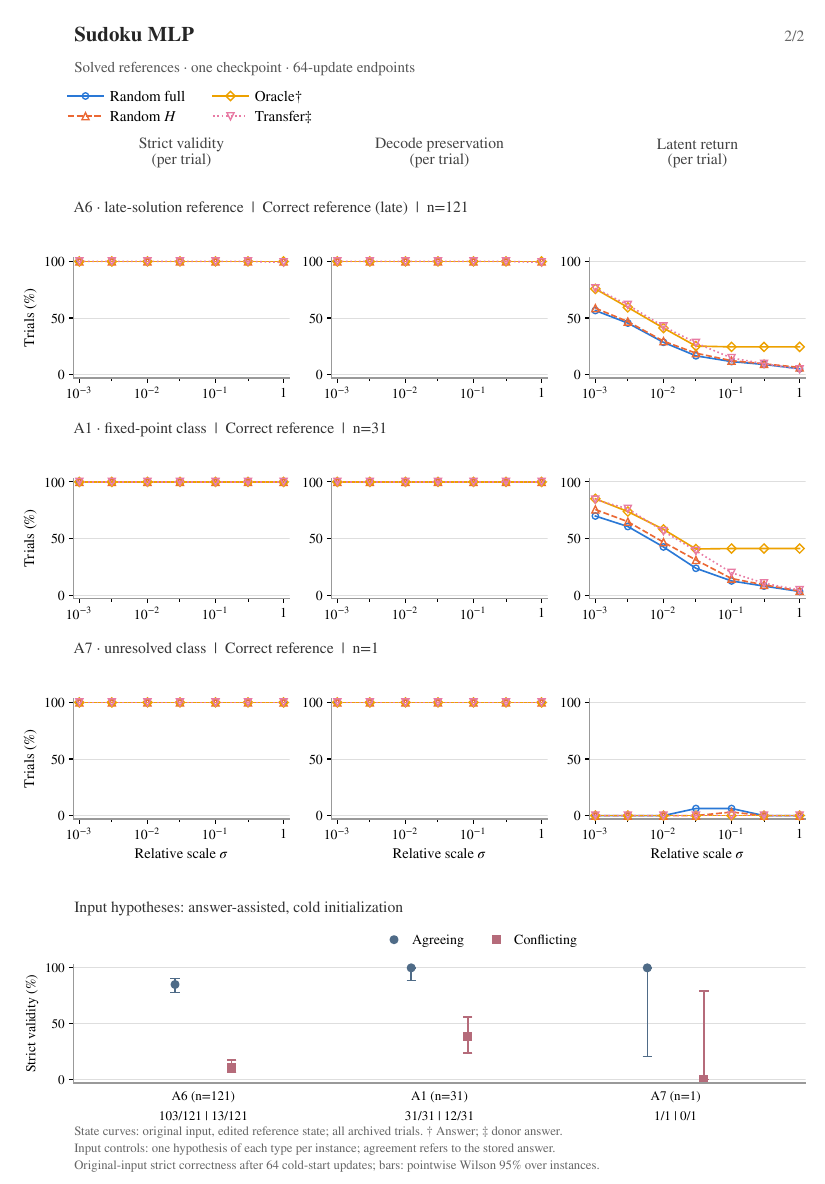}
\caption{Sudoku MLP: solved-reference diagnostics. Top: per-trial state-edit outcomes under the original input. Bottom: separate answer-assisted input hypotheses agreeing or conflicting with the stored answer, evaluated from cold initialization; these are not ICS performance estimates. Bars are pointwise Wilson 95\% intervals across instances; counts follow legend order.}
\label{fig:sigma-sudoku-trm-seed0-p2}
\end{figure}
\clearpage

\begin{figure}[p]
\centering
\includegraphics[width=0.84\linewidth]{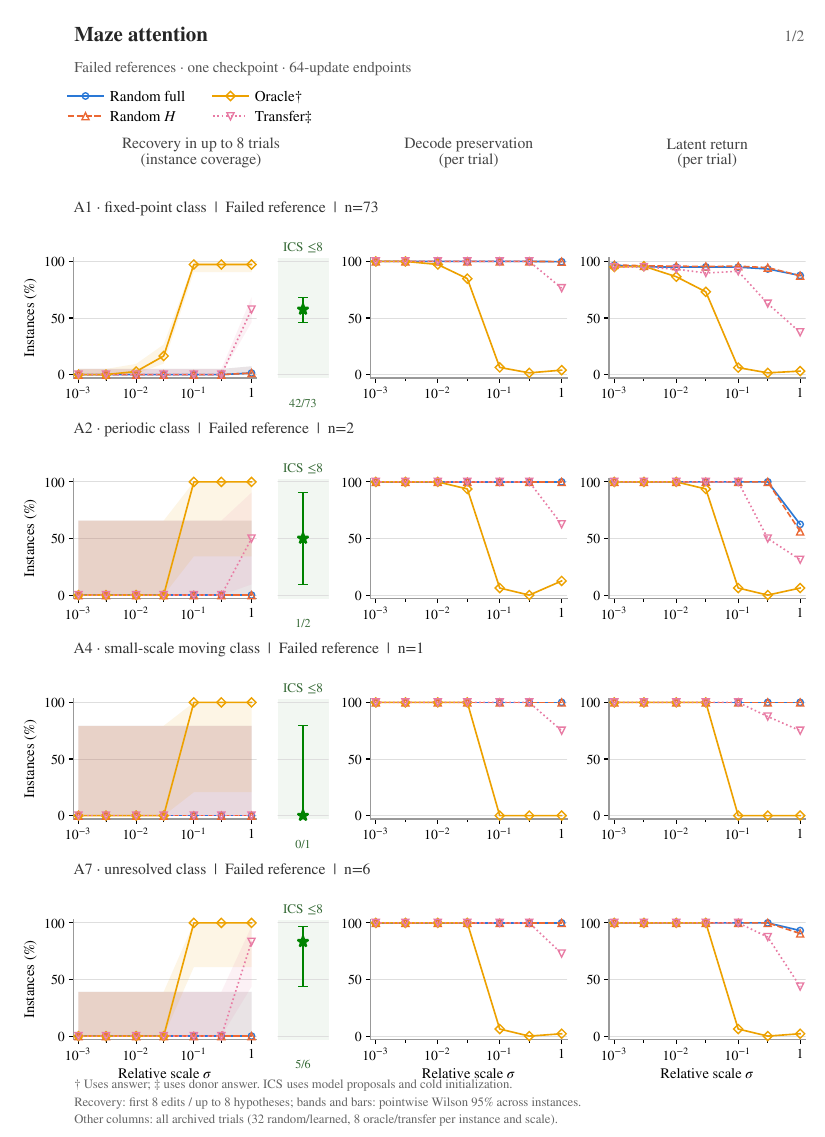}
\caption{Maze attention: failed-reference diagnostics. Left: first-eight strict recovery coverage; ICS uses up to eight cold model hypotheses. Shading and bars are pointwise Wilson 95\% intervals across instances. Middle/right: per-trial decode preservation and latent return.}
\label{fig:sigma-maze-seed0-p1}
\end{figure}
\clearpage

\begin{figure}[p]
\centering
\includegraphics[width=0.84\linewidth]{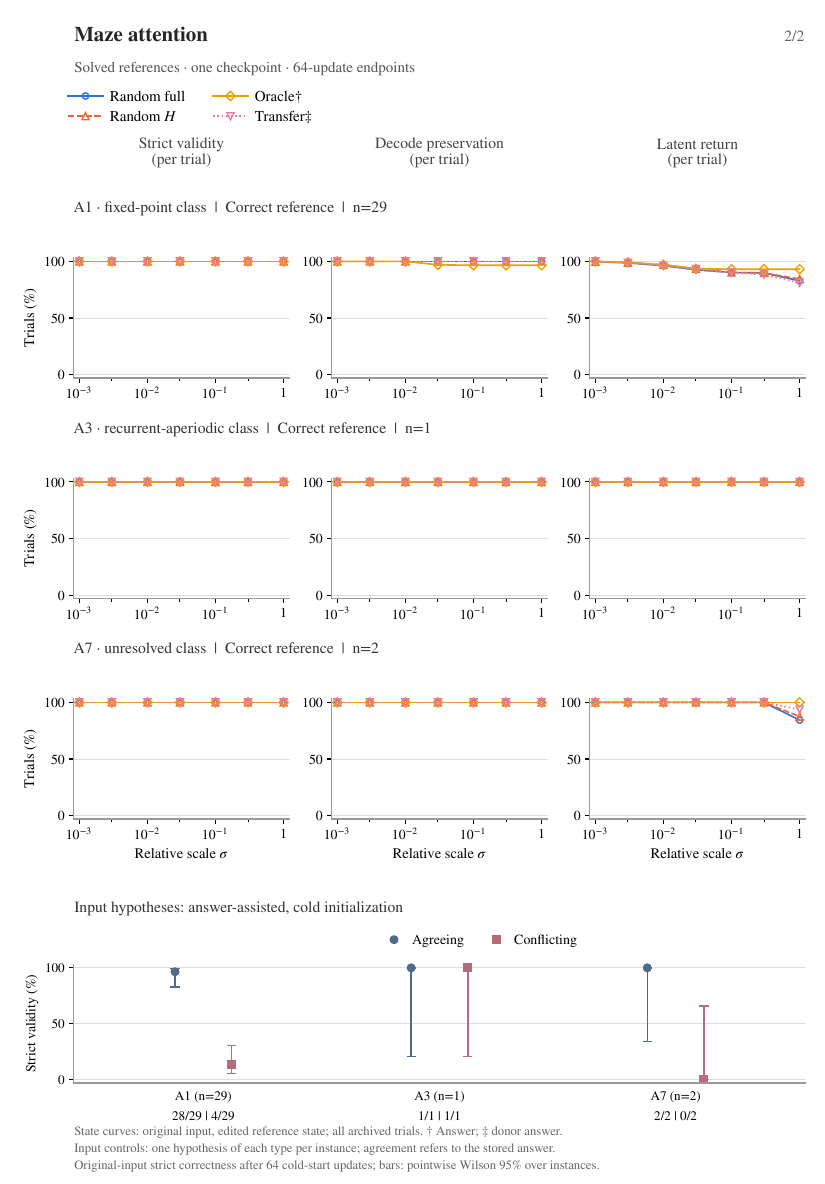}
\caption{Maze attention: solved-reference diagnostics. Top: per-trial state-edit outcomes under the original input. Bottom: separate answer-assisted input hypotheses agreeing or conflicting with the stored answer, evaluated from cold initialization; these are not ICS performance estimates. Bars are pointwise Wilson 95\% intervals across instances; counts follow legend order.}
\label{fig:sigma-maze-seed0-p2}
\end{figure}
\clearpage

\begin{figure}[p]
\centering
\includegraphics[width=0.84\linewidth]{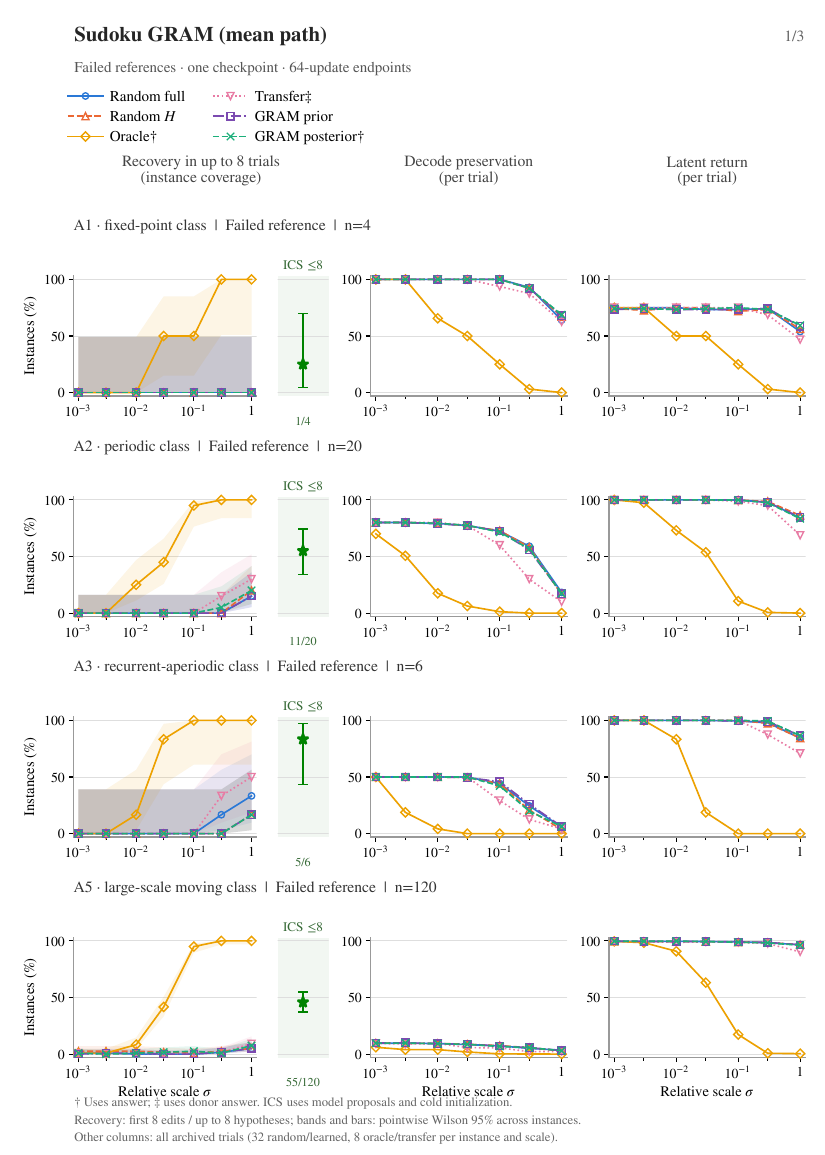}
\caption{Sudoku GRAM (mean path): failed-reference diagnostics. Left: first-eight strict recovery coverage; ICS uses up to eight cold model hypotheses. Shading and bars are pointwise Wilson 95\% intervals across instances. Middle/right: per-trial decode preservation and latent return.}
\label{fig:sigma-sudoku-gram-s0-p1}
\end{figure}
\clearpage

\begin{figure}[p]
\centering
\includegraphics[width=0.84\linewidth]{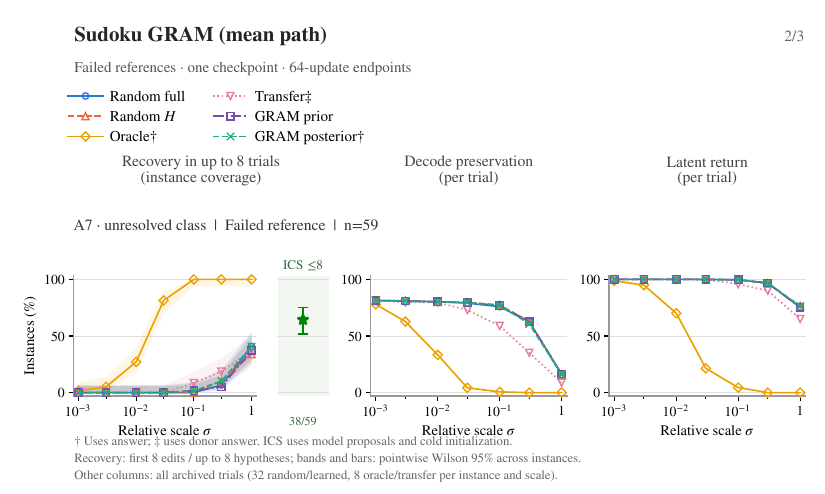}
\caption{Sudoku GRAM (mean path): failed-reference diagnostics. Left: first-eight strict recovery coverage; ICS uses up to eight cold model hypotheses. Shading and bars are pointwise Wilson 95\% intervals across instances. Middle/right: per-trial decode preservation and latent return.}
\label{fig:sigma-sudoku-gram-s0-p2}
\end{figure}
\clearpage

\begin{figure}[p]
\centering
\includegraphics[width=0.84\linewidth]{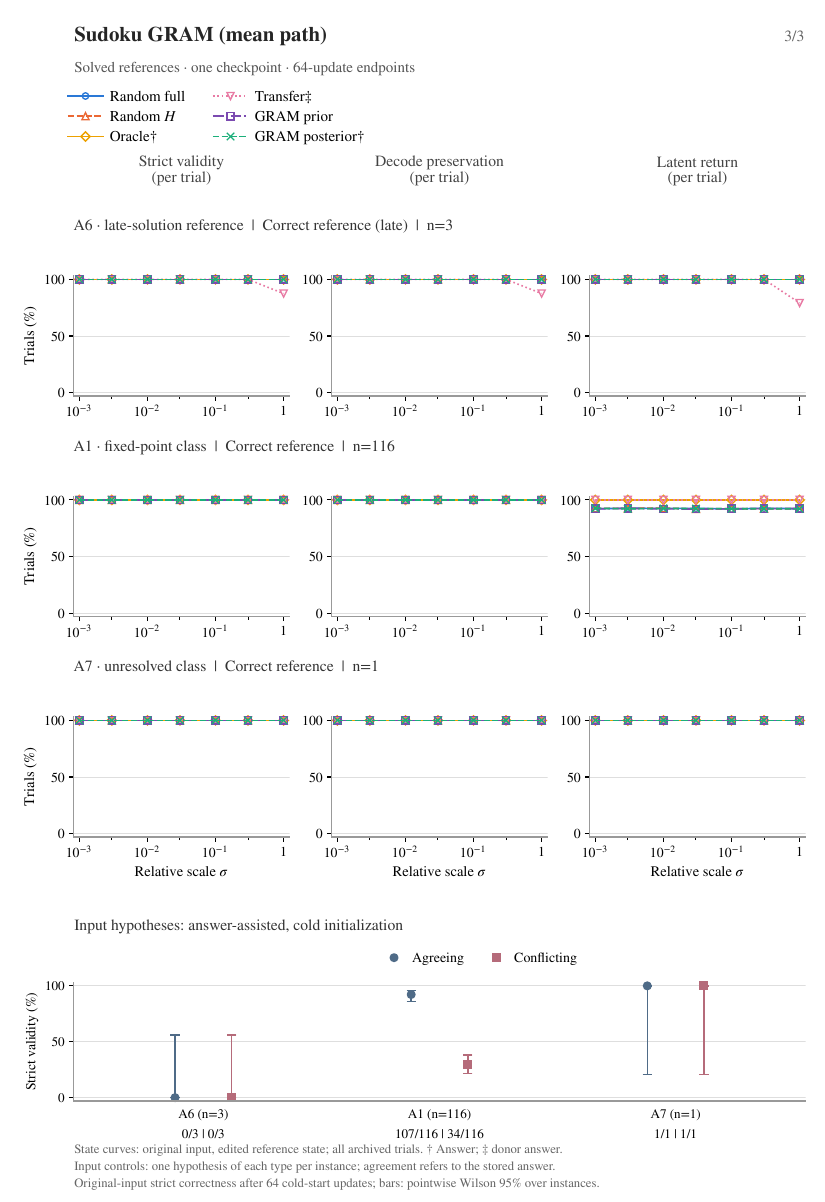}
\caption{Sudoku GRAM (mean path): solved-reference diagnostics. Top: per-trial state-edit outcomes under the original input. Bottom: separate answer-assisted input hypotheses agreeing or conflicting with the stored answer, evaluated from cold initialization; these are not ICS performance estimates. Bars are pointwise Wilson 95\% intervals across instances; counts follow legend order.}
\label{fig:sigma-sudoku-gram-s0-p3}
\end{figure}
\clearpage

\begin{figure}[p]
\centering
\includegraphics[width=0.84\linewidth]{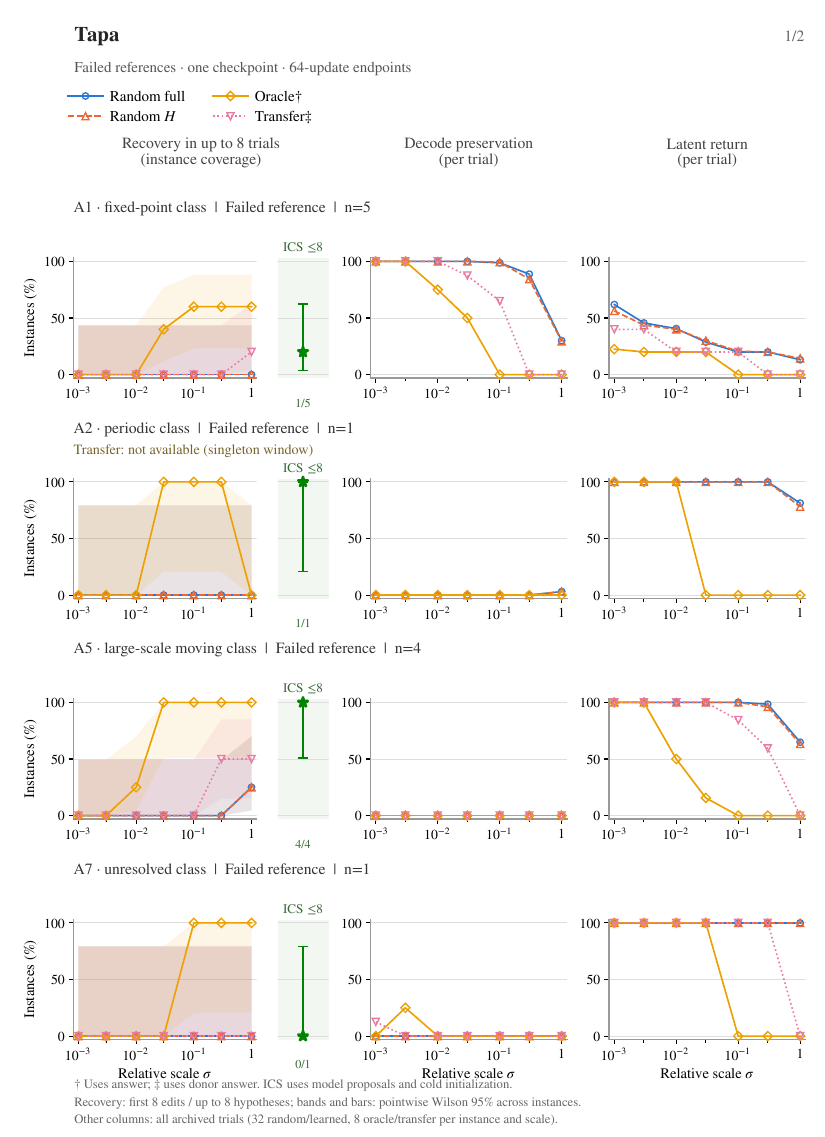}
\caption{Tapa: failed-reference diagnostics. Left: first-eight strict recovery coverage; ICS uses up to eight cold model hypotheses. Shading and bars are pointwise Wilson 95\% intervals across instances. Middle/right: per-trial decode preservation and latent return.}
\label{fig:sigma-ppb-tapa-ppb-tapa-s0-p1}
\end{figure}
\clearpage

\begin{figure}[p]
\centering
\includegraphics[width=0.84\linewidth]{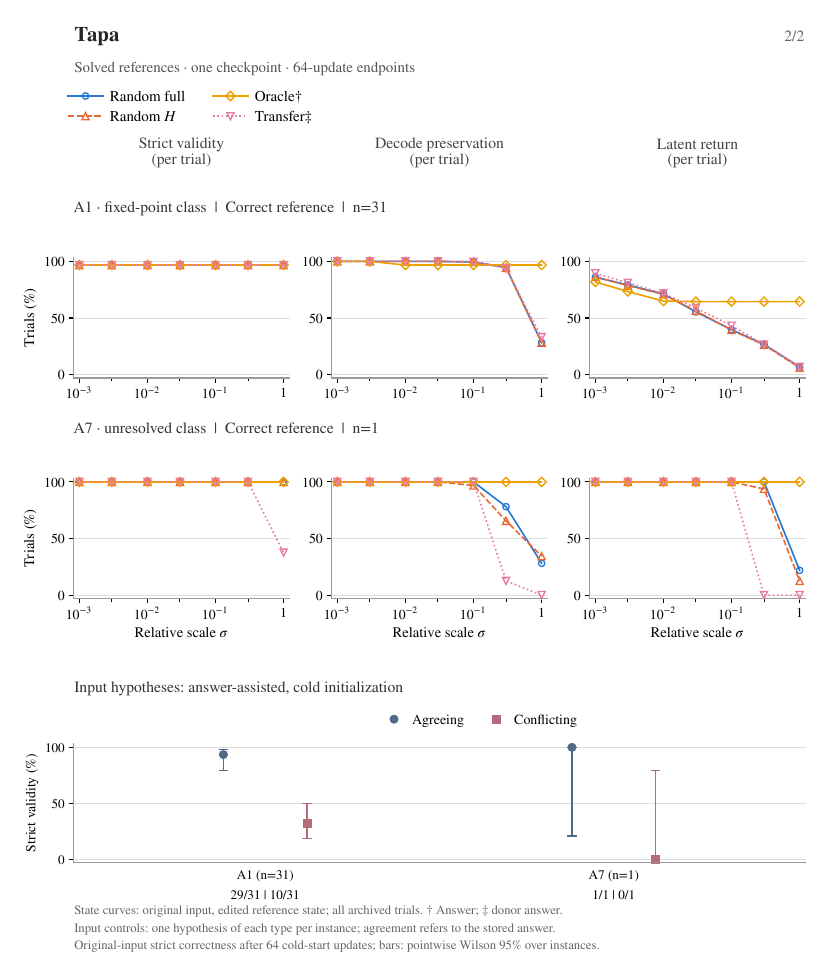}
\caption{Tapa: solved-reference diagnostics. Top: per-trial state-edit outcomes under the original input. Bottom: separate answer-assisted input hypotheses agreeing or conflicting with the stored answer, evaluated from cold initialization; these are not ICS performance estimates. Bars are pointwise Wilson 95\% intervals across instances; counts follow legend order.}
\label{fig:sigma-ppb-tapa-ppb-tapa-s0-p2}
\end{figure}
\clearpage

\begin{figure}[p]
\centering
\includegraphics[width=0.84\linewidth]{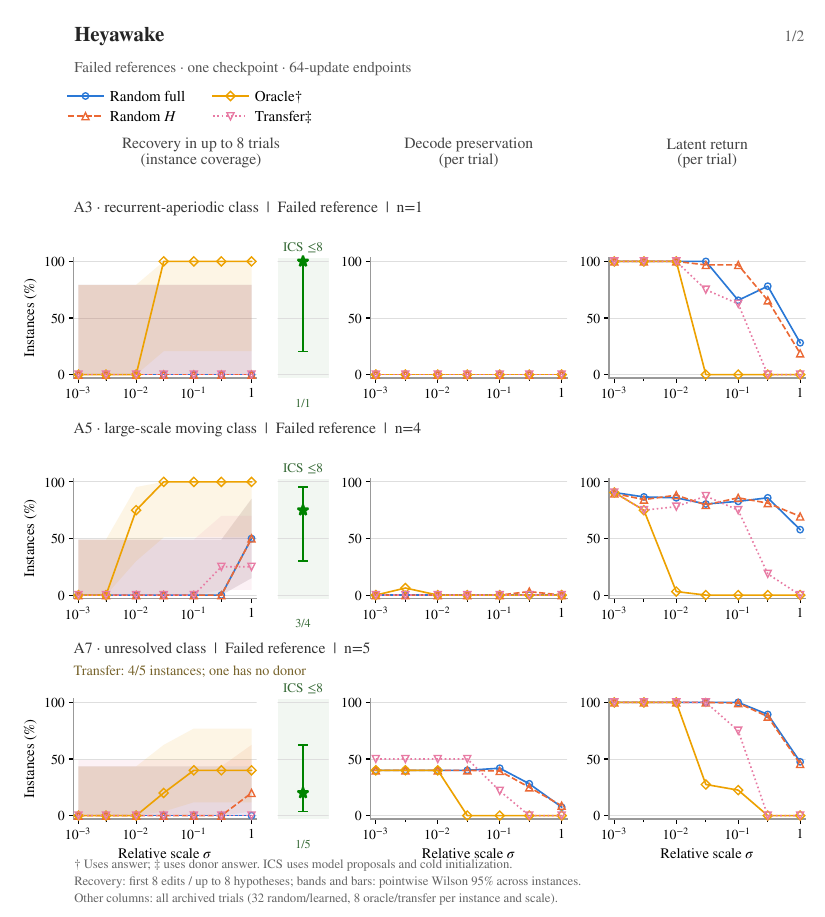}
\caption{Heyawake: failed-reference diagnostics. Left: first-eight strict recovery coverage; ICS uses up to eight cold model hypotheses. Shading and bars are pointwise Wilson 95\% intervals across instances. Middle/right: per-trial decode preservation and latent return.}
\label{fig:sigma-ppb-heyawake-ppb-heyawake-s0-p1}
\end{figure}
\clearpage

\begin{figure}[p]
\centering
\includegraphics[width=0.84\linewidth]{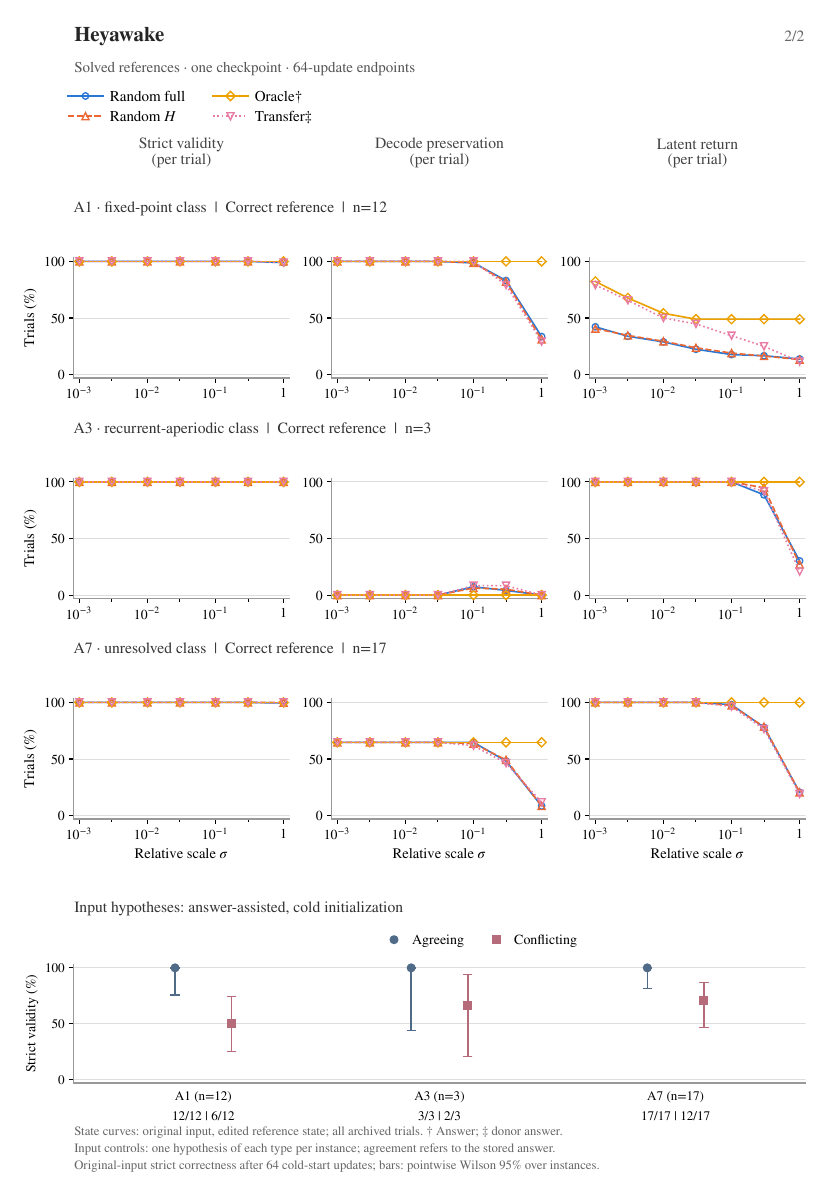}
\caption{Heyawake: solved-reference diagnostics. Top: per-trial state-edit outcomes under the original input. Bottom: separate answer-assisted input hypotheses agreeing or conflicting with the stored answer, evaluated from cold initialization; these are not ICS performance estimates. Bars are pointwise Wilson 95\% intervals across instances; counts follow legend order.}
\label{fig:sigma-ppb-heyawake-ppb-heyawake-s0-p2}
\end{figure}
\clearpage

\begin{figure}[p]
\centering
\includegraphics[width=0.84\linewidth]{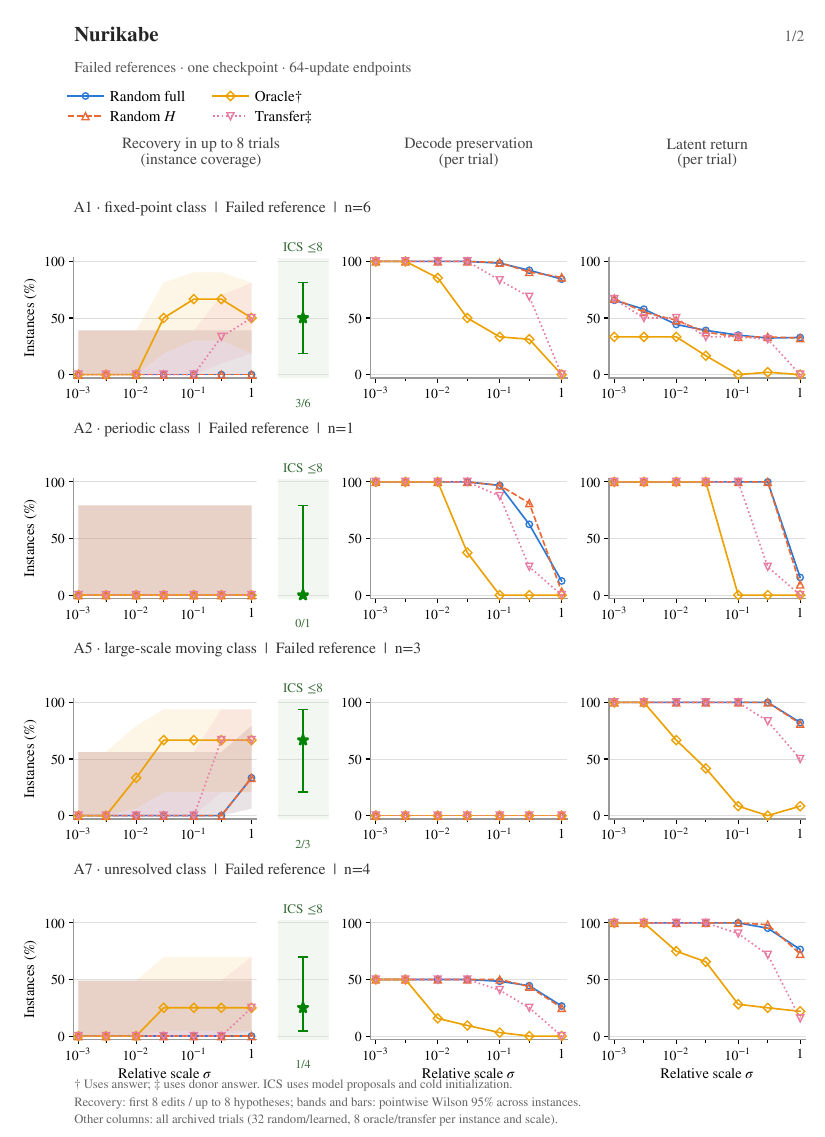}
\caption{Nurikabe: failed-reference diagnostics. Left: first-eight strict recovery coverage; ICS uses up to eight cold model hypotheses. Shading and bars are pointwise Wilson 95\% intervals across instances. Middle/right: per-trial decode preservation and latent return.}
\label{fig:sigma-ppb-nurikabe-ppb-nurikabe-s0-p1}
\end{figure}
\clearpage

\begin{figure}[p]
\centering
\includegraphics[width=0.84\linewidth]{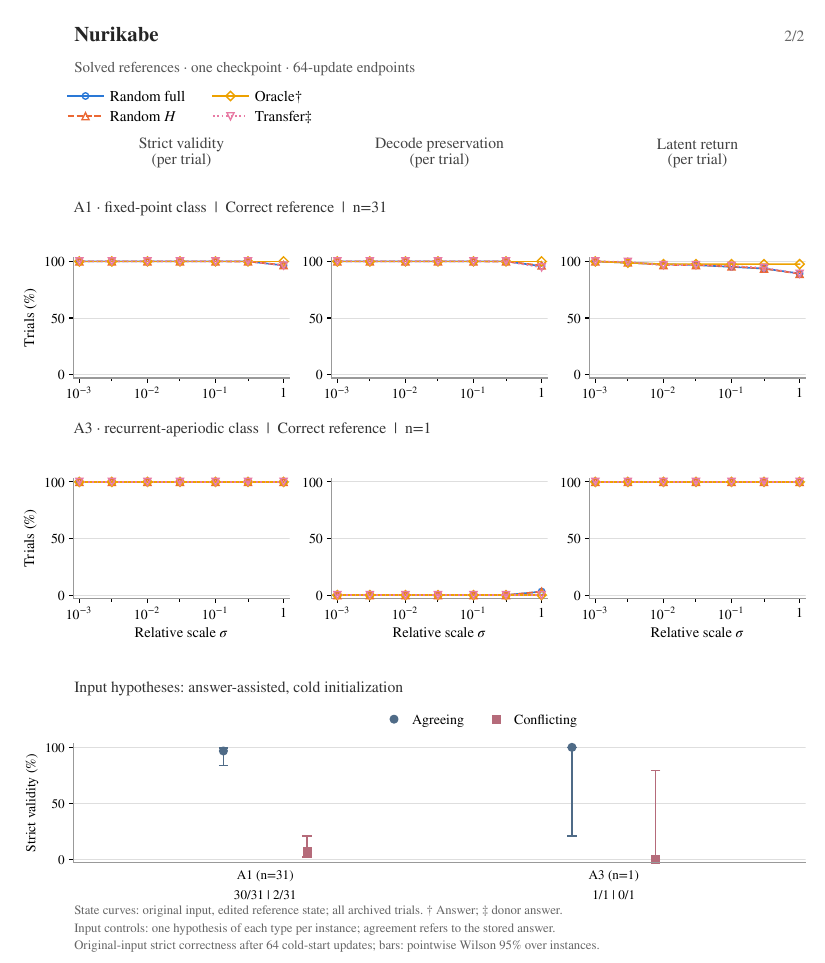}
\caption{Nurikabe: solved-reference diagnostics. Top: per-trial state-edit outcomes under the original input. Bottom: separate answer-assisted input hypotheses agreeing or conflicting with the stored answer, evaluated from cold initialization; these are not ICS performance estimates. Bars are pointwise Wilson 95\% intervals across instances; counts follow legend order.}
\label{fig:sigma-ppb-nurikabe-ppb-nurikabe-s0-p2}
\end{figure}
\clearpage

\begin{figure}[p]
\centering
\includegraphics[width=0.84\linewidth]{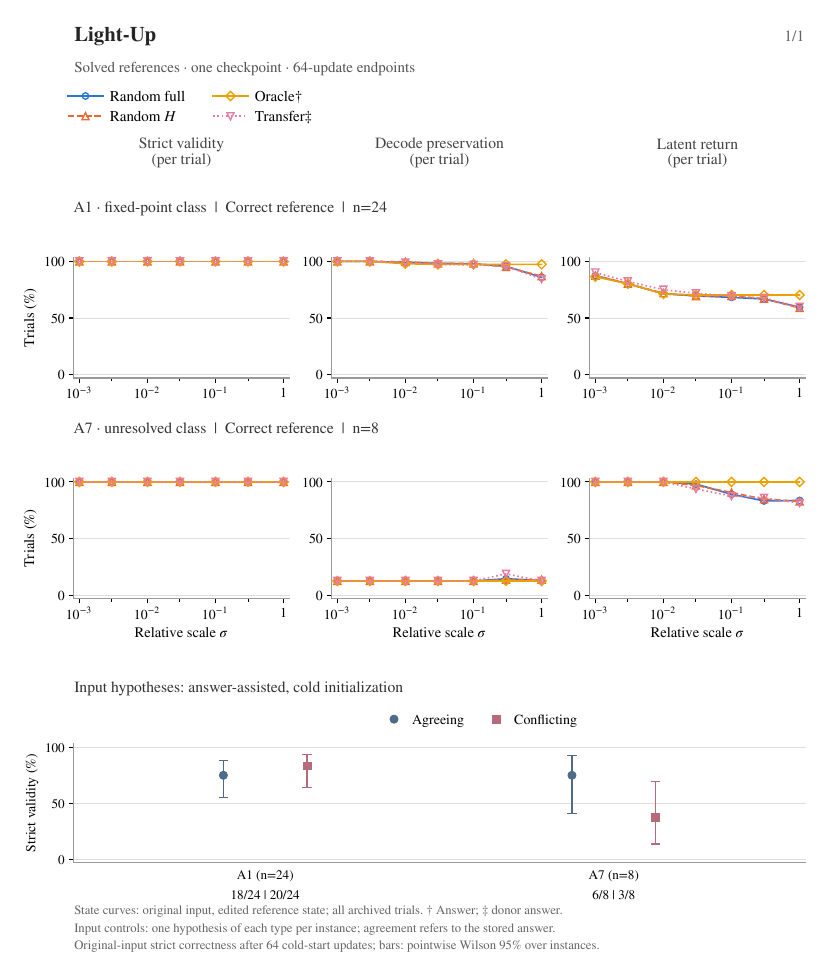}
\caption{Light-Up: solved-reference diagnostics. Top: per-trial state-edit outcomes under the original input. Bottom: separate answer-assisted input hypotheses agreeing or conflicting with the stored answer, evaluated from cold initialization; these are not ICS performance estimates. Bars are pointwise Wilson 95\% intervals across instances; counts follow legend order.}
\label{fig:sigma-lightup-lu26b-s0-p1}
\end{figure}
\clearpage

\subsection{Comparisons Across Failure and Successful States}
\label{app:perturbation-classes}

\paragraph{Coverage of the extended study.}
Figures~\ref{fig:sigma-sudoku-trm-seed0-p1}--%
\ref{fig:sigma-lightup-lu26b-s0-p1}
report all 40 recorded cohorts across six tasks and seven model--task combinations.
Each combination uses one trained checkpoint. Table~\ref{tab:perturbation-cohorts} summarizes the sample sizes; individual numerical classes remain separate in the figures.

This extended study includes a separate cohort of 106 incorrect fixed-point-class Sudoku TRM instances, rather than the 239-instance cohort used in Figure~\ref{fig:sigma}(a). The moving-failure cohorts of 96 Sudoku TRM and 120 Sudoku GRAM instances are shared with Figure~\ref{fig:sigma}(b,c).

References are grouped by whether the saved prediction passes the original-task checker before intervention. The late-solution references, labeled A6, originate from previously failed instances whose classification rollouts have already reached accepted endpoints. They therefore enter this study as successful references. A7 denotes unresolved numerical behavior.

\begin{table}[t]
\centering
\small
\setlength{\tabcolsep}{6pt}
\begin{tabular}{lrrr}
\toprule
Model and task & Failed references & Correct references & Total\\
\midrule
Sudoku TRM       & 202 & 153 & 355\\
Maze TRM        & 82  & 32  & 114\\
Sudoku GRAM      & 209 & 120 & 329\\
Tapa TRM        & 11  & 32  & 43\\
Heyawake TRM    & 10  & 32  & 42\\
Nurikabe TRM    & 14  & 32  & 46\\
Light-Up TRM    & 0   & 32  & 32\\
\bottomrule
\end{tabular}
\caption{
Diagnostic cohorts in the extended perturbation study. Correct references include late-solution references. Counts describe the sampled model--instance pairs, not the frequency of each outcome in the full test sets.
}
\label{tab:perturbation-cohorts}
\end{table}

\paragraph{Three complementary readouts.}
All state perturbations in these extended figures start from the saved reference state $z_\star$ and are followed by 64 updates under the original input. We report three outcomes.

\emph{Strict correctness} requires the endpoint answer to satisfy the original task after restoring givens, and additionally requires its raw decode to respect
all original givens. This adds a given-consistency requirement to the rule-valid metric used in Figure~\ref{fig:sigma}.

\emph{Decode preservation} requires the raw endpoint decode to equal the raw reference decode at every output coordinate, before restoring givens.

\emph{Latent return} measures proximity to the saved reference state and its trailing trajectory. Let $S_\star$ contain $z_\star$ and the final 16 saved full-dimensional states. We count a return when
\begin{equation}
    \min_{s\in S_\star}
    \|z_{\mathrm{end}}-s\|_2
    \leq
    \max\left\{
        3f\|z_\star\|_2,\,
        \operatorname{diam}(S_\star)
    \right\},
\label{eq:latent-return}
\end{equation}
where $f$ is the numerical residual floor. Distances are computed in the full latent space. This readout measures return to the recorded reference neighborhood.

\paragraph{Aggregation within each figure.}
For failed references, the correctness column reports instance coverage: at least one strictly correct endpoint among the first eight perturbation attempts. The separate ICS markers use up to eight cold-start, model-proposed input hypotheses and the same strict endpoint criterion.

For successful references, the correctness column instead reports the fraction of individual interventions that retain strict correctness. Decode preservation and latent return are likewise reported per intervention for both reference groups. These rates use all recorded attempts: 32 per scale for random and learned directions, and eight for solution-guided and transferred directions.

Pointwise nominal Wilson 95\% intervals accompany instance-level coverage and the separate input controls. Per-intervention rates are shown descriptively without treating repeated attempts on the same instance as independent samples.

Transferred directions require a donor instance. The Tapa periodic-failure cohort has no available transfer donor, and the Heyawake unresolved-failure transfer results cover four of its five instances. Each plotted rate uses its available denominator; unavailable controls are omitted.

\paragraph{Input controls on successful references.}
The successful-reference figures additionally show one answer-assisted agreeing hypothesis and one conflicting hypothesis per instance when available. Agreement is defined relative to the stored reference answer. Each modified input is evaluated from fresh initialization for 64 updates and scored against the original task.

These controls examine recomputation under added assumptions, rather than preservation of the saved successful latent state. A returned answer need not obey the additional hypothesis to satisfy the original task.

\paragraph{Correctness, decode preservation, and latent
return can separate.}
The successful Sudoku TRM references provide a clear example. At $\sigma=1$, full-state random perturbations retain both strict correctness and the original decode in all trials of the near-fixed-point cohort, while only $3.2\%$ return to the saved latent neighborhood (Figure~\ref{fig:sigma-sudoku-trm-seed0-p2}). Preserving an answer therefore does not require returning to the same recorded latent state.

Correctness can also persist without exact decode preservation. For the regular-moving Nurikabe reference, all full-state random trials at $\sigma=1$ remain strictly correct, while only $3.1\%$ reproduce the complete raw reference decode (Figure~\ref{fig:sigma-ppb-nurikabe-ppb-nurikabe-s0-p2}).
These readouts capture different aspects of the response to an intervention.

Across failed references, recovery varies with the numerical class and intervention direction. The complete figures retain the small periodic, regular-moving, irregular-moving, and unresolved cohorts alongside the larger fixed-point groups. They support the comparisons within these sampled cohorts; the smaller groups provide descriptive examples rather than estimates of task-wide behavior.

\subsection{Hypothesis Content, Initialization, and Persistence of Near-Fixed Points}
\label{app:hypothesis-interventions}

We examine whether a hypothesis enables recovery by changing the input condition, whether recovery requires a fresh initialization, and whether incorrect near-fixed points persist under the conditioned update map.
These controlled interventions use one frozen Sudoku Extreme TRM-MLP checkpoint. They study individual hypotheses rather than the complete ICS search procedure.

\paragraph{Paired interventions and scoring.}
We use 4,320 previously classified failed instances: 239 in the incorrect near-fixed-point class, 4,073 with moving failures, two with periodic failures, and six in other or unresolved classes.
For each instance, we generate 16 single-cell hypotheses from the decoder scores at its saved failed state $z_\star$. Mutable positions are ordered by increasing top-two logit margin, and we consider two alternative tokens per position, excluding its current prediction. Original givens are never modified. These are the same ordered hypotheses used for these instances in the extended perturbation study.

For each hypothesis $h=(i,v)$, we keep the modified input $x_h$ fixed and run the same map $F_{x_h}$ for 512 recursion steps from either $z_\star$ (\emph{warm start}) or $z_0$ (\emph{cold start}). Neither run stops early. Recovery means that the prediction at step 512 is valid for the original task after restoring its givens. We do not force the injected value into the returned answer. Since these Sudoku instances have unique solutions, the recorded reference-match score after restoring givens agrees with rule validity. Unless stated otherwise, percentages pool instance--hypothesis pairs, rather than count whether any hypothesis succeeds for an instance.

We call a hypothesis \emph{compatible} when its proposed value agrees with the reference solution at the edited position, and \emph{incompatible} otherwise. This label is assigned after proposal generation and is used only for analysis. In particular, compatibility does not imply that the model will find a solution under $F_{x_h}$.

\begin{figure}[t]
    \centering
    \includegraphics[width=\linewidth]{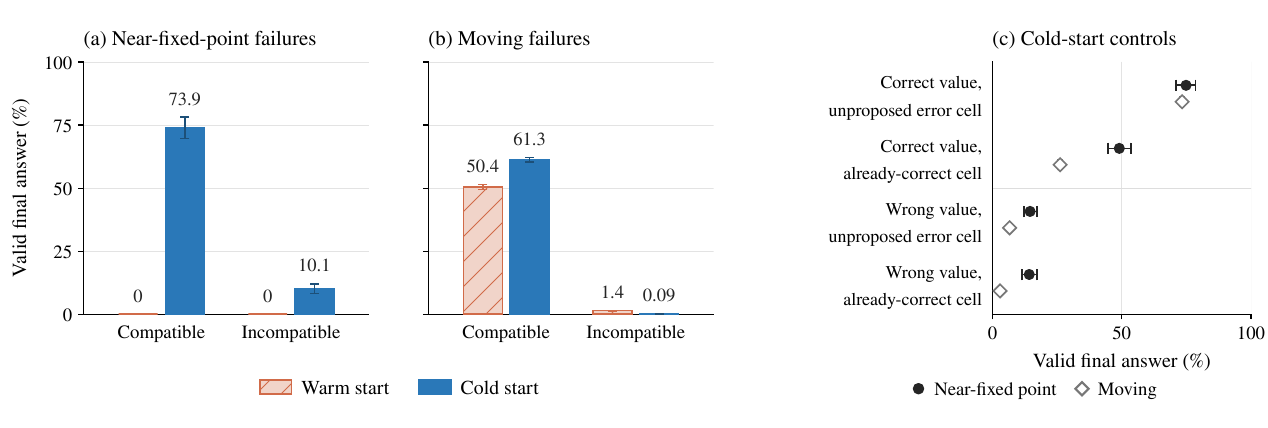}
    \caption{\textbf{Recovery depends on initialization and hypothesis content.}
    All panels use the frozen Sudoku-Extreme TRM-MLP checkpoint
    (training seed~0).
    (a,b) Warm and cold starts use the same instance--hypothesis pairs
    under $F_{x_h}$. Compatible hypotheses agree with the reference
    solution at the edited cell; this label is assigned after proposal
    generation.
    (c) Cold-start, label-informed controls inject correct or wrong
    values at unproposed erroneous cells or already-correct mutable cells.
    All rates measure validity of the final answer after 512 updates
    and restoration of original givens, pooled over
    instance--hypothesis pairs; they do not measure whether any of
    several hypotheses succeeds per instance. Injected values are not
    forced into the returned answer, so incompatible hypotheses can
    still yield valid outputs. Error bars show 95\% confidence intervals
    from 1,000 bootstrap resamples of instances. Sample counts are given
    in Tables~\ref{tab:hypothesis-warm-cold}
    and~\ref{tab:hypothesis-content}.}
    \label{fig:hypothesis-interventions}
\end{figure}

\begin{table}[t]
\centering
\small
\setlength{\tabcolsep}{5pt}
\begin{tabular}{llrrr}
\toprule
Failure cohort & Hypothesis & Pairs & Warm (\%) & Cold (\%)\\
\midrule
Near-fixed point (239)
    & Compatible   & 602    & 0.00   & 73.92\\
    & Incompatible & 3,222  & 0.00   & 10.09\\
Moving (4,073)
    & Compatible   & 17,381 & 50.43  & 61.26\\
    & Incompatible & 47,787 & 1.43   & 0.09\\
Periodic (2)
    & Compatible   & 8      & 100.00 & 100.00\\
    & Incompatible & 24     & 12.50  & 0.00\\
Other/unresolved (6)
    & Compatible   & 9      & 0.00   & 44.44\\
    & Incompatible & 87     & 36.78  & 2.30\\
\bottomrule
\end{tabular}
\caption{Final-answer recovery under paired warm and cold starts.
Each pair uses the same instance and injected hypothesis for both
initializations, with 512 updates under the conditioned map.
Parentheses in the first column give instance counts; the denominator
of each percentage is the number of instance--hypothesis pairs.
The periodic and other/unresolved groups are reported descriptively
because they contain very few instances.}
\label{tab:hypothesis-warm-cold}
\end{table}

\paragraph{A useful hypothesis need not remove an incorrect near-fixed point.}
Figure~\ref{fig:hypothesis-interventions}(a,b) and Table~\ref{tab:hypothesis-warm-cold} compare recovery under the two initializations.
For the near-fixed-point cohort, warm starts recover no solutions, whereas cold starts recover 445 of 602 compatible pairs (73.9\%).
Across all 3,824 hypotheses in this cohort, 3,760 warm endpoints (98.3\%) remain near-fixed and decode to the same incorrect answer as $z_\star$; 3,674 also remain within $0.05\lVert z_\star\rVert_2$ of that state.
Here, an endpoint is classified as near-fixed when every relative step residual in the final 16 updates is at most $2.5\times10^{-6}$.

Thus, input conditioning can permit recovery from a fresh initialization while leaving an incorrect near-fixed state accessible from the failed initialization. For moving failures, compatible hypotheses recover 50.4\% of warm pairs and 61.3\% of cold pairs. The benefit of restarting therefore depends on the starting state and hypothesis; it is not a uniform advantage for every intervention.

\paragraph{The content and location of an injected assignment matter.}
We separate agreement with the reference solution from the location of the edit using four cold-start controls. We inject a correct or incorrect value either at an already-correct mutable cell or at an erroneous cell not selected by the model's proposer. Each control samples up to four eligible cells per instance without replacement. Incorrect control values exclude both the reference value and the saved prediction. These label-informed controls are diagnostic interventions, not inference procedures available to ICS. Control families are paired by instance, but need not select the same cells.

\begin{table}[t]
\centering
\small
\setlength{\tabcolsep}{4pt}
\resizebox{\linewidth}{!}{%
\begin{tabular}{lrr}
\toprule
Injected assignment
    & Near-fixed-point failures
    & Moving failures\\
\midrule
Model proposal, compatible
    & 73.9\% (445/602)
    & 61.3\% (10,648/17,381)\\
Model proposal, incompatible
    & 10.1\% (325/3,222)
    & 0.09\% (45/47,787)\\
Correct value at unproposed error cell
    & 74.9\% (715/955)
    & 73.3\% (11,950/16,292)\\
Correct value at already-correct cell
    & 49.1\% (469/956)
    & 26.2\% (4,262/16,292)\\
Wrong value at unproposed error cell
    & 14.6\% (139/955)
    & 6.6\% (1,082/16,292)\\
Wrong value at already-correct cell
    & 14.2\% (136/956)
    & 2.9\% (480/16,292)\\
\bottomrule
\end{tabular}}
\caption{Cold-start recovery after 512 updates, stratified by hypothesis content. Counts are successful/attempted instance--assignment pairs. An already-correct cell is an editable cell correctly predicted at the saved state, not an original given. Compatible model proposals occur on 225 of 239 near-fixed-point instances and 4,060 of 4,073 moving instances. One near-fixed-point instance has only three eligible unproposed error cells, yielding 955 rather than 956 error-cell controls.}
\label{tab:hypothesis-content}
\end{table}

Correcting an erroneous cell is more effective than repeating an already-correct value
(Figure~\ref{fig:hypothesis-interventions}(c); Table~\ref{tab:hypothesis-content}). Averaging paired differences between per-instance recovery fractions gives gains of 25.8 percentage points for near-fixed-point failures (95\% interval: 22.3--29.8) and 47.2 points for moving failures (46.3--48.1). Intervals use 1,000 bootstrap resamples of instances, preserving all interventions within each instance.

The controls show that recovery depends on what information is injected, not merely on changing an input token. They do not establish that the model's location ordering is optimal: correctly filling unproposed error cells also produces strong recovery. Incompatible hypotheses sometimes recover valid answers because the model is free to revise or disregard an injected value when decoding the original task.

\paragraph{Spectral diagnostics under the conditioned map.}
We additionally estimate the spectral radius of the Jacobian at original and conditioned near-fixed states. Computation uses the full recurrent state $(z_H,z_L)$ in fp32. For each state and input map, we run 40 Jacobian--vector-product power iterations from each of two Gaussian initial vectors. Each run takes the geometric mean of its last 30 norm growth ratios; $\hat\rho$ is the larger of the two estimates.

The implementation checks whether the relative range of the final ten ratios is below $10^{-2}$ for both initial vectors, but records this convergence flag jointly across a batch of states. Not all batches satisfy the check. We report all finite-iteration estimates rather than treating the flag as an individual-state convergence certificate.

\begin{table}[t]
\centering
\small
\setlength{\tabcolsep}{3.5pt}
\resizebox{\linewidth}{!}{%
\begin{tabular}{llrlrr}
\toprule
State & Map & $n$ & Median $\hat\rho$ [Q1, Q3]
    & $\hat\rho<1$ & Batch check\\
\midrule
Original incorrect state
    & $F_x$ & 239
    & 0.9185 [0.8926, 0.9429] & 239 & 128\\
Retained incorrect, compatible
    & $F_{x_h}$ & 585
    & 0.9085 [0.8846, 0.9366] & 585 & 84\\
Retained incorrect, incompatible
    & $F_{x_h}$ & 3,089
    & 0.9076 [0.8829, 0.9339] & 3,087 & 449\\
Recovered correct, matched pairs
    & $F_{x_h}$ & 424
    & 0.9113 [0.8849, 0.9320] & 424 & 47\\
After restoring $x$, same pairs
    & $F_x$ & 424
    & 0.9175 [0.8986, 0.9380] & 424 & 9\\
Retained correct, solved controls
    & $F_{x_h}$ & 6,534
    & 0.9147 [0.8882, 0.9371] & 6,524 & 604\\
\bottomrule
\end{tabular}}
\caption{Spectral-radius estimates at near-fixed states.
The first row has one state per instance; subsequent rows count instance--hypothesis pairs.
Retained endpoints preserve the original decoded answer and lie within $0.05\lVert z_\star\rVert_2$ of the starting state.
The two matched rows use exactly the same 424 compatible interventions on 203 initially failed instances, before and after restoring $x$. 
The last row uses the separate 431-instance solved cohort described below. The final two columns count estimates below one and endpoints belonging
to batches that pass the ratio-stability check, respectively; the latter is not an individual-state convergence rate. Quartiles describe the distribution of estimates, not uncertainty in an
individual spectral radius.}
\label{tab:hypothesis-spectral}
\end{table}

All 585 retained incorrect endpoints under compatible hypotheses have $\hat\rho<1$, with median 0.9085 (Table~\ref{tab:hypothesis-spectral}).
More directly, 422 compatible instance--hypothesis pairs yield a retained incorrect near-fixed endpoint from the warm start and a correct near-fixed endpoint from the cold start, with both spectral radius estimates below one under the same $F_{x_h}$.
This provides numerical evidence consistent with persistent local attraction under the conditioned map, in addition to small update residuals and unchanged incorrect decoding. 
It does not prove stability: the endpoints are approximate and the finite-iteration estimates are not certified bounds. Estimates above one also occur for two retained incorrect endpoints under incompatible hypotheses and ten retained correct endpoints in the solved controls (maxima 1.0051 and 1.0365, respectively).

\paragraph{Preserving a solution differs from reaching it.}
For each cold endpoint, we restore the original input $x$ and run another 512 updates. Among the 10,977 cold endpoints that are correct and near-fixed in the two largest failure cohorts, 10,761 (98.0\%) remain correct and near-fixed and move by at most $0.05\lVert z_\star\rVert_2$. 
The matched spectral rows in Table~\ref{tab:hypothesis-spectral} 
further compare 424 compatible near-fixed-point interventions satisfying this return criterion. 
These tests establish endpoint correspondence at the measured tolerance, not continuous tracking of a fixed-point branch between the two maps.

As a complementary control, we retain 431 instances whose original-input 512-step rollout reaches a correct near-fixed state, from 450 tested instances. We propose 16 alternative-value hypotheses per instance. All 6,896 hypotheses are incompatible with the reference solution, because the starting predictions are correct.
Warm continuation keeps all final answers valid; 6,835 endpoints (99.1\%) remain near-fixed with the same decoded solution, including 6,534 within
the retention tolerance. Cold starts under these identical hypotheses recover only 1,040 of 6,896 pairs (15.1\%). 
Thus, a conditioned map can preserve an existing solution even when that solution is difficult to reach from initialization.

Together, these interventions support searching over hypothesis content and recomputing candidates from a fresh state. 
They show that recovery can improve without eliminating every incorrect near-fixed point, and that a solution's persistence alone does not ensure its accessibility from a particular initialization.
\subsection{Sudoku component and proposal ablations}
\label{app:ablation}

We compare a fixed-depth TRM baseline with seven search configurations using the same seed Sudoku model; the search runs use inference seed $7$. The first eight disjoint windows in test-loader order contain 6,144 instances each, totaling 49,152. 

\paragraph{Components and selection.}
A is TRM with depth scaling to $D=20$: it executes all 20 recursion steps on the original input and returns the highest-$q$ prediction along the trajectory. The same output is evaluated in both columns, giving $87.85\%$ accuracy at 20 steps ($0.375$ TFLOPs) per instance; A does not use a task certificate to choose its answer.  B adds hypothesis search with 16 steps per child; C adds up to 32 symmetry restatements of the original input, each with 16 steps. Thus AB, AC, and ABC retain the original search settings, while A is a fixed-depth reference. Their differences from A include adaptive original-input computation as well as the added search stages.

For the search configurations, certificate selection checks each candidate against the original Sudoku rules. Model selection uses the engine's $q$ terminal without a task checker: the original-input and hypothesis stages can accept a positive halting logit; hypothesis candidates are ordered by assumption contradictions, original-input consistency, and descending $q$. The restatement stage first accepts a positive-$q$ candidate, then uses agreement of at least two complete outputs, and finally falls back to the highest-$q$ restatement if needed. This protocol is not a fixed-pool best-$q$ rule. All reported accuracies are the offline rule-validity rates of delivered answers after restoring original givens; reference-solution accuracy agrees in these records. Without restatements, an unaccepted AB run returns the last search level's top-ranked candidate. These outputs also enter accuracy.

\paragraph{Search budgets and proposal controls.}
Both certificated-based selection and model-based selection use two alternative values per proposed location, subject to the proposal cap. The small budget arms cap proposals at six per parent and total children at 16; the respective levels and beam widths are unchanged. The implementation caps the number of proposals, not the number of distinct locations. Random-location controls retain model-ranked alternative values. 
The joint random control also samples alternative values without replacement from the digit vocabulary, excluding the current predicted value; it does not filter proposals by Sudoku constraint satisfaction.

For the search configurations, each window stores accuracy to four decimal places. With its known denominator, the number of correct outputs is uniquely recoverable; we pool those counts. A reports the 20-step baseline accuracy.

\begin{table}[t]
\centering
\small
\setlength{\tabcolsep}{4pt}
\resizebox{\linewidth}{!}{%
\begin{tabular}{lrrrrrr}
\toprule
& \multicolumn{3}{c}{Certificate selection} & \multicolumn{3}{c}{Model selection}\\
\cmidrule(lr){2-4}\cmidrule(lr){5-7}
Configuration & Acc. (\%) & Steps & TFLOPs & Acc. (\%) & Steps & TFLOPs\\
\midrule
A: TRM + depth scaling ($D=20$) & 87.85 & 20.0 & 0.38 & 87.85 & 20.0 & 0.38\\
AB: + hypotheses & 99.92 & 74.0 & 1.39 & 99.75 & 58.8 & 1.10\\
AC: + restatements & 98.83 & 63.7 & 1.19 & 98.82 & 63.6 & 1.19\\
ABC: full HIS & 99.94 & 74.4 & 1.39 & 99.81 & 59.8 & 1.12\\
ABC: random locations and values & 99.85 & 75.5 & 1.42 & 98.81 & 58.8 & 1.10\\
ABC: random locations & 99.90 & 73.8 & 1.38 & 99.78 & 59.4 & 1.11\\
ABC: small budget & 99.63 & 56.8 & 1.06 & 99.42 & 54.8 & 1.03\\
ABC: small budget, random locations and values & 99.34 & 64.3 & 1.20 & 97.45 & 50.4 & 0.95\\
\bottomrule
\end{tabular}}
\caption{ICS ablation on Sudoku-extreme (with 49,152 instances).
A uses the same best-$q$ output in both columns; its exact cost is $0.375$ TFLOPs.
Search configurations retain adaptive original-input attempts.
Both accuracy columns evaluate delivered answers with the original task rules;
only certificate-selected search consults those rules during selection.
Steps and TFLOPs are mean executed model computation per instance, charging
every launched evaluation. One step costs $1.875\times10^{10}$ FLOPs.
Search accuracies pool reconstructed counts; search costs use rounded window means. No multi-seed uncertainty is implied.}
\label{tab:ablation-complete}
\end{table}

\paragraph{Interpretation.}
Relative to full ICS, omitting hypotheses (AC) lowers accuracy by 1.11 percentage points with a certificate and 0.99 points under model selection. Omitting restatements (AB) lowers accuracy by 0.02 and 0.06 points, respectively. 
These comparisons use the shared adaptive original-input schedule of the search configurations. 
Random locations only result in a small gap from the full ICS. Randomizing values instead lowers it by 1.00 point, whereas certificate selection remains above 99.85\%. Under the small budget, the joint random control loses 1.96 points under model selection and 0.29 points with a certificate. The results support useful information in the model's value proposals. Because the terminal rule affects which later candidates are evaluated, the two selection columns, which also differ in search budgets, do not isolate scoring error on a common candidate pool.

\section{Input-Conditioning Experiments}
\label{app:input-conditioning}

The preceding experiments examine state interventions under the original-input update map $F_x$. We now examine interventions that change the input condition and therefore the update map. With model parameters held fixed, a hypothesis $H$ replaces $F_x$ with $F_{\tilde{x}}=F_{x_H}$ or a restatement $\Phi$ replaces $F_x$ with $F_{\tilde{x}}=F_{\Phi(x)}$.

Input conditioning can locally eliminate incorrect attracting fixed points. Numerical continuation in the benchmark models provides evidence of the underlying folds, and computer-assisted certificates in separately trained small reasoners establish this mechanism (Appendices~\ref{app:fixed-point-continuation}, \ref{app:small-model-certificate}, and~\ref{app:fold-theory}). The experiments also examine how input conditions and initialization affect access to correct solutions.

We first describe the construction of the update-residual surfaces and recorded trajectories in Figure~\ref{fig:input-conditioning}, then present the continuation and certification results. We subsequently examine aligned restatement dynamics, compare fresh initialization with continuation from a saved failed state, and vary the duration of hypothesis conditioning. Throughout these experiments, answer correctness is evaluated against the original task $x$.

\subsection{Construction of the update-residual visualization}
\label{app:landscape-construction}

Figure~\ref{fig:input-conditioning} visualizes a Maze-Hard instance using the seed-0 TRM checkpoint. The surfaces are constructed from model evaluations, and the overlaid paths are projections of recorded full-state trajectories.

\paragraph{Reference states and input intervention.}
We run the original-input model for 512 recursion steps from $z_0$ to obtain an incorrect near-fixed point $s_w$. For this diagnostic example, the hypothesis $H$ introduces one additional "path" token at a mutable position.

Under the resulting map $F_{x_H}$, we run two trajectories for 1,024 updates: a cold-start trajectory from $z_0$ and a warm-start trajectory from $s_w$. 
We use the cold-start endpoint as the correct near-fixed point $s_c$. 

\paragraph{A shared two-dimensional section.}
We construct an affine plane through $z_0$, $s_w$, and $s_c$ in the full recurrent state space. Its orthonormal basis is
\begin{equation}
\begin{aligned}
e_1 &=
\frac{s_w-z_0}{\|s_w-z_0\|_2},\\
a &=
(s_c-z_0)-\langle s_c-z_0,e_1\rangle e_1,
\qquad
e_2=\frac{a}{\|a\|_2}.
\end{aligned}
\end{equation}
A point on the section corresponds to the full latent state
\begin{equation}
    z(u,v)=z_0+u e_1+v e_2.
\end{equation}
Both input conditions are evaluated at the same section
coordinates, so differences between their surfaces arise
from changing the input condition.

\paragraph{Measured residuals and interpolation.}
For each sampled state, we evaluate the relative
one-step update residual
\begin{equation}
    r_x(z)=
    \frac{\|F_x(z)-z\|_2}{\|z\|_2},
    \qquad
    r_{x_H}(z)=
    \frac{\|F_{x_H}(z)-z\|_2}{\|z\|_2}.
\end{equation}
Panels (a) and (c) display the corresponding base-10
log residuals.
Panel (b) displays
\begin{equation}
    \Delta(u,v)=
    \log_{10} r_{x_H}(z(u,v))
    -
    \log_{10} r_x(z(u,v)).
\end{equation}
Positive values therefore indicate that conditioning increases the relative update magnitude at the same state.

We begin with a $45\times45$ grid and add nested $9\times9$ grids around the coordinates of $s_w$ and $s_c$, with half-widths $32/4^j$ for $j=0,\ldots,7$. After deduplication, the section contains 3,195 sampled coordinates. Both maps are evaluated in FP32. We construct a shared triangulation in the original section coordinates and use piecewise-linear interpolation of the measured log residuals. The interpolant is defined before applying any display coordinate transformation.

\paragraph{Recorded trajectories.}
Each full-state trajectory is projected onto the section
using coordinates
\[
    \bigl(
    \langle z_t-z_0,e_1\rangle,\,
    \langle z_t-z_0,e_2\rangle
    \bigr).
\]
For visualization, its projected path is placed on the corresponding interpolated residual surface. These displayed heights describe the residual at the reconstructed section coordinate; the original trajectory can also have components outside the plane.

The illustrative perturbation in panel (a) is applied at $z_0$, corresponding to $t=0$ in the general annotation $z_t+\varepsilon_t^{(k)}$. It is an in-plane displacement of radius $800$ at angle $225^\circ$, selected from trajectories that return to $s_w$. This example illustrates a returning perturbed trajectory; the quantitative comparisons of perturbation methods use the separate protocol described with Figure~\ref{fig:sigma}.

The perturbed initialization projects outside the sampled
rectangle. For rendering, out-of-domain projected coordinates are clamped to its boundary before surface interpolation. Consequently, the displayed perturbed starting point and its connector are schematic. The underlying trajectory is unchanged, and no residuals are extrapolated outside the sampled section.

\paragraph{Display coordinates.}
The residual minima occupy small regions relative to the full trajectory extent. To make both visible, all panels use the same nonlinear horizontal coordinate transformation, which magnifies neighborhoods of $s_w$ and $s_c$ and compresses distant
context.
A monotone display-height transformation compresses extreme log-residual values while retaining shallow relief. The residual panels share the same height and color scales, and axis and colorbar labels report the original values.

The same horizontal transformation is applied to the surfaces, trajectories, and markers. Local transparency adjustments reveal surface regions otherwise obscured by foreground plateaus. These rendering choices leave the measured residuals and recorded trajectories unchanged, but displayed widths, distances, and slopes are not latent-space metric measurements.
The surfaces represent update residuals, rather than an energy function whose gradient generates the dynamics.

\paragraph{Endpoint checks.}
Direct evaluation at the recorded state $s_w$ gives $r_x(s_w)\approx2.41\times10^{-7}$ and $r_{x_H}(s_w)\approx0.22465$. Thus, the same state that is nearly stationary under the original input has a substantial update under the hypothesis-conditioned map. Both conditioned trajectories return valid answers, and their terminal states differ by only $2.40\times10^{-7}$ in relative norm. These endpoint measurements use the saved full states, rather than values read from the interpolated surfaces.

\subsection{Input conditioning induces folds in incorrect fixed-point branches}
\label{app:fixed-point-continuation}

Numerical continuation in the trained Maze-Hard and Sudoku-Extreme models identifies local folds in incorrect fixed-point branches as the input condition changes. At a fold, two fixed-point branches meet and terminate locally on one side of a critical parameter value. This provides a mechanism through which changing the input can remove an incorrect attracting fixed point. We first present the numerical evidence, then establish this mechanism with computer-assisted certificates in trained small reasoners (Appendix~\ref{app:small-model-certificate}).

We use $\lambda$ to interpolate between the original input embedding ($\lambda=0$) and the hypothesis-conditioned embedding ($\lambda=1$). The curves below track fixed points across these conditions; each plotted state approximately satisfies the fixed-point equation for its own condition.

\begin{figure}[t]
\centering
\includegraphics[width=\linewidth]{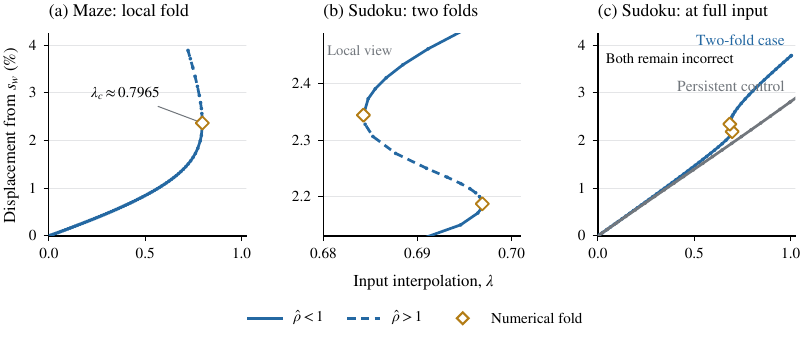}
\caption{Numerical continuation in the trained Maze-Hard and Sudoku-Extreme models. (a) The Maze example exhibits a fold near $\lambda=0.79650$. (b) A Sudoku example has two folds, revealing an intervening unstable branch. (c) Its full continuation and a persistent Sudoku control reach the fully conditioned input. Points are connected in continuation order; vertical coordinates report $100\|z-s_w\|_2/\|s_w\|_2$. Panel (b) magnifies the fold region. Stability styles use numerical multiplier estimates, not interval certificates.}
\label{fig:fixed-point-continuation}
\end{figure}

\paragraph{A local fold in Maze-Hard.}
For the Maze-Hard landscape example (instance 461, hypothesis at cell 576), refinement locates a fold at $\lambda_c\approx0.796498891$. The relative fixed-point residual is $2.03\times10^{-15}$ and the unit-eigenvector equation residual is $7.01\times10^{-14}$. The leading multiplier approaches $+1$, while the next-largest computed multiplier modulus is approximately $0.609$. Automatic differentiation gives nonzero fold coefficients $a\approx-294.62$ and $b\approx-0.12645$, using the normalization in Appendix~\ref{app:fold-theory}. A finer continuation step gives the same fold location to approximately eight decimal places. Together, these measurements provide numerical evidence of a local saddle-node involving the incorrect attracting branch.

\paragraph{Condition-dependent branch structure in Sudoku.}
For Sudoku instance 60398 with a hypothesis at cell 40, continuation follows the original attracting branch to a fold near $\lambda=0.696896$. It then reverses direction in $\lambda$, follows an unstable segment to a second fold near $0.684264$, and increases $\lambda$ again along another attracting branch to reach $\lambda=1$ (Figure~\ref{fig:fixed-point-continuation}). This is the order of fixed-point continuation, not a sequence of reasoning steps or a monotone increase of the input parameter. The tracked states retain the original incorrect decode.

A different hypothesis on this instance, and the controls on Maze instance 69 and Sudoku instance 357158, reach the fully conditioned map without detected folds. Together, these cases show that the structure of incorrect fixed-point branches depends on the input condition: a branch can terminate locally at a fold, while another incorrect attracting branch can persist under the final condition.

\paragraph{Continuous input intervention.}
Let $e(x)$ and $e(x_H)$ be the frozen model's input embeddings. We define
\begin{equation}
 e_\lambda=(1-\lambda)e(x)+\lambda e(x_H),
 \qquad G(z,\lambda)=F_\lambda(z)-z,
 \label{eq:embedding-continuation}
\end{equation}
where $F_\lambda$ evaluates the recursive update with embedding $e_\lambda$. Thus $F_0=F_x$ and $F_1=F_{x_H}$. Intermediate embeddings need not encode literal puzzles; this interpolation is a diagnostic path between the two conditions, not an additional ICS inference operation. The complete recurrent state and all model parameters are held in the same coordinates throughout.

\paragraph{Following stable and unstable branches.}
Starting from an incorrect near-fixed point, we solve $G(z,\lambda)=0$ using FP64 Newton--Krylov correction and pseudo-arclength continuation. An additional arclength equation allows the continuation parameter $\lambda$ to reverse direction at a fold, so the method can follow unstable fixed points that ordinary forward iteration would leave. Jacobian--vector products support the matrix-free linear solves and Arnoldi estimates of leading multipliers. We refine candidate folds by solving the augmented fixed-point, unit-multiplier, and eigenvector-normalization equations described in Appendix~\ref{app:fold-theory}. These calculations concern the full recurrent state; projections are used only for display.

\subsection{Certified elimination of incorrect attracting fixed points}
\label{app:small-model-certificate}

Input conditioning can locally eliminate incorrect attracting fixed points in trained recursive reasoners. We establish this result with 51 complete computer-assisted certificates across selected model--instance--hypothesis cases in separately trained $4\times4$ Sudoku models (Table~\ref{tab:small-model-folds}). Each certificate verifies the sufficient conditions of Appendix~\ref{app:fold-theory}: an incorrect attracting branch meets an unstable companion and both cease to exist locally as the input interpolation crosses the critical value. The verified coefficient signs place the attracting branch on the lower-$\lambda$ side.

The complete certificates cover 42 solution-consistent and nine inconsistent hypotheses. They establish local elimination as an attainable outcome of input conditioning in learned networks. We report the certified instances and their subsequent reasoning outcomes first, followed by the model, selection, and verification protocols.

\begin{table}[t]
\centering
\small
\setlength{\tabcolsep}{4pt}
\begin{tabular}{ccrrrrrr}
\toprule
& & \multicolumn{4}{c}{Solution-consistent hypotheses}
& \multicolumn{2}{c}{Inconsistent hypotheses}\\
\cmidrule(lr){3-6}\cmidrule(lr){7-8}
Width & Seed & Tested & Folds & Roots & Complete & Tested & Complete\\
\midrule
8 & 0 & 71 & 21 & 19 & 19 & 9 & 0\\
8 & 1 & 56 & 1 & 1 & 1 & 24 & 3\\
8 & 2 & 61 & 13 & 12 & 12 & 19 & 4\\
16 & 0 & 23 & 6 & 6 & 3 & 17 & 0\\
16 & 1 & 45 & 7 & 7 & 7 & 35 & 2\\
16 & 2 & 8 & 0 & 0 & 0 & 8 & 0\\
\midrule
\multicolumn{2}{l}{Total} & 264 & 48 & 45 & 42 & 112 & 9\\
\bottomrule
\end{tabular}
\caption{Fold detection and certification in six trained small models. Counts concern selected checkpoint--instance--hypothesis pairs, not independent puzzles. All nine detected folds under inconsistent hypotheses have complete certificates. Failure to complete a certificate does not establish absence of a fold.}
\label{tab:small-model-folds}
\end{table}

\begin{figure}[t]
\centering
\includegraphics[width=\linewidth]{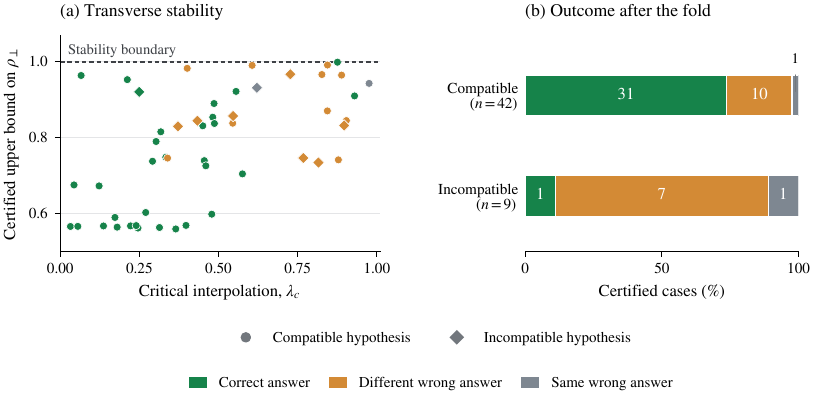}
\caption{Certified local folds and subsequent warm-start outcomes in trained small models. (a) Fold locations and certified bounds on the noncritical spectral radius for all 51 complete certificates; markers distinguish solution-consistent and inconsistent hypotheses, and colors indicate the 512-update endpoint category after crossing the fold. Each bound concerns the noncritical spectrum; the critical multiplier itself equals one. (b) Endpoint categories at $\lambda=\min\{1,\lambda_c+0.02\}$: 42 consistent and nine inconsistent hypotheses. Bar widths show percentages and labels give case counts. ``Same wrong answer'' denotes output agreement with the initial failure, not return to the same latent state.}
\label{fig:certified-folds}
\end{figure}

\paragraph{Recovery after a certified fold.}
Among the 42 certified cases with solution-consistent hypotheses, 31 produce the correct solution in the post-fold warm-start diagnostic (Figure~\ref{fig:certified-folds}). We initialize each rollout at the original incorrect state and run 512 updates at $\lambda=\min\{1,\lambda_c+0.02\}$. Ten of the remaining cases end at another incorrect decode and one at the same incorrect decode; the latter latent state has nevertheless moved and remains nonstationary.

Among the nine inconsistent cases, one ends at the solution, seven at another incorrect decode, and one at the same incorrect decode. At $\lambda=1$, their counts are one, six, and two, respectively. These are endpoint outcomes, not the frequency of ever visiting a correct prediction. The results separate the certified change to the local fixed-point structure from the empirically measured answer reached by subsequent recursion.

\paragraph{Models and diagnostic data.}
We use $4\times4$ Sudoku with $L=16$ positions and hidden widths $h\in\{8,16\}$, each trained with three seeds. Both recurrent components are retained, giving full state dimensions $n=2Lh\in\{256,512\}$. The shared MLP block uses position-mixing and channelwise SwiGLU transformations with RMS normalization; each recursion step updates $z_L$ and then $z_H$. Training uses cross-entropy supervision at eight recursion steps, detached recurrent carries, and AdamW with batch size 256, learning rate $10^{-3}$, weight decay $1$, and 4,000 optimizer updates.

Of the 288 completed Sudoku grids, 200 supply training puzzles and 88 supply the evaluation pools. Puzzles are generated by deleting cells and retaining only uniquely solvable inputs. Checkpoints are selected by 32-step accuracy on the standard pool with 6--10 givens, evaluated every 250 optimizer updates. This pool therefore serves model selection and is not an unbiased test set. The mechanism study uses a separate pool of 3,520 harder puzzles with 4--5 givens, generated from the same 88 grids.

\paragraph{Case and hypothesis selection.}
For each checkpoint, we scan this pool in its fixed order, run 512 updates from initialization, and stop after collecting ten incorrect near-fixed points or exhausting the pool. Near-stationarity requires relative residuals below $10^{-10}$ throughout the last 16 updates. An incorrect output differs from the unique reference completion; merely satisfying row, column, and box constraints without respecting the original givens is insufficient. The six scans yield 47 retained checkpoint--instance pairs.

Each retained state receives eight single-cell hypotheses. The first inserts the reference value at an incorrectly predicted empty cell; remaining proposals are ranked by decoder logit margin. We distinguish hypotheses that agree with the unique solution (solution-consistent, or compatible) from those that disagree (inconsistent, or incompatible). This reference-informed construction is a diagnostic control, separate from the label-free ICS performance evaluation.

\paragraph{Certificate construction and replay.}
For each pair, we perform dense full-state pseudo-arclength continuation, refine detected folds, and attempt interval certification at multiple box radii. We separately count detected folds, verified roots of the augmented fold equations, and complete certificates. A complete certificate additionally establishes nondegenerate fold coefficients, a simple unit multiplier with the complementary spectrum inside the unit circle, and a strictly wrong prediction at an original non-given position throughout the enclosing box. Appendix~\ref{app:fold-theory} states the local implication and verification conditions. The certificate is for the real-valued map defined by the stored weights, with numerical enclosures accounting for evaluation error.

We replay all 54 saved candidate certificates without repeating candidate search. All 54 pass augmented-root, nondegeneracy, and wrong-decoding verification; 51 also pass transverse-stability verification. The remaining three are unverified at this last stage, not classified as unstable. The replay uses outward-rounded interval arithmetic, independently checks the decoder-margin lower bounds with exact rational arithmetic, and checks uniqueness of the original puzzles.

\paragraph{A representative certificate.}
For a representative width-16, seed-0 case (instance 305, hypothesis at cell 0), $\lambda_c\approx0.482176$. The verified root bound satisfies $Y+ZR<1.049\times10^{-9}<R$, with $R\approx1.737\times10^{-9}$. Outward-widened coefficient enclosures are $a\in[-0.27288,-0.27215]$ and $b\in[-0.10222,-0.10202]$, while the noncritical spectral radius is below $0.85388$. Four original non-given positions have certified wrong tokens, with all their decoder margins exceeding $7.21535$. This example verifies both transverse stability and wrong decoding around the critical point.

\paragraph{Scope and interpretation.}
The theorem and certificates establish that changing the input embedding can locally eliminate an incorrect attracting fixed point. The large-model continuation results are numerical evidence; the rigorous certificates concern the separately trained small models. Certification covers a neighborhood of each critical point along the specified interpolation, rather than absence of every incorrect attractor or absence at the final condition $\lambda=1$. The connection from the original failed state to the critical branch is tracked numerically, not certified along its entire length.

The study selects failed states and hypotheses for mechanism analysis, so the case counts do not estimate the prevalence of this mechanism on the main test sets. Post-fold rollouts evaluate recovery separately from local elimination. Their successful and unsuccessful outcomes, together with the persistent branches above, motivate searching over input conditions and evaluating the candidates that those conditions produce.

\subsection{Restatement-induced dynamics in aligned coordinates}
\label{app:restatement-dynamics}

Validity-preserving restatements express the same task in different
representations. We test whether their effect extends beyond a change
of coordinates: after aligning latent states, do they change the
recursive update, and does the resulting recovery depend on
initialization and continued conditioning?

\paragraph{Aligning the update maps.}
We restrict this study to spatial restatements. For a cell permutation
$\Phi$, let $P_\Phi$ apply the corresponding permutation to the
sequence positions of both recurrent components $(z_H,z_L)$, while
leaving hidden-feature coordinates and the 16 puzzle-embedding prefix
positions unchanged. We express the restated-input update in the
original coordinates as
\begin{equation}
    \widetilde F_\Phi(z)
    = P_\Phi^{-1}F_{\Phi(x)}(P_\Phi z).
    \label{eq:restatement-aligned-map}
\end{equation}
The tested TRM models initialize each recurrent component by
broadcasting a shared vector over positions, so $P_\Phi z_0=z_0$;
the implementation checks this identity for every batch. Their
shared token decoder permits predictions to be mapped back by
$\Phi^{-1}$. Thus, the original and restated cold-start trajectories
begin at the same aligned state. Equivariance of the recursive update
would additionally require $\widetilde F_\Phi=F_x$, which we do not
assume.

\paragraph{Models, instances, and restatements.}
We use a frozen TRM checkpoint for Sudoku-Extreme and
a frozen checkpoint for Maze-Hard. The archived
Sudoku failure cohort contains 4,320 instances: 239 in the incorrect
near-fixed-point class, 4,073 moving failures, two periodic failures,
and six other or unresolved cases. The Maze cohort contains 42
incorrect near-fixed-point instances. For each instance, we use its
saved full latent endpoint $z_{\mathrm{fail}}$ from the trajectory
diagnostics in Appendix~\ref{app:latent-trajectories}.

For Sudoku, we sample eight nonidentity spatial symmetries per
instance by composing band and stack permutations, row permutations
within bands, column permutations within stacks, and optional
transposition. Sampling uses the fixed per-instance seed
$7\cdot1{,}000{,}003+\mathrm{id}(x)$; duplicate sampled symmetries are
not removed. Digit relabeling is excluded from this study because
Equation~\ref{eq:restatement-aligned-map} defines a transport of
spatial positions. For Maze, we use all seven nonidentity rotations
and reflections of the grid. These yield 34,560 Sudoku and 294 Maze
instance--restatement pairs. They are diagnostic cohorts from one
checkpoint per task, rather than full-test, multi-seed evaluations
of ICS.

\paragraph{Paired rollouts and scoring.}
For each pair $(x,\Phi)$, we run $\widetilde F_\Phi$ for 512 recursion
steps from either $z_0$ (\emph{cold}) or $z_{\mathrm{fail}}$
(\emph{warm}). Original-input controls run $F_x$ for 512 steps from
the same two starting states. All rollouts use eager FP32 inference
with frozen parameters, no injected latent noise, and no early
stopping or model-based candidate selection.

At every step, predictions are mapped to the original representation
and passed through the task-specific output projection. Sudoku
scoring restores the original givens and checks agreement with the
unique reference solution. Maze scoring restores the original
structural tokens and applies the shortest-path validity criterion
to the projected output. These operations affect the scored
prediction, not the latent recurrence. Recovery records whether
any scored prediction is correct within the observation window.
We report both recovery averaged over instance--restatement pairs
and instance-level coverage by any of the $K$ restatements. The
original-input controls have one rollout per instance and start.
Confidence intervals use 1,000 bootstrap resamples of instances,
retaining all associated restatements and paired conditions.

\paragraph{Restatements change the aligned update.}
At a shared state $z$, define
\begin{equation}
\begin{aligned}
    \Delta_x(z)&=F_x(z)-z,
    &\Delta_\Phi(z)&=\widetilde F_\Phi(z)-z,\\
    d_\Phi(z)&=
    \frac{\|\Delta_\Phi(z)-\Delta_x(z)\|_2}
         {\max\{\|\Delta_x(z)\|_2,10^{-12}\}}.
\end{aligned}
\label{eq:restatement-update-change}
\end{equation}
Norms include both recurrent components and their prefix positions.
Across the failed-instance pairs, the median $d_\Phi(z_0)$ is
0.259 for Sudoku and 0.482 for Maze; the corresponding median
cosine similarities between updates are 0.966 and 0.884.
The updates therefore differ even after undoing the spatial
coordinate transformation.

At original near-fixed points, $\|\Delta_x(z)\|_2$ is very small,
so we instead compare the relative step residual
$\|F(z)-z\|_2/\|z\|_2$ under the two maps. For the 239 Sudoku
near-fixed-point instances, its median increases from
$6.02\times10^{-7}$ under $F_x$ to $0.0494$ under
$\widetilde F_\Phi$. For the 42 Maze instances, it increases from
$9.22\times10^{-7}$ to $0.672$. Thus, these recorded states cease
to be nearly stationary under the restated maps. This observation
does not by itself establish recovery or removal of an entire
incorrect attractor.

\begin{figure}[t]
    \centering
    \includegraphics[width=\linewidth]{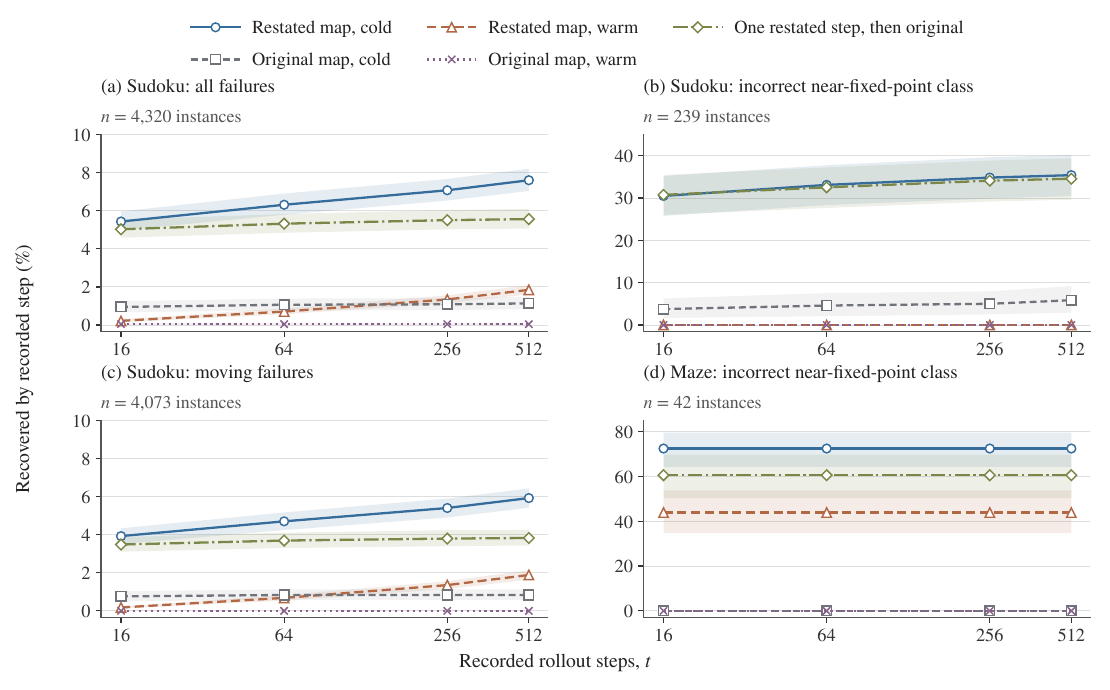}
    \caption{\textbf{Recovery under spatial restatements depends on
    initialization and the duration of conditioning.}
    Recovery is the fraction of instance--restatement pairs with
    a correct prediction by the recorded rollout step; original-map
    controls average over instances. Bands show pointwise 95\%
    instance-bootstrap intervals. Each restated cold/warm rollout
    uses the same task and transformation, differing only in its
    initial state. The one-step control first applies
    $\widetilde F_\Phi$ once, then records $t$ updates under $F_x$:
    its total cost is $t+1$ steps, and the preparatory state is not
    included in its recovery window. Panel $y$-axis ranges differ.}
    \label{fig:restatement-dynamics}
\end{figure}

\begin{table}[t]
\centering
\small
\setlength{\tabcolsep}{4pt}
\resizebox{\linewidth}{!}{%
\begin{tabular}{lrrrr}
\toprule
 & \multicolumn{3}{c}{Sudoku-Extreme} & Maze-Hard\\
\cmidrule(lr){2-4}
Condition & All failures & Near-fixed & Moving & Near-fixed\\
Instances & 4,320 & 239 & 4,073 & 42\\
\midrule
Original map, cold & 1.13 & 5.86 & 0.83 & 0.00\\
Original map, warm & 0.05 & 0.00 & 0.00 & 0.00\\
Restated map, cold & 7.60 & 35.41 & 5.92 & 72.45\\
Restated map, warm & 1.85 & 0.00 & 1.88 & 43.88\\
One restated step, then original$^{\dagger}$
                  & 5.56 & 34.57 & 3.83 & 60.54\\
\midrule
Any-of-$K$ coverage, restated cold
                  & 19.58 & 53.97 & 17.53 & 92.86\\
Any-of-$K$ coverage, restated warm
                  & 7.82 & 0.00 & 8.20 & 85.71\\
\bottomrule
\end{tabular}}
\caption{Recovery and coverage (\%) on the archived failure cohorts.
The upper rows report recovery within 512 recorded updates; restated
conditions average over instance--restatement pairs, and original-map
controls average over instances. The final two rows count instances
recovered by at least one of $K=8$ Sudoku or $K=7$ Maze restatements.
$\dagger$: this control uses one preparatory restated update followed
by 512 original-map updates, for 513 total steps.}
\label{tab:restatement-recovery}
\end{table}

\paragraph{Changing the map does not remove the dependence on initialization.}
Figure~\ref{fig:restatement-dynamics} and
Table~\ref{tab:restatement-recovery} show that cold restatements recover
35.41\% of Sudoku near-fixed-point pairs, whereas warm restatements
recover none. On Maze, both initializations recover solutions, but
cold recovery is higher: 72.45\% versus 43.88\%. The paired cold-minus-warm
difference is 35.41 percentage points for Sudoku near-fixed-point
pairs (95\% interval: 30.49--40.17) and 28.57 points for Maze
(21.08--36.05). Allowing any tested restatement to succeed yields
cold-start instance coverage of 53.97\% and 92.86\%, respectively.
In particular, the increased residual at a Sudoku failed state
does not imply access to a correct solution from that state.

\paragraph{One changed update versus continued restatement.}
To distinguish the first changed update from continued use of the
restated map, we initialize a control at
$z_1=\widetilde F_\Phi(z_0)$ and then apply $F_x$ for 512 additional
steps. As implemented, this control therefore has 513 total steps;
only its 512 original-map predictions enter the recorded recovery
window. It is a mechanism control, not an exactly matched-compute
comparison.

For Sudoku near-fixed-point pairs, this control recovers 34.57\%,
close to the 35.41\% obtained with continued restatement. The paired
difference is 0.84 percentage points (95\% interval:
$-0.42$--$2.25$). Continued restatement has a larger advantage on
Sudoku moving failures, 2.10 points (1.83--2.38), and on Maze,
11.90 points (5.10--19.06). Thus, a single restated update can
provide much of the recovery observed in the Sudoku near-fixed-point
cohort, whereas retaining the restated map provides additional
benefit on Maze. The benefit of restatement need not require
continuous use of the alternative map in every setting.

\paragraph{Controls starting from solved instances.}
We also evaluate 444 Sudoku and 38 Maze instances that produce valid
endpoints in a separate 512-step original-input screening rollout.
We compare restated cold starts with restated warm starts from these
saved valid endpoints, using the same transformation families and
a further 512 steps. Table~\ref{tab:restatement-solved} reports
endpoint invalidity, rather than whether an intermediate solution
was encountered. Original-input cold reruns are included because
screening and evaluation are separate executions. The results show
that successful task representations are not interchangeable for
the learned solver: cold restatements can return invalid outputs
on previously solved instances, while continuation from a valid
state usually preserves validity.

\begin{table}[t]
\centering
\small
\setlength{\tabcolsep}{5pt}
\begin{tabular}{lrr}
\toprule
Condition & Sudoku-Extreme & Maze-Hard\\
\midrule
Original map, cold & 0.45\% (2/444) & 0.00\% (0/38)\\
Restated map, cold & 0.53\% (19/3,552) & 4.89\% (13/266)\\
Restated map, warm & 0.00\% (0/3,552) & 0.38\% (1/266)\\
\bottomrule
\end{tabular}
\caption{Invalid final predictions after 512 updates on the
separately screened solved-control cohorts. Counts are
invalid/total rollouts. Restated rows contain eight transformations
per Sudoku instance and seven per Maze instance.}
\label{tab:restatement-solved}
\end{table}

These experiments establish that spatial restatements can change
the aligned recursive update and redirect reasoning from the same
initial state without changing model parameters or task validity.
Together with the initialization and one-step controls, they support
restatements as an input-conditioning operation within ICS while
showing that their effectiveness depends on both the starting state
and how long the condition is retained.

\subsection{Initialization and hypothesis retention}
\label{app:retention}

We examine whether a hypothesis that enables recovery from fresh initialization also enables recovery from a saved incorrect state, and how recovery depends on the duration of conditioning.

\paragraph{Models, instances, and hypotheses.}
We study three TRM configurations: Sudoku MLP, Maze attention, and Sudoku attention, using one checkpoint per configuration. For each model, we use a fixed cohort of 20 incorrect instances: 16 with a successful hypothesis in an earlier cold-start screening experiment and four without such a success.
Each instance receives eight model-proposed single-cell hypotheses derived from its saved state's decoder logits. The same hypotheses are used in every initialization and retention condition.

These selected cohorts support paired mechanism comparisons rather than estimates of test-set accuracy.

\paragraph{Rollout and correctness conventions.}
All rollouts use frozen parameters, FP32 state updates, and disabled halting.  Correctness is always evaluated against the original
task $x$.

Cold-start trajectories apply $F_{x_H}$ from the standard initialization:
\begin{equation}
    z_t^{\mathrm{cold},H}=F_{x_H}^{\,t}(z_0).
\end{equation}
We run each trajectory for 512 updates and record endpoint correctness at intermediate horizons.

Warm-start trajectories begin from the same saved incorrect state $w$. They retain the hypothesis for $m$ updates and then restore the original input:
\begin{equation}
    z_T^{H,m}
    =
    F_x^{\,T-m}\!\left(F_{x_H}^{\,m}(w)\right),
    \qquad T=512,
    \quad m\in\{0,1,2,16,64,512\}.
    \label{eq:retention-appendix}
\end{equation}
The latent state is carried continuously across the input switch.
At $m=0$, the entire rollout uses the original input; at $m=512$, the hypothesis remains active throughout. The original-input control is evaluated once per instance and shared across its hypotheses.

\paragraph{Conditional and full-pool comparisons.}
Within each model's 16-instance screening cohort, there are 128 instance--hypothesis pairs. Of these, 37 for Sudoku MLP, 25 for Maze attention, and 26 for Sudoku attention are strictly correct after 64 cold-start updates. These pairs define the \emph{cold-successful} subset.

\begin{figure}[t]
\centering
\includegraphics[width=\linewidth]{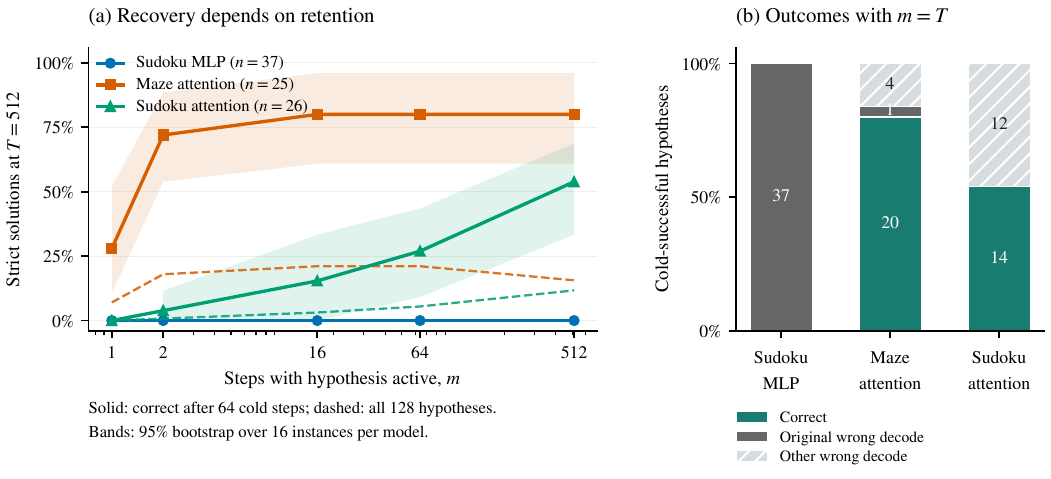}
\caption{
Input retention has model-dependent effects. (a) Strict correctness after 512 updates from a saved incorrect state. Solid curves: hypotheses successful after 64 cold-start updates; dashed curves: all 128 hypotheses from the 16-instance screening cohort. Bands show nondegenerate 95\% instance-bootstrap intervals. (b) Endpoint decode categories among cold-successful hypotheses when conditioning is retained throughout ($m=T$).
}
\label{fig:mechanisms}
\end{figure}

Figure~\ref{fig:mechanisms} and Table~\ref{tab:mechanism-retention} report warm-start outcomes for both this conditional subset and the full 128-pair pool. Thus, the conditional comparison holds the instance and hypothesis fixed while testing recovery from a different initialization. All warm-start results are scored at the same total horizon of 512 updates.

\begin{table}[t]
\centering
\small
\setlength{\tabcolsep}{4pt}
\begin{tabular}{llccccc}
\toprule
Model & Hypothesis pool
& $m=1$ & $m=2$ & $m=16$ & $m=64$ & $m=512$\\
\midrule
Sudoku MLP
& All
& $0/128$ & $0/128$ & $0/128$ & $0/128$ & $0/128$\\
& Cold-successful
& $0/37$ & $0/37$ & $0/37$ & $0/37$ & $0/37$\\
\midrule
Maze attention
& All
& $9/128$ & $23/128$ & $27/128$ & $27/128$ & $20/128$\\
& Cold-successful
& $7/25$ & $18/25$ & $20/25$ & $20/25$ & $20/25$\\
\midrule
Sudoku attention
& All
& $0/128$ & $1/128$ & $4/128$ & $7/128$ & $15/128$\\
& Cold-successful
& $0/26$ & $1/26$ & $4/26$ & $7/26$ & $14/26$\\
\bottomrule
\end{tabular}
\caption{
Strictly correct warm-start endpoints after 512 updates. The hypothesis is retained for the first $m$ updates. Each model's full pool contains eight hypotheses for each of 16 screened instances. Cold-successful subsets are selected by correctness after 64 updates from fresh initialization. Denominators count instance--hypothesis pairs, not distinct instances.
}
\label{tab:mechanism-retention}
\end{table}

\paragraph{Initialization changes recoverability.}
For Sudoku MLP, all 37 cold-successful hypotheses also remain correct at the cold-start 512-step endpoint. Yet none recovers from the saved incorrect state at any tested retention duration. All 128 warm-start pairs retain the original incorrect endpoint decode. Thus, a hypothesis can support a correct solution from fresh initialization while failing to recover from an existing incorrect state under the same conditioned map.

\paragraph{The effect of retention varies across models.}
Maze attention often recovers after brief conditioning, whereas Sudoku attention benefits from longer conditioning within the tested schedules. Longer retention is not uniformly beneficial: in the full Maze pool, restoring the original input after 16 or 64 updates yields 27 correct endpoints, compared with 20 when the hypothesis remains active throughout.

Because the total horizon is fixed, increasing $m$ also reduces the time available after restoring $x$. These curves therefore compare complete intervention schedules, rather than identifying a universal minimum conditioning duration.

\paragraph{Additional controls and uncertainty.}
Without changing the input, all 60 incorrect instances retain their original incorrect endpoint decode. Among the four additional instances per model, neither Sudoku configuration yields a warm-start success at any tested duration. Maze yields one success among 32 hypotheses at $m=16$ and $m=64$, and none at the other durations.

For uncertainty estimates in the retention curves, we resample instances with replacement using 2,000 bootstrap replicates and recompute the pooled hypothesis success rate. This preserves dependence among hypotheses from the same instance.

\paragraph{Longer continuations.}
We selectively extend a retention trial to 2,048 updates if its maximum relative update over the last 16 steps exceeds $10f_{\mathrm{num}}$, or if its endpoint classification remains unresolved, where $f_{\mathrm{num}}=2.5\times10^{-7}$. Continuation uses the map active at step 512.

There are 820 incorrect-instance trials per model: one original-input control and 40 hypothesis-duration combinations for each of 20 instances. The extension criterion is triggered by zero trials for Sudoku MLP, 28 for Maze attention, and 144 for Sudoku attention. These follow-up diagnostics are kept separate from the common 512-step outcomes reported above.

\subsection{Matched-norm perturbations and solution round trips}
\label{app:matched-and-roundtrip}

We use two additional controls to examine the effect of input conditioning. The first compares a single conditioned update with random state displacements of the same magnitude. The second tests whether correct solutions can be recovered after switching between the original and hypothesis-conditioned inputs. Both controls use the strict correctness criterion.

\paragraph{Matching the first conditioned displacement.}
For each incorrect instance, we select the hypothesis with the largest halting score among those strictly correct after 64 cold-start updates. If none is correct, we select the hypothesis with the largest score. This selection uses cold-rollout outcomes to define the diagnostic comparison.

For the selected hypothesis $H$, let
\begin{equation}
    d_H=\|F_{x_H}(w)-w\|_2
\end{equation}
be the magnitude of its first update from the saved incorrect state $w$.
We draw eight independent Gaussian vectors and normalize them to obtain isotropic unit directions:
\begin{equation}
    u_k=\frac{\xi_k}{\|\xi_k\|_2},
    \qquad
    \xi_k\sim\mathcal{N}(0,I_d),
    \qquad k=1,\ldots,8.
\end{equation}
The matched random interventions and the one-step input intervention are then compared through
\begin{equation}
\begin{aligned}
    z_T^{\mathrm{random},k}
    &=F_x^{\,T-1}(w+d_Hu_k),\\
    z_T^{\mathrm{input}}
    &=F_x^{\,T-1}\!\left(F_{x_H}(w)\right),
    \qquad T=512.
\end{aligned}
\end{equation}
Thus, both conditions use 511 original-input updates after the intervention. The matched quantity is the Euclidean displacement
of the complete recurrent state.

Across the 16 screened instances per model, none of the 128 random-direction continuations produces a strictly correct prediction, for any of the three models.
The selected one-step input intervention succeeds on 6 of 16 Maze instances, and on none of the Sudoku instances.
The additional four-instance cohorts also yield
no random-intervention successes, with 32 trials
per model. On the sampled Maze instances, matching displacement
magnitude alone therefore does not reproduce the
recovery induced by a conditioned update.

\paragraph{Solution round trips.}
We next examine correspondence between solution states under the two input conditions. Using one recorded hypothesis per screened instance, we construct the endpoints
\begin{equation}
\begin{aligned}
    s_H &= F_{x_H}^{\,2048}(z_0),\\
    s_x &= F_x^{\,2048}(s_H),\\
    \widetilde{s}_H &= F_{x_H}^{\,512}(s_x).
\end{aligned}
\end{equation}
The first two endpoints are saved from the preceding
long-rollout experiment. The final phase reapplies the same hypothesis while carrying the state continuously from $s_x$.

For the solution-preservation comparison, we select pairs for which both $s_H$ and $s_x$ are strictly correct and satisfy the numerical near-fixed-point classification under their respective maps. This yields 16 Sudoku MLP, 15 Maze attention, and seven Sudoku attention instances.

We separately measure whether $\widetilde{s}_H$ is strictly correct, whether it has the same raw decode as $s_H$, and whether it returns to the recorded latent neighborhood of $s_H$. 
Let $S_H$ contain $s_H$ and the final 16 states of its reference trajectory. The latent-return criterion is
\begin{equation}
    \min_{s\in S_H}\|\widetilde{s}_H-s\|_2
    \leq
    \max\!\left\{
        10f_{\mathrm{num}}\|s_H\|_2,\,
        \operatorname{diam}(S_H)
    \right\},
    \qquad f_{\mathrm{num}}=2.5\times10^{-7}.
\end{equation}
Distances are computed in the full latent space.

\begin{table}[t]
\centering
\small
\setlength{\tabcolsep}{7pt}
\begin{tabular}{lccc}
\toprule
Model
& Strictly correct
& Same decode
& Latent return\\
\midrule
Sudoku MLP
& $16/16$ & $16/16$ & $16/16$\\
Maze attention
& $15/15$ & $15/15$ & $15/15$\\
Sudoku attention
& $7/7$ & $7/7$ & $5/7$\\
\bottomrule
\end{tabular}
\caption{
Solution round trips. The selected pairs have strictly correct, numerically near-fixed endpoints under both input conditions before the final phase. Results describe $\widetilde{s}_H$ after reapplying the hypothesis for 512 updates. Same-decode and latent-return measurements use $s_H$ as their reference. Each instance contributes one trial.
}
\label{tab:solution-roundtrips}
\end{table}

All selected round trips recover a strictly correct endpoint with the same decoded answer as $s_H$ (Table~\ref{tab:solution-roundtrips}). Latent return also holds for every selected Sudoku MLP and Maze instance, but for only five of the seven Sudoku attention instances. Correct-answer preservation and return to a recorded latent neighborhood are therefore distinct outcomes.

The remaining saved pairs are reported separately. The one remaining Maze pair produces a correct endpoint and returns to its reference neighborhood. Among nine remaining Sudoku attention pairs, one produces a correct endpoint and seven return to their reference neighborhoods. Because these reference pairs do not all represent correct solutions, their return rates do not measure
solution preservation.

Together with the initialization comparison, these results distinguish preserving an available solution from recovering one from an incorrect state. In particular, Sudoku MLP can recover the recorded solution after an input-condition round trip even when the same family of hypotheses fails to recover from saved incorrect states.

\subsection{One-step response to hypothesis injection}
\label{app:one-step-response}

We measure how input conditioning changes the next update from a saved incorrect state. Using the same models, instances, and hypotheses as Section~\ref{app:retention}, we compare one update under $F_x$ with one under $F_{x_H}$, holding the initial latent state and model parameters fixed.

We use the relative update residuals defined in Section~\ref{app:landscape-construction}:
\begin{equation}
    r_x(w)=\frac{\|F_x(w)-w\|_2}{\|w\|_2},
    \qquad
    r_{x_H}(w)=
    \frac{\|F_{x_H}(w)-w\|_2}{\|w\|_2}.
\end{equation}
The norm includes both recurrent components of the complete latent state.

\begin{table}[t]
\centering
\small
\setlength{\tabcolsep}{10pt}
\begin{tabular}{lcc}
\toprule
Model
& Original input
& Hypothesis input\\
& $r_x(w)$ & $r_{x_H}(w)$\\
\midrule
Sudoku MLP
& $1.92\times10^{-7}$ & $2.80\times10^{-2}$\\
Maze attention
& $2.37\times10^{-7}$ & $6.67\times10^{-2}$\\
Sudoku attention
& $4.75\times10^{-7}$ & $1.65\times10^{-1}$\\
\bottomrule
\end{tabular}
\caption{
Median relative one-step displacement from saved incorrect states. Each model contributes 20 instances. The original-input column uses one measurement per instance; the hypothesis-input column pools eight hypotheses per instance, giving 160 state--hypothesis pairs without filtering by cold-start success or halting score. Both quantities are dimensionless and normalized by the same starting state's norm.
}
\label{tab:mechanism-forward}
\end{table}

Under the original input, median relative updates are of order $10^{-7}$. Under hypothesis conditioning, they increase to $2.80\%$, $6.67\%$, and $16.50\%$ for Sudoku MLP, Maze attention, and Sudoku attention, respectively (Table~\ref{tab:mechanism-forward}). Thus, changing the input produces percent-level movement from states that are nearly stationary under the original map.

This response does not necessarily yield a correct answer. For example, Sudoku MLP retains its original incorrect endpoint decode throughout the tested retention schedules despite the initial displacement. Persistent incorrect predictions therefore need not imply an absence of latent movement. The initialization and retention experiments examine whether that movement ultimately enables recovery.

\section{Training Supervision and Inference Behavior}
\label{app:training-interventions}

\begin{figure}[t]
\centering
\includegraphics[width=\linewidth]{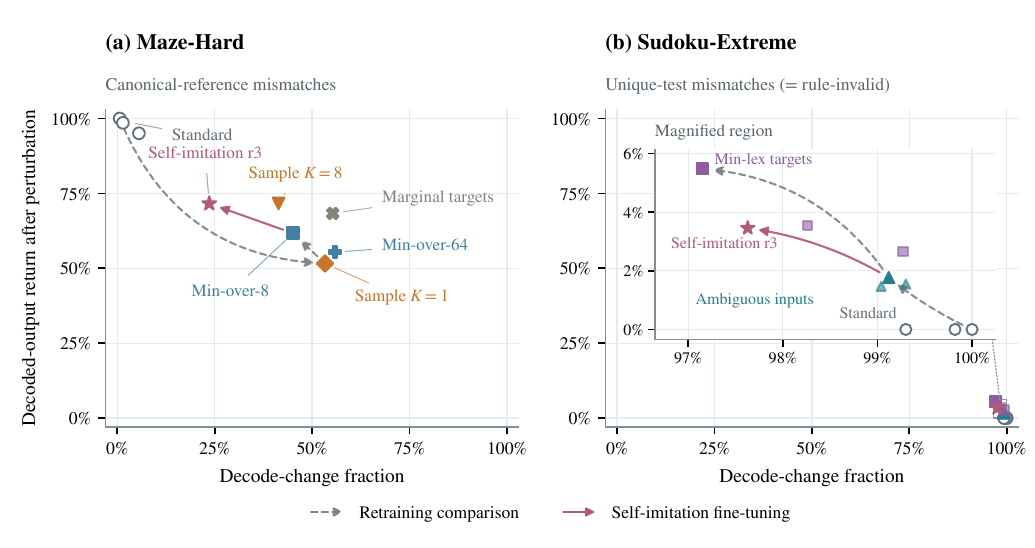}
\caption{
\textbf{Training supervision changes decoded behavior.}
Each point summarizes one trained model's reference-mismatch
cohort after 16 recursion steps. The horizontal axis measures
subsequent decoded changes; the vertical axis measures return
to the original decoded prediction after perturbation.
Dashed arrows compare separately trained conditions, whereas
solid arrows connect a self-imitation initialization to its
third fine-tuning round. The Sudoku inset magnifies the same
observations. Standard Sudoku models are probed on 3,072 test
instances and retrained models on 422,786; Maze models use
the same 1,000-instance test split. Maze reference mismatches
can include valid alternative solutions.
}
\label{fig:training-interventions}
\end{figure}

\begin{figure}[t]
\centering
\includegraphics[width=\linewidth]{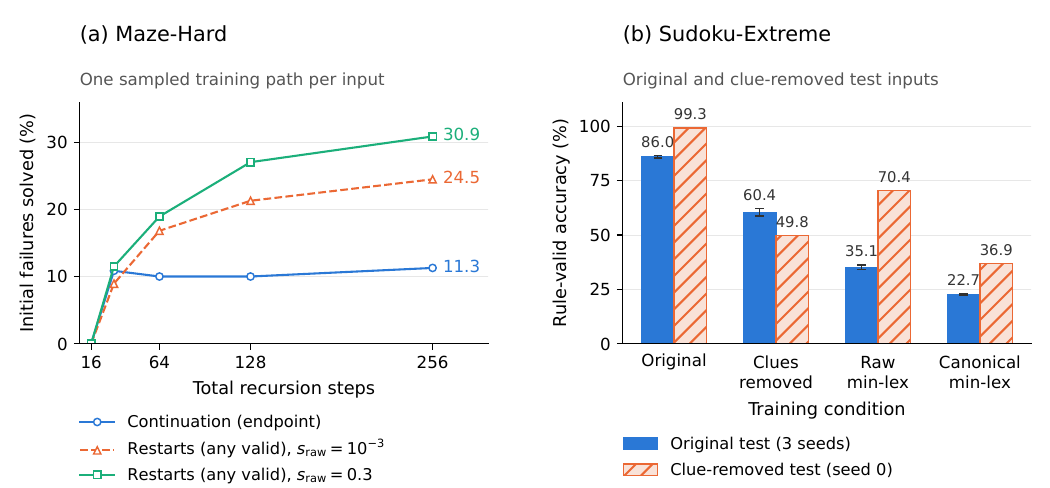}
\caption{
\textbf{Task performance under training interventions.}
(a) Additional inference on the 469 initially rule-invalid
instances of the Maze Sample $K=1$ model.
Continuation measures correctness at the budget endpoint;
restart measures coverage across trial endpoints.
Noise scales are standard deviations in raw state units.
(b) Sudoku rule-valid accuracy under changes to training
inputs and target conventions.
Original-test bars show three-seed means and standard
deviations; clue-removed results use seed~0.
}
\label{fig:training-performance}
\end{figure}

Our finite-horizon diagnostics reveal a marked contrast between the studied Sudoku-Extreme and Maze-Hard models: incorrect predictions on Sudoku frequently continue to change under further recursion, whereas those on Maze are predominantly persistent. Nurikabe, Tapa, and Heyawake exhibit mixtures of these behaviors.

One potential contributor is the relationship between solution multiplicity and training supervision. The original Sudoku-Extreme training and test puzzles have unique solutions. Maze-Hard, by contrast, admits multiple shortest paths, while each training instance is supervised using a single reference path selected by the dataset's path-generation procedure.

These observations motivate two related questions: does solution multiplicity affect inference behavior, and does the way supervision selects among valid solutions matter? We investigate these questions through complementary training interventions. On Maze-Hard, we keep the training inputs fixed and change how shortest-path targets are selected or combined. On Sudoku-Extreme, we introduce multiple solutions by removing givens and then vary the target-selection convention. We measure both decoded behavior and task accuracy
(Figures~\ref{fig:training-interventions}
and~\ref{fig:training-performance}).

\subsection{Maze-Hard: changing supervision on fixed inputs}

We first investigate whether the choice of training targets affects inference behavior independently of solution multiplicity. We retain the same 1,000 Maze-Hard training inputs, so each input's set of valid shortest paths remains unchanged, and vary how these paths enter supervision. All models are evaluated on the unchanged 1,000-instance test split.

\paragraph{Replacing the reference-path convention.}
The original dataset provides one reference shortest path per input. In \emph{Sample $K=1$}, we replace this target with one uniformly sampled shortest path, fixed throughout training. Thus, each input still has a single training target, but the targets no longer follow the original path-selection convention. In \emph{Sample $K=8$}, we present eight sampled paths per input as separate training examples, increasing the training set to 8,000 input--target pairs.

Figure~\ref{fig:training-interventions}(a) shows that these changes substantially alter decoded behavior. For the standard models, only $0.63$--$5.56\%$ of their reference-mismatch cohorts meet the decode-change criterion, and $95.06$--$100\%$ return to their original decoded prediction after perturbation. With Sample $K=1$, these fractions become $53.3\%$ and $51.6\%$, respectively; Sample $K=8$ gives $41.4\%$ and $71.6\%$.

These measurements use each model's own reference-mismatch  cohort, which can include valid alternative paths. They demonstrate a change in decoded behavior without changing the underlying puzzles or their solution sets. However, the target replacements also reduce task accuracy: Sample $K=1$ and Sample $K=8$ achieve $53.1\%$ and $23.0\%$ rule-valid accuracy, respectively.

\paragraph{Supervising sets of valid solutions.}
To examine alternatives to presenting different valid paths as separate targets, we train models with \emph{min-over-8} and \emph{min-over-64} objectives. Each input is associated with a sampled target set, and training minimizes cross-entropy over its members. The associated reference-matching supervision also accepts membership in that set. We additionally evaluate per-cell shortest-path inclusion probabilities as soft targets.

Min-over-8 improves rule-valid accuracy to $64.2\%$, while retaining substantial decoded changes ($45.1\%$) and lower decoded-output return ($61.7\%$) than the standard models. Increasing the target set to 64 paths yields $57.5\%$ accuracy, and marginal targets yield $7.8\%$. Thus, providing broader supervision over valid solutions
does not produce a monotonic accuracy improvement.

\paragraph{Learning from the model's own solutions.}
Starting from the min-over-8 checkpoint, we perform three rounds of collecting rule-valid model predictions and fine-tuning on the resulting target sets. Each round collects eight predictions per training input: one deterministic 16-step rollout and seven noisy rollouts. Valid predictions populate the new target set; inputs
without a valid prediction retain their original sampled
targets.

After three rounds, the decode-change fraction decreases from $45.1\%$ to $23.6\%$, while decoded-output return increases from $61.7\%$ to $71.6\%$. Accuracy nevertheless changes from $64.2\%$ to $62.8\%$. Self-imitation therefore moves these behavioral diagnostics toward the standard models without recovering their solution accuracy.

\paragraph{Additional inference after an incorrect prediction.}
We next examine whether additional computation can recover instances that remain incorrect after 16 deterministic recursion steps. For each trained model, we identify these instances using the task rules and compare two inference procedures.

\emph{Continuation} extends the original trajectory to 256 total recursion steps and evaluates its final prediction. 
\emph{Noisy restart} uses the same total budget for the initial 16-step rollout followed by 15 fresh, 16-step trials. During each restart trial, independent Gaussian noise is added to $z_L$ before every update, with per-coordinate standard deviation $s_{\mathrm{raw}}\in\{10^{-3},0.3\}$. This scale is expressed in raw state units, rather than the relative perturbation scale used in the
state-intervention experiments.

Continuation reports correctness at its final step, whereas restart reports whether any trial endpoint is rule-valid. The latter measures candidate coverage and does not require a model-based selector to identify the correct candidate.

For Sample $K=1$, continuation ends with a valid prediction on $53/469=11.3\%$ of the initial failures. Noisy restart provides at least one valid candidate for $115/469=24.5\%$ at $s_{\mathrm{raw}}=10^{-3}$ and $145/469=30.9\%$ at $s_{\mathrm{raw}}=0.3$ (Figure~\ref{fig:training-performance}(a)).

These experiments show that the retrained models can produce valid alternatives to their initial incorrect predictions through noisy rollouts. Together with the behavioral diagnostics, they motivate evaluating training supervision through both its initial solution accuracy and the opportunities it leaves for additional inference.

\subsection{Sudoku-Extreme: introducing solution ambiguity}
The Maze experiments change supervision while preserving each input's solution set. We next take the complementary approach: introduce multiple solutions into Sudoku training inputs and examine whether this produces the predominantly persistent incorrect predictions observed on Maze.

\paragraph{Introducing ambiguity while retaining the targets.}
Starting from the original Sudoku-Extreme training data, we randomly remove givens until each input admits at least two solutions, while retaining its original target. An exact solution-counting procedure, capped at two, verifies this property for all 1,001,000 augmented training examples.

We evaluate the resulting models on both the unchanged unique-solution test set of 422,786 instances and a separate 3,072-instance clue-removed test set. All reported accuracies use task validity, so valid alternative completions are accepted even when they differ from the reference answer.

Training on ambiguous inputs reduces original-test accuracy to $60.38\pm1.72\%$, compared with $86.0\pm0.6\%$ for the standard models. On the clue-removed test set, the seed-zero ambiguity-trained model achieves $49.80\%$, compared with $99.25\%$ for the standard model. Thus, the reduction is also present when evaluation inputs admit multiple solutions.

Despite this change in supervision, incorrect predictions on the original test set continue to change: the decode-change fraction remains approximately $99.0$--$99.3\%$ across three training seeds (Figure~\ref{fig:training-interventions}(b)). Introducing solution ambiguity alone therefore does not reproduce the persistent-output behavior observed on Maze in these experiments.

\paragraph{Selecting targets by a consistent convention.}
We next replace the targets on the same ambiguous training inputs using two deterministic conventions. \emph{Raw min-lex} selects the lexicographically smallest valid completion in row-major order. \emph{Canonical-frame min-lex} compares solutions after relabeling digits by their order of first appearance, with a deterministic raw-digit tie-break for digits absent from the givens.

Raw min-lex supervision raises seed-zero accuracy on the clue-removed test set from $49.80\%$ to $70.38\%$, but reduces original-test accuracy to $35.13\pm1.05\%$. The canonical-frame variant achieves
$22.71\pm0.38\%$ on the original test set and $36.88\%$ on the clue-removed test set (Figure~\ref{fig:training-performance}(b)). Moreover, the raw min-lex models' incorrect predictions on the original test set retain high decode-change fractions of approximately $97.2$--$99.3\%$.

These results show that imposing a deterministic target convention does not, by itself, reproduce Maze-like persistent failures or restore solution accuracy. The particular convention and its interaction with learning remain relevant.

\paragraph{Self-imitation.}
We also fine-tune the seed-zero ambiguity-trained model using its own rule-valid predictions. Each of three rounds uses the first 100,000 ambiguous training examples, replacing an original target with a valid model prediction when one is found and retaining the original target otherwise. Each round performs 8,000 fine-tuning updates.

After three rounds, original-test accuracy is $56.88\%$, compared with $62.24\%$ for the initialization. The decode-change fraction decreases from $99.12\%$ to $97.63\%$, while decoded-output return increases from $1.77\%$ to $3.46\%$. Unlike the larger behavioral shift observed for Maze, this model remains dominated by changing incorrect predictions in the diagnostic window.

\paragraph{What explains the contrast?}
The two directions of intervention distinguish solution multiplicity from the choice of training targets. On Maze-Hard, changing the targets substantially changes decoded behavior while preserving the training inputs and their valid solution sets. Even Sample $K=1$, which retains one fixed target per input, produces a different behavioral profile from the original supervision. The observed behavior therefore depends on more than the number of valid solutions.

Conversely, introducing multiple solutions into Sudoku does not reproduce the predominantly persistent incorrect predictions observed on Maze. Providing deterministic min-lex targets also leaves incorrect predictions largely changing within the diagnostic window. These results identify training supervision as a contributing factor, but do not support a simple correspondence between solution multiplicity, a deterministic target convention, and persistent failures.

The interventions also separate behavioral changes from improvements in reasoning performance. For example, Maze self-imitation increases decoded-output return while slightly reducing solution accuracy. Training objectives should therefore be evaluated by both the quality of the resulting predictions and the effectiveness of additional inference, rather than output persistence alone.

Whereas ICS searches input conditions to change the update map of a frozen reasoner, these experiments motivate a complementary direction: learning update rules whose dynamics support accurate and efficient test-time search.
\section{Conditional Analysis of Input-Conditioned Reasoning}
\label{app:conditional-analysis}

This section establishes sufficient conditions under which input conditioning locally eliminates an incorrect attracting fixed point. The fold analysis in Appendix~\ref{app:fold-theory} gives the proof, and the computer-assisted certificates in Appendix~\ref{app:small-model-certificate} verify these conditions in trained small reasoners. Alongside this result, we characterize the response of a saved state to an input change and give conditions for reaching a correct solution from a chosen initialization.

For a fixed model, let
\[
F = F_\theta(\cdot;x),
\qquad
G_H = F_\theta(\cdot;\tilde{x})
\]
denote the original and hypothesis-conditioned update maps. We begin with the one-step residual response, then prove the local fold criterion and describe its verification. The final two subsections address persistence of a correct fixed point and its accessibility from different initializations.

\subsection{One-step response at an incorrect near-fixed point}
\label{app:one-step-analysis}

Let $w$ be a saved state with an incorrect decoded answer. Using the Euclidean norm of the complete latent state, define
\begin{align}
\eta_F(w) &= \|F(w)-w\|_2,\\
\Delta_H(w) &= \|G_H(w)-F(w)\|_2,\\
\delta_H(w) &= \|G_H(w)-w\|_2.
\end{align}
Here, $\eta_F(w)$ measures the original update residual, $\Delta_H(w)$ measures how much the hypothesis changes the update at the same state, and $\delta_H(w)$ measures the resulting residual under the conditioned map.

\paragraph{A residual bound.}
The identity
\[
G_H(w)-w
=
\bigl(G_H(w)-F(w)\bigr)+\bigl(F(w)-w\bigr)
\]
and the triangle inequality give
\begin{equation}
\bigl|\delta_H(w)-\Delta_H(w)\bigr|
\le \eta_F(w).
\label{eq:input-change-residual}
\end{equation}
Equivalently,
\begin{equation}
\max\{0,\Delta_H(w)-\eta_F(w)\}
\le \delta_H(w)
\le \Delta_H(w)+\eta_F(w).
\end{equation}

If $w$ is an exact fixed point of $F$, then 
$\eta_F(w)=0$ and
$\delta_H(w)=\Delta_H(w)$.
Any nonzero change in the update therefore makes $w$
cease to be a fixed point of $G_H$.

For an approximate fixed point, let $\tau$ be an absolute residual tolerance. If
\begin{equation}
\eta_F(w)\le\tau,
\qquad
\Delta_H(w)>\eta_F(w)+\tau,
\end{equation}
then $\delta_H(w)>\tau$: the same state no longer satisfies the approximate-stationarity criterion under the conditioned map.
For relative residuals, the same bounds hold after dividing all three quantities by the common denominator
$\|w\|_2>0$.

\paragraph{Implication for hypothesis injection.}
When the original residual is small relative to the input-induced update change, Equation~\ref{eq:input-change-residual} implies that
$\delta_H(w)$ closely tracks $\Delta_H(w)$. This explains how hypothesis injection can initiate motion from a previously near-stationary state without directly perturbing that state.

The bound concerns departure from stationarity at $w$, rather than escape from an entire region of incorrect predictions. The conditioned trajectory may still converge to another state that decodes to an incorrect answer. Determining whether a correct solution persists under $G_H$, and whether a particular initialization reaches it, requires additional conditions.

\subsection{Local annihilation of an incorrect attracting fixed point}
\label{app:fold-theory}

Input conditioning can locally eliminate incorrect attracting fixed points. We establish this mechanism through a saddle-node result and 51 computer-assisted certificates in separately trained $4\times4$ Sudoku reasoners (Appendix~\ref{app:small-model-certificate}). The proposition below gives sufficient conditions for this local annihilation, and the certificates verify those conditions. Figure~\ref{fig:certified-folds} relates the certified events to the observed reasoning outcomes.

Let $e_0=e(x)$ and $e_1=e(x_H)$ be fixed input embeddings, and define
\begin{equation}
e_\lambda=(1-\lambda)e_0+\lambda e_1,
\qquad
F_\lambda(z)=F_\theta(z;e_\lambda),
\qquad \lambda\in[0,1].
\label{eq:fold-homotopy}
\end{equation}
This family varies the input embedding while keeping the model parameters fixed. The analysis concerns the smooth real-valued map defined by the stored, frozen parameters and embedding endpoints. Write $\widehat y_\lambda(z)$ for the decoded answer after the original-task projection $\Pi_x$; correctness is always assessed by $V(x,\widehat y_\lambda(z))$.

The local mechanism is a collision between an attracting fixed point and an unstable fixed point. Under the conditions below, the pair disappears on one side of the collision. If both states decode incorrectly near the collision, this eliminates an incorrect fixed-point attractor.

\paragraph{Proposition: a local wrong-answer fold.}
Suppose $F_\lambda$ is $C^3$ jointly in $(z,\lambda)$ near $(z_c,\lambda_c)$, with $\lambda_c\in(0,1)$, and:
\begin{enumerate}
\item $F_{\lambda_c}(z_c)=z_c$. The Jacobian $J_c=D_zF_{\lambda_c}(z_c)$ has an algebraically simple eigenvalue $+1$; all other eigenvalues have modulus strictly below one.
\item For real right and left eigenvectors $v,u$, normalized by $u^\top v=1$, the coefficients
\begin{equation}
a=u^\top\partial_\lambda F_{\lambda_c}(z_c),
\qquad
b=\tfrac12u^\top D_z^2F_{\lambda_c}(z_c)[v,v]
\label{eq:fold-coefficients}
\end{equation}
satisfy $a\ne0$ and $b\ne0$.
\item In a neighborhood of $(z_c,\lambda_c)$, decoding remains incorrect for the original task: $V(x,\widehat y_\lambda(z))=0$.
\end{enumerate}
Then there are sufficiently small state and parameter neighborhoods in which $F_\lambda$ has exactly two fixed points when $-a(\lambda-\lambda_c)/b>0$, and none when this expression is negative. At $\lambda=\lambda_c$, the two branches meet at $z_c$. One branch is locally asymptotically stable; the other has one unstable direction. Both decode incorrectly. In particular, if $ab>0$, increasing $\lambda$ through $\lambda_c$ locally annihilates an incorrect attracting fixed point and its unstable companion.

\emph{Proof.}
Apply Lyapunov--Schmidt reduction to $F_\lambda(z)-z=0$ at the critical point, writing $z=z_c+\xi v+w$, with $u^\top w=0$. The restriction of $J_c-I$ to the complementary subspace is invertible. The implicit function theorem therefore determines $w=w(\xi,\delta)$, where $\delta=\lambda-\lambda_c$, and $\partial_\xi w(0,0)=0$. Projecting the remaining equation onto $u$ gives
\begin{equation}
0=a\delta+b\xi^2+
O\!\left(\delta^2+|\delta\xi|+|\xi|^3\right).
\label{eq:fold-reduced-equation}
\end{equation}
The nonzero coefficients give the stated local pair/no-pair alternative of the saddle-node theorem~\citep{kuznetsov2004bifurcation}. Along the two branches, the critical eigenvalue is $1+2b\xi+O(|\delta|+\xi^2)$. It lies below one on one branch and above one on the other; sufficiently close to the fold it remains above $-1$. The remaining eigenvalues stay inside the unit circle by continuity. This establishes the stability distinction. The decoding assumption makes both branches incorrect. \hfill$\square$

\paragraph{Computer-assisted verification.}
We certify an exact critical point and its eigenvector by enclosing a zero of the augmented system
\begin{equation}
\mathcal H(z,v,\lambda)=
\begin{pmatrix}
F_\lambda(z)-z\\
(D_zF_\lambda(z)-I)v\\
c^\top v-1
\end{pmatrix},
\label{eq:fold-augmented-system}
\end{equation}
where $c$ is a fixed normalization vector. For an approximate root $\widehat\omega$, let $X=\widehat\omega+[-R,R]^{2d+1}$ and choose a numerical inverse preconditioner $A$. Outward-rounded interval evaluations bound
\begin{equation}
Y\ge\|A\mathcal H(\widehat\omega)\|_\infty,
\qquad
Z\ge\sup_{\omega\in X}\|I-A D\mathcal H(\omega)\|_\infty.
\end{equation}
The inequalities $Z<1$ and $Y+ZR<R$ make $\omega\mapsto\omega-A\mathcal H(\omega)$ a contraction mapping $X$ into itself. They certify a unique augmented root in $X$. Verified linear solves enclose its left eigenvector; interval evaluation then checks the nonzero left--right pairing and excludes zero from both coefficients in Equation~\ref{eq:fold-coefficients}.

These checks also establish simplicity of the critical eigenvalue. The verified bound on $I-A D\mathcal H$ implies that $D\mathcal H$ is invertible. If $\ker(J-I)$ had dimension greater than one, a nonzero change to $v$ in that kernel could preserve $c^\top v$, producing a kernel vector of $D\mathcal H$. Thus the eigenspace is one-dimensional. Moreover, a generalized eigenvector $w$ with $(J-I)w=v$ would imply $u^\top v=u^\top(J-I)w=0$, contradicting the certified pairing. The eigenvalue $+1$ is therefore algebraically simple.

Transverse stability is certified by a Lyapunov bound on the noncritical directions. Define at the exact enclosed root
\begin{equation}
B=J-\frac{vu^\top}{u^\top v}.
\label{eq:fold-deflation}
\end{equation}
The rank-one term is the spectral projector onto the critical direction. Thus $B$ replaces the eigenvalue $+1$ by zero and retains the other eigenvalues of $J$. Interval operations enclose $B$ using the enclosed $J,u,v$, including uncertainty in their pairing. A numerical Lyapunov solve proposes a symmetric matrix $P$; separate interval checks verify
\begin{equation}
P\succ0,
\qquad
r^2P-B^\top P B\succ0,
\qquad 0<r<1.
\label{eq:fold-lyapunov}
\end{equation}
The interval verification of these inequalities certifies that every noncritical eigenvalue has modulus below $r$.

As an alternative, interval bounds on a similarity-transformed Jacobian can give Gershgorin disks that isolate the unit multiplier and place every remaining disk strictly inside the unit circle. Four of the 51 complete certificates use this criterion; the other 47 use Equation~\ref{eq:fold-lyapunov}.

Strict margins for an incorrect decoder output certify that the local fixed-point branches remain wrong for the original task. For the linear decoder used in these experiments, token $a$ at position $i$ is constant over a state box if every competing token $b$ satisfies
\begin{equation}
\inf_{z\in X_z}(w_a-w_b)^\top z_{H,i}>0.
\label{eq:fold-decoder-margin}
\end{equation}
Strict margins at all positions make the complete decode locally constant. On a unique-solution task, a single certified admissible token $a\ne y_i^\star$ at an original non-given position already proves incorrectness, provided $\Pi_x$ preserves that token. The witness therefore remains wrong after the original givens are restored.

\paragraph{What the result establishes.}
The proposition and its verified instances establish that varying the input condition can locally annihilate an incorrect attracting fixed point. The certificate concerns the smooth real-valued map along the specified embedding interpolation; floating-point rollouts approximate this map, and intermediate embeddings need not represent valid discrete inputs. It certifies disappearance in a neighborhood of one critical point, rather than absence of other incorrect attractors or absence throughout the path to $\lambda=1$. Ordinary numerical continuation supplies evidence connecting the critical branch to the observed failure at $\lambda=0$; it is not a validated connection certificate. Thus the local result does not certify that the discrete ICS input replacement eliminates the original failure branch at its endpoint. Subsequent recovery and the inference cost of discovering useful conditions remain empirical outcomes, reported separately from the local proof.

\subsection{Local persistence of a correct solution}
\label{app:local-persistence}

We next ask when a correct solution can persist after changing the input condition. Let $s$ be a state that produces a valid answer under the original input. Under the conditioned input, define the returned prediction at state $z$ as
\begin{equation}
\widehat y_H(z)
=
\Pi_x\!\left(g_\theta(z;\tilde{x})\right),
\end{equation}
where $\Pi_x$ is the original-task output projection used by ICS. Correctness is always evaluated using
$V(x,\widehat y_H(z))$.

Because the decoder may depend on the input, correctness at $s$ under $x$ does not automatically imply correctness under $x_H$. The condition below explicitly concerns decoding under the conditioned input and evaluation against the original task.

\paragraph{A sufficient local condition.}
For $r>0$, let
\[
B(s,r)=\{z:\|z-s\|_2\le r\}
\]
be a closed ball around $s$. Suppose that:

\begin{enumerate}
\item Every state in the ball produces a valid answer
to the original task:
\[
V(x,\widehat y_H(z))=1
\qquad\text{for all }z\in B(s,r).
\]

\item The conditioned update is contractive on the ball:
there exists $0\le\kappa<1$ such that
\[
\|G_H(u)-G_H(v)\|_2
\le \kappa\|u-v\|_2
\qquad\text{for all }u,v\in B(s,r).
\]

\item Its displacement at the center satisfies
\begin{equation}
\delta_H(s):=\|G_H(s)-s\|_2
\le (1-\kappa)r.
\label{eq:local-persistence-condition}
\end{equation}
\end{enumerate}

Then $G_H$ has a unique fixed point $s_H$ within $B(s,r)$, and every trajectory initialized in this ball converges to $s_H$. Every iterate, including the limiting fixed point, produces a valid answer to the original task. Moreover,
\begin{equation}
\|s_H-s\|_2
\le \frac{\delta_H(s)}{1-\kappa}.
\label{eq:fixed-point-displacement}
\end{equation}

\emph{Proof.}
For any $z\in B(s,r)$,
\begin{align}
\|G_H(z)-s\|_2
&\le \|G_H(z)-G_H(s)\|_2+\|G_H(s)-s\|_2\\
&\le \kappa r+\delta_H(s)
\le r.
\end{align}
Thus, $G_H$ maps the closed ball into itself. The contraction mapping theorem gives a unique fixed point in the ball and convergence from every initialization within it. At this fixed point,
\[
\|s_H-s\|_2
\le \kappa\|s_H-s\|_2+\delta_H(s),
\]
which gives Equation~\ref{eq:fixed-point-displacement}. Correctness follows from the first assumption.
\hfill$\square$

\paragraph{Interpreting the conditions.}
The factor $\kappa$ bounds how strongly one update contracts distances within the ball. The displacement $\delta_H(s)$ measures how far the conditioned update moves its center. Equation~\ref{eq:local-persistence-condition} requires this displacement to be small enough for the ball to remain invariant.

A locally correct decoding region can arise when the conditioned logits vary continuously with the latent state and have a unique highest-scoring token at every output position at $s$. If the resulting prediction is valid, sufficiently small changes in the state preserve that prediction. This establishes the existence of a local decoding neighborhood, but does not by itself determine its radius or verify contraction.

If $s$ is an exact fixed point of the original map, then $\delta_H(s)=\Delta_H(s)$. For an approximate fixed point, the preceding residual bound instead gives
\[
\delta_H(s)\le\eta_F(s)+\Delta_H(s).
\]
Consequently, a sufficient condition for Equation~\ref{eq:local-persistence-condition} is
\begin{equation}
\eta_F(s)+\Delta_H(s)\le(1-\kappa)r.
\end{equation}

The result allows the correct fixed point to move: $s_H$ need not coincide with $s$. It guarantees persistence and convergence within the specified neighborhood, but does not establish that either a fresh initialization or a previously incorrect state reaches that neighborhood.

\subsection{Persistence does not imply accessibility}
\label{app:accessibility}

The preceding result guarantees convergence and correctness for trajectories initialized inside a specified neighborhood. It does not determine whether the fresh initialization $z_0$ or a previously incorrect state $w$ reaches that neighborhood.

\paragraph{Entry into a correct neighborhood.}
Let $B(s,r)$ satisfy the conditions of Appendix~\ref{app:local-persistence}, with conditioned fixed point $s_H$. For an initialization $u$, define its first entry time as
\begin{equation}
t_{\mathrm{ent}}(u)
=
\inf\left\{
t\in\mathbb{N}_0:
G_H^t(u)\in B(s,r)
\right\},
\label{eq:correct-region-entry}
\end{equation}
where $\inf\varnothing=\infty$.

If $t_{\mathrm{ent}}(u)$ is finite, invariance of the ball ensures that all subsequent states remain inside it.
Consequently, for every integer $T\ge t_{\mathrm{ent}}(u)$,
\begin{equation}
V\!\left(x,\widehat y_H(G_H^T(u))\right)=1.
\end{equation}
Moreover, for every integer $n\ge1$,
\begin{equation}
\left\|
G_H^{t_{\mathrm{ent}}(u)+n}(u)-s_H
\right\|_2
\le
\kappa^n
\left\|
G_H^{t_{\mathrm{ent}}(u)}(u)-s_H
\right\|_2.
\end{equation}

Local contraction therefore controls convergence \emph{after entry}; it supplies no bound on the entry time itself. For a finite inference budget $T$, $t_{\mathrm{ent}}(u)\le T$ is a sufficient condition for a correct endpoint. It is not necessary: a trajectory may produce a valid answer elsewhere, without entering this particular ball.

\paragraph{Cold success can coexist with warm failure.}
Suppose the fresh initialization reaches the correct neighborhood, so that $t_{\mathrm{ent}}(z_0)<\infty$. Now suppose the previously incorrect state lies in a region $U$ satisfying
\begin{equation}
w\in U,
\qquad
G_H(U)\subseteq U,
\qquad
V(x,\widehat y_H(z))=0
\quad\text{for all }z\in U.
\label{eq:invariant-incorrect-region}
\end{equation}
Then the cold-start trajectory produces valid answers from its entry time onward, whereas the warm-start trajectory remains incorrect at every step. The latter follows directly by induction:
$G_H^t(w)\in U$ for all $t\ge0$.

Thus, the same conditioned map can support a correct solution accessible from $z_0$ while retaining an incorrect outcome accessible from $w$. In particular, successful cold reasoning does not require every incorrect fixed point or invariant incorrect region to disappear.

\paragraph{Implication for hypothesis search.}
The existence of a correct solution under a hypothesis and its accessibility from the chosen initialization are separate requirements.
ICS evaluates each hypothesis from $z_0$, so its relevant accessibility question concerns trajectories starting from that initialization. Warm-start interventions probe a different question: whether the conditioned map can recover directly from a previously failed state.

These conditions explain how the two outcomes can differ; they do not establish that cold initialization is universally preferable. The entry and invariance properties depend on the particular model, input, hypothesis, and initialization.

\paragraph{Scope and empirical interpretation.}
The fold certificates in Appendix~\ref{app:small-model-certificate} verify sufficient conditions for local elimination of particular incorrect attracting fixed points. The residual bound additionally relates the measured input-induced response to departure from stationarity at the saved state.

The correct-solution persistence and accessibility results require separate assumptions about decoding neighborhoods, contraction, and trajectory entry. Center residuals, estimated Jacobian spectral radii, and sampled perturbation responses provide complementary diagnostics, but neither these measurements nor the fold certificates verify all of those assumptions throughout a neighborhood. Correct predictions can also occur on moving trajectories, so convergence to a fixed point is not necessary for successful reasoning.

The analysis identifies mechanisms through which input conditioning and initialization affect reasoning. Which hypotheses ICS discovers, and the computation required to find and select a correct answer, are evaluated empirically.

\end{document}